%% file: main.tex
\documentclass[11pt, a4paper]{memoir}
\usepackage[english, science, titlepage, dropcaps]{ku-frontpage}
\usepackage[utf8]{inputenc}
\usepackage[T1]{fontenc}
\usepackage{graphicx, wrapfig, subcaption, setspace, booktabs}
\usepackage{tabularx}
\usepackage{caption}
\usepackage[colorlinks=true,linkcolor=blue,citecolor=blue]{hyperref}
\usepackage{xcolor}

\hypersetup{
            colorlinks=true,
            linkcolor=black,
            citecolor=blue,
}

\assignment{MSc thesis in Computer Science}
\author{Lukas Muttenthaler}

\title{Subjective Question Answering}
\subtitle{Deciphering the inner workings of Transformers in the realm of subjectivity}
\date{Handed in: \today}
\advisor{Advisors: Johannes Bjerva, Isabelle Augenstein}

\begin{document}
\maketitle

\tableofcontents
\newpage

\chapter*{Abstract}
\label{section:abstract}
\input{abstract.tex}

\chapter{Overview}
\label{section:overview}
\input{overview.tex}

\chapter{Introduction}
\label{section:intro}
\input{introduction.tex}

\chapter{Background}
\label{section:background}
\input{background.tex}

\chapter{Methodology}
\label{section:method}
\input{method.tex}

\chapter{Data}
\label{section:datasets}
\input{data.tex}

\chapter{Quantitative Analyses}
\label{section:results}
\input{quantitative_analysis.tex}

\chapter{Qualitative Analyses}
\label{section:qualitative_analysis}
\input{qualitative_analysis.tex}

\chapter{Discussion}
\label{section:discussion}
\input{discussion.tex}

\chapter{Summary}
\label{section:summary}
\input{summary.tex}

\chapter{Acknowledgments}
\label{section:acknowledgements}
\input{acknowledgements.tex}

%% BIBLIOGRAPHY %%
\bibliographystyle{splncs04}
\bibliography{references}

\end{document}

%% file: abstract.tex
Understanding subjectivity demands reasoning skills beyond the realm of common knowledge. It requires a machine learning model to process sentiment and to perform opinion mining. In this work, I've exploited a recently released dataset for span-selection Question Answering (QA), namely SubjQA \cite{subjqa2020}. SubjQA is the first QA dataset to date that contains questions that ask for subjective opinions corresponding to review paragraphs from six different domains, namely books, electronics, grocery, movies, restaurants, and TripAdvisor. Hence, to answer these subjective questions, a learner must extract opinions and process sentiment for various domains, and additionally, align the knowledge extracted from a paragraph with the natural language utterances in the corresponding question, which together enhance the difficulty of a QA task. In the scope of this master's thesis, I inspected different variations of BERT \cite{devlin2018bert}, a neural architecture based on the recently released Transformer \cite{DBLP:conf/nips/VaswaniSPUJGKP17}, to examine which mathematical modeling approaches and training procedures lead to the best answering performance. However, the primary goal of this thesis was not to solely demonstrate state-of-the-art performance but rather to investigate into the inner workings (i.e., latent representations) of a Transformer-based architecture to contribute to a better understanding of these not yet well understood "black-box" models.

One of the key insights of this work reveals that a Transformer's hidden representations, with respect to the true answer span, are clustered more closely in vector space than those representations corresponding to erroneous predictions. This observation holds across the top three Transformer layers for both objective and subjective questions, and generally increases as a function of layer dimensions. Moreover, the probability to achieve a high cosine similarity among hidden representations in latent space concerning the true answer span tokens is significantly higher for correct compared to incorrect answer span predictions. These statistical results have decisive implications for down-stream applications, where it is crucial to know about why a neural network made mistakes, and in which point in space and time the mistake has happened (e.g., to automatically predict correctness of an answer span prediction without the necessity of labeled data).

Quantitative analyses have shown that Multi-task Learning (MTL) does not significantly improve over Single-task Learning (STL). This might be due to one of the leveraged auxiliary tasks being unsolvable. It appears as if BERT produces domain-invariant features by itself, although further investigation must go into this line of research to determine whether this observation holds across other datasets and domains.  Fine-tuning BERT with additional Recurrent Neural Networks (RNNs) on top improves upon BERT with solely one linear output layer for QA. This is most likely due to a more fine-grained encoding of temporal dependencies between tokens through recurrence forward and backward in time, and is in line with recent work.

%% file: overview.tex
I will begin this thesis with an  \hyperref[section:intro]{\textbf{Introduction}} comprising an overview of the topic being explored. In so doing, I will explain my motivations in conducting this research and outline the importance of continued research on various neural architectures in this field. Following that, I will outline the \hyperref[section:rq]{\textbf{Research Questions (RQs)}} I aim to answer. Thereafter, in the \hyperref[section:background]{\textbf{Background}} section  I will introduce the task of \hyperref[section:question_answering]{\textbf{Question Answering (QA)}} and discuss to which of the various QA versions I will confine myself in the scope of this master's thesis. 

This is followed by an overview of the model architectures that will be leveraged in the different experiments. I will start with explaining the \hyperref[section:transformers]{\textbf{Transformer}}, elaborate on the mechanisms behind \hyperref[section:bert_intro]{\textbf{BERT}} which is a transformer-based architecture. Moreover, I will discuss the mathematical details with respect to \hyperref[method:birnns]{\textbf{Recurrent Neural Networks (RNNs)}} and \hyperref[method:highway]{\textbf{Highway Networks}}. To conclude the background section, I will discuss the notion of \hyperref[section:multitask]{\textbf{Multi-task learning}}. 

In the \hyperref[section:method]{\textbf{Methodology}} section of the thesis, I will explain the different models, the task(s), and most importantly all relevant computations that are necessary to optimize the models concerning the respective task(s). 

The elaboration of the methods is followed by a detailed overview of the \hyperref[section:datasets]{\textbf{datasets}} that are exploited to train and evaluate the neural architectures. In this section, I will provide an in-depth analysis of the datasets to both qualitatively and quantitatively assess their nature before any model training.

In the \hyperref[section:results]{\textbf{Quantitative Analysis}} section, results concerning all conducted experiments will be presented, explained and discussed. Note that a thorough interpretation of the results will follow in the \hyperref[section:discussion]{\textbf{Discussion}} part and hence interpretation is constrained in this section. Ad hoc elaboration on results may be provided but I refer the interested reader to the \hyperref[section:discussion]{\textbf{Discussion}} section for in-depth interpretations.

Numeric results must be connected to visualizations of models' feature representations in vector space in order to understand the breakthroughs and shortcomings of Machine Learning (ML) models. Hence, an in-depth \hyperref[section:qualitative_analysis]{\textbf{Qualitative Analysis}} of the hidden representations with respect to selected neural architectures follows the depiction of quantitative results. Alongside this I provide an error analysis to identify the issues the models faced at inference time. Here, I will try to answer \textbf{where} along the way and \textbf{why} a learner made mistakes.

Last but not least, I will \hyperref[section:discussion]{\textbf{discuss}} the results obtained from both types of analyses, draw conclusions and close with a concise \hyperref[section:summary]{\textbf{Summary}} of the thesis to provide a synopsis free of the hefty details.

%% file: introduction.tex
Thoroughly understanding the full nature of subjectivity is a daunting task for both humans and machines \cite{banfield1982,quirk1985comprehensive,DBLP:conf/aaai/Wiebe00,DBLP:journals/lre/WiebeWC05}. Whether it is a subjective thought, an opinion, a question, or an answer, all of it highly depends on the context the respective natural language utterance appears in \cite{DBLP:conf/acl/MihalceaBW12,DBLP:journals/lre/WiebeWC05}. It is often not simple to decipher what is and what is not subjective \cite{banfield1982,DBLP:conf/acl/MihalceaBW12}. A question might be subjective but its answer contains an objective, measurable fact, and vice versa \cite{DBLP:conf/aaai/Wiebe00,subjqa2020}. Due to the frequent exchange of opinions in a world greatly embedded in social media, subjectivity in natural language has become highly pervasive. This fact alone makes the task of examining how machines read natural language texts that contain subjective opinions worth pursuing. However, I would like to further stress why I encourage the field of Artificial Intelligence (AI) to shed light on the development of systems that possess the ability to answer questions concerning subjective opinions.

Machine Reading, also called span-selection Question Answering (QA) or Reading Comprehension (RC), has a long-standing history in the fields of Information Retrieval (IR) and Natural Language Processing (NLP). Over the past two decades, of which the last in particular yielded breakthroughs in NLP, machine reading has recorded vast advancements. An array of systems has been developed to enhance machine comprehension systems \cite{DBLP:journals/corr/TrischlerWYHSBS16,DBLP:journals/corr/SeoKFH16,DBLP:journals/corr/WeissenbornWS17,DBLP:journals/corr/abs-1806-08727} and numerous different RC datasets have been created to train these \cite{downeygetting}. Although much work has been going on in the entirety of open-domain QA \cite{DBLP:conf/semweb/CabrioCAMLG12,DBLP:journals/ipm/RyuJK14,DBLP:conf/acl/ChenFWB17,DBLP:conf/aaai/WangYGWKZCTZJ18}, I will in this project exclusively focus on the task of finding an answer span in a corresponding natural language context, i.e. span-selection RC.

\begin{figure}[h!]
    \centering
    \includegraphics[width=.63\textwidth]{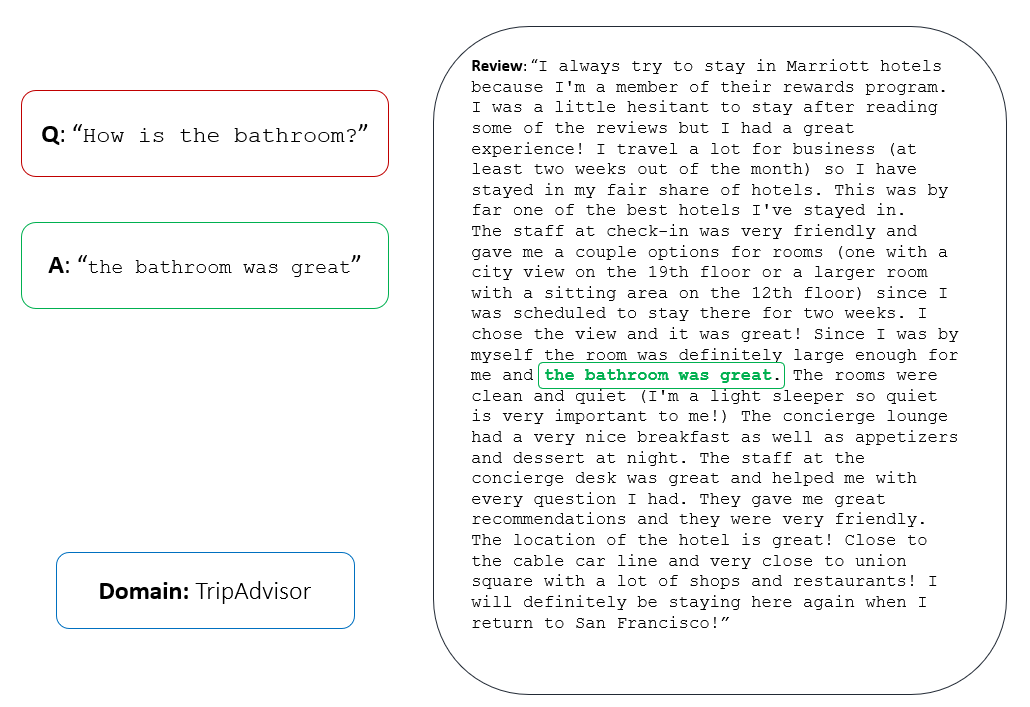}
    \caption{QA example from SubjQA \cite{subjqa2020}. The correct answer is a text span of $n$ character sequences in the review paragraph corresponding to the subjective question. The span was identified through human crowd-workers before model training. As such, QA is considered a span-selection task, where both start and end position of the correct text span must be predicted by a neural network.}
    \label{fig:qa_example}
\end{figure}

The task of answering questions that contain objective, measurable facts appears to be resolved to a large extent for answerable questions \cite{DBLP:journals/corr/RajpurkarZLL16,DBLP:journals/corr/abs-1804-07461}. In so doing, SQuAD v1.0 \cite{DBLP:journals/corr/RajpurkarZLL16} was the first-ever large-scale dataset that fostered the latter development. As a result, the researchers centered around SQuAD recently developed a more complex dataset that consists of questions that are not answerable, namely SQuAD v2.0 \cite{DBLP:journals/corr/abs-1806-03822}. As can be inferred from the publicly available leader board regarding this task, it appears to be more difficult for models across the board to understand that a question cannot be answered with the given context. This might sound paradoxical at first sight, but if one recalls that humans frequently face the task of acknowledging the fact that in certain cases there simply is no answer, the latter becomes more apparent.

What has been lacking until recently, however, was both a dataset that not only includes unanswerable questions that consist of an objective, measurable fact but is also rich in questions that contain a subjective opinion and a corresponding machine reading system that is capable of understanding and hence answering such questions. At this point, I would like to stress the fact that when I speak about \textsc{understanding} subjectivity, I refer to reading a paragraph and finding the correct answer span within this paragraph (see Figure~\ref{fig:qa_example} for an example). I am aware of the fact that utterly \textsc{understanding} subjectivity is beyond the scope of current methods in ML \cite{DBLP:conf/acl/MihalceaBW12,DBLP:journals/lre/WiebeWC05,DBLP:conf/aaai/Wiebe00}. 

The vast majority of QA datasets is factoid and concerns solely a single domain such as SQuAD v1.0 \cite{DBLP:journals/corr/RajpurkarZLL16}, SQuAD v2.0 \cite{DBLP:journals/corr/abs-1806-03822}, WikiQA \cite{yang-etal-2015-wikiqa}, WikiReading \cite{DBLP:conf/acl/HewlettLJPFHKB16} or CNN/Daily Mail \cite{NIPS2015-5945} of which all but the last dataset are exclusively based on Wikipedia. Recent research in NLP that scrutinized QA datasets revealed that such datasets do not necessarily examine Natural Language Understanding (NLU) abilities, as complex reasoning skills are often not required to perform well \cite{downeygetting,DBLP:journals/corr/abs-1911-09241}. SubjQA, the dataset that I am going to exploit in this study, is the first dataset to date that includes subjective opinions extracted directly from reviews written by humans, does consist of texts corresponding to multiple domains, and includes a high number of unanswerable questions \cite{subjqa2020}. The latter set of questions has been proven to be particularly difficult since a machine reading system must understand that a question cannot be answered from the given context \cite{DBLP:journals/corr/abs-1806-03822}.

Although originality often makes research questions worth pursuing, any research question must come with both a purpose for society and an adequate justification for the pursued avenue. If we, as researchers in the field of ML, develop methods to better understand and analyze how subjectivity and the respective context it appears in is reflected in a neural model, society might benefit from more sophisticated chatbots, search engines, and voice assistants among others shortly. Hence, I will in this thesis contribute to the investigation of both natural language data that contains subjectivity and the behavior of neural machines when faced with the latter as much as time and space allow. In the following sections, I will introduce the general task of QA, both different neural network architectures and training techniques that play a crucial role with respect to the systems that I plan to inspect.

\section{Related Work}
\label{section:related_work}

I will, in this work, investigate different neural architectures for span-selection Question Answering (QA)  based on the Transformer \cite{DBLP:conf/nips/VaswaniSPUJGKP17} (see Section~\ref{section:question_answering} in the Background section for a detailed elaboration on QA, and Figure~\ref{fig:qa_example} for a general overview of the task). In so doing, I will inspect different mechanisms (e.g., multi-task learning, sequential knowledge transfer)  and training procedures (e.g., adversarial training, different task sampling strategies) to enhance performance concerning subjective QA. In addition, I will look deeper into the inner workings, i.e. hidden representations, of Transformer models at each layer stage. This is done to decipher \texttt{how} neural networks answer subjective questions, and unveil \texttt{where} along the way and \texttt{why} they make mistakes. A thorough qualitative analysis of model behavior appears crucial, given that deep neural networks (DNNs) are often considered "black-box" models that require better understanding by the community \cite{lecun2015deep}.

The recent advent of Transformer models \cite{DBLP:conf/nips/VaswaniSPUJGKP17} and NLP models based on the latter such as ELMo \cite{DBLP:conf/naacl/PetersNIGCLZ18}, BERT \cite{devlin2018bert}, and RoBERTa \cite{DBLP:journals/corr/abs-1907-11692}, has yielded an enormous flux of studies that draws attention to NLP in general and open-domain QA in particular. One recent study that is similar to this work conducted a layer-wise analysis of BERT's Transformer layers to investigate how BERT answers questions \cite{bert-qa-layerwise}. For each of BERT's Transformer layers, they projected the model's high-dimensional hidden representations into $\mathbf{R}^{2}$ to visually depict how BERT clusters different parts of an input sequence (i.e., question, context, answer) while searching for the correct answer span in latent space. The main difference, however, is that the aforementioned study exclusively conducted a qualitative analysis of BERT's hidden representations without the endeavor to implement different model versions to quantitatively inspect performance concerning QA. Moreover, BERT was fine-tuned on factoid and not on subjective questions which most likely yields different QA behavior and hence different feature representation patterns in latent space as both opinions and sentiment are more relevant than objective, measurable facts to answer a subjective question. Their attempt to explain QA behavior through thoroughly analyzing BERT's hidden representations at various layer stages was, nevertheless, remarkable and a crucial step forward towards explainability in AI which is why I will follow their approach concerning the qualitative analysis of feature representations in vector space, and inspect whether their results are replicable for the realm of subjectivity. 

Another study that has been published recently, developed a dataset that significantly differs from most recent QA datasets. As mentioned at the beginning of this section, the vast majority of QA datasets is factoid and concerns only a single domain \cite{DBLP:journals/corr/RajpurkarZLL16,yang-etal-2015-wikiqa,NIPS2015-5945}. Their dataset followed the attempt to explicitly avoid questions that may be answered with common knowledge or knowledge about one domain \cite{downeygetting}. This attempt follows a similar motivation for the development of SubjQA \cite{subjqa2020}. They created a dataset that consists of four domains of which two cannot be answered with pre-training on corpora that contain common knowledge (e.g., "During which period was Bill Clinton president of the United States of America?"). However, the dataset is with 800 texts not particularly large and does not include any questions with respect to subjective opinions of humans. Moreover, the study exclusively focused on dataset development and analysis without looking into the behaviour of SOTA NLP models while answering questions. The latter is decisive to both understand how neural networks process natural language utterances contained in the dataset which potentially yields insights into the quality of the respective dataset, and whether human annotations are reliable sources.

Arkhangelskaia et al., 2019 \cite{DBLP:journals/corr/abs-1910-06431} investigated which tokens in question - context sequence pairs receive particular attention by BERT's self-attention mechanisms to answer a question, and how the multi-headed attention weights change across the different layers in BERT. Similarly to \cite{bert-qa-layerwise}, the authors did solely conduct a qualitative analysis of the model. Contrary to \cite{bert-qa-layerwise}, the study focused on a single implementation of BERT and exclusively exploited SQuAD \cite{DBLP:journals/corr/RajpurkarZLL16,DBLP:journals/corr/abs-1806-03822} without inspecting BERT's behaviour with respect to other, more challenging QA datasets where contexts belong to different domains. 

Both Bingel \& Søgaard, 2017 \cite{DBLP:conf/eacl/SogaardB17}, and Bjerva, 2017 \cite{DBLP:conf/nodalida/Bjerva17} investigated the relatedness between auxiliary and main tasks in different multi-task learning (MTL) paradigms. They analyzed the importance of relations between tasks, and under which conditions auxiliary tasks unfold to be beneficial for the main task(s). However, they exclusively scrutinized classification and structured prediction tasks. Structured prediction tasks are tasks, where the model is meant to predict symbols rather than real values. In NLP such tasks are summarized under the umbrella term of sequence tagging or labeling (e.g., Part-of-Speech tagging, Named Entity Recognition). Moreover, they solely deployed Recurrent Neural Networks (RNNs) \cite{DBLP:journals/cogsci/Elman90}. In contrast, I will for the first time investigate different MTL settings for span-selection QA and in so doing leverage the knowledge of a pre-trained Transformer model \cite{DBLP:conf/nips/VaswaniSPUJGKP17}, namely BERT \cite{devlin2018bert}.

Numerous studies have worked on the development of QA datasets or put effort into the advancement of RC models that perform well on them. Little, however, has gone into the examination of non-factoid questions, which are questions that cannot be answered with objective, measurable facts but require models to understand subjectivity without resorting to common knowledge that might be present in large pre-training corpora. BERT is complex enough to perform incredibly well on factoid questions that correspond to paragraphs from a single domain and frequently require common knowledge to be answered \cite{devlin2018bert,DBLP:journals/corr/abs-1804-07461}. Whether BERT does also achieve close to human performance on datasets that barely consist of factoid questions, require little common knowledge, and contain texts from multiple domains is yet to be deciphered and will play a major role in the current study.

\section{Research Questions}
\label{section:rq}

In this study, I will examine the following research questions (RQs) as thoroughly as space and time allow.

\begin{enumerate}
    \item \texttt{Is} it necessary to fine-tune a Transformer model on a span-selection QA dataset that consists of subjective questions and multiple domains, namely SubjQA \cite{subjqa2020}, to achieve SOTA performance with respect to the latter? Or is it sufficient to leverage the knowledge of a pre-trained Transformer model, namely BERT \cite{devlin2018bert}, that was previously fine-tuned on SQuAD? SQuAD is a span-selection QA dataset that exclusively contains objective questions with respect to a single domain \cite{DBLP:journals/corr/RajpurkarZLL16,DBLP:journals/corr/abs-1806-03822}.
    \item \texttt{Do} additional encoding layers that are not exploited by neural architectures based on the Transformer \cite{DBLP:conf/nips/VaswaniSPUJGKP17}, such as Long-Short Term Memories (LSTMs) \cite{hochreiter1997long} or Highway Networks \cite{DBLP:journals/corr/SrivastavaGS15}, on top of BERT enhance information processing to an extent such that QA performance is increased? This research question is based on one recent study that has shown that encoding temporal dependencies among tokens through Recurrent Neural Networks (RNNs) \cite{DBLP:journals/cogsci/Elman90} helps BERT for QA \cite{hu2019question}. However, this was the first study to date that has investigated the latter and their experiments were performed solely with respect to SQuAD. Thus, it is both worth replicating their results and analysing whether this hold for subjective questions too.
    \item \texttt{To} which extent does multi-task learning (MTL) and adversarial learning techniques enhance BERT's answering behaviour with respect to subjective questions corresponding to multiple domains? In so doing, I will leverage the crowd-sourced human labels concerning the degree of subjectivity \cite{subjqa2020}, and adversarial training methods such as Gradient Reversal Layers (GRLs) \cite{ganin2014unsupervised}.
    \item \texttt{How} does model performance differ as a function of review domains? Since SubjQA contains review paragraphs corresponding to multiple domains it appears crucial to investigate a learner's performance with respect to these domains, and inspect whether some domains are more difficult than others.
    \item \texttt{Is} it possible to infer the difficulty of subjective questions from the interrogative words (e.g., \texttt{how}, \texttt{what}, \texttt{which}) that introduce them? If so, \texttt{how} does the degree of difficulty differ among them?
    \item \texttt{Which} qualitative insights can be extracted from a Transformer's hidden representations in latent space? Is it possible to decipher \texttt{where} along the way (i.e., in which layer) a learner made mistakes to better understand \texttt{why} an answer span was not predicted correctly? Moreover, is there a difference between subjective and objective questions with respect to a model's answer span search in latent space? If there is a difference, \texttt{which} is it, and \texttt{how} could this insight be leveraged?
\end{enumerate}

%% file: background.tex
\section{Question Answering} 
\label{section:question_answering}

The task of Question Answering (QA) has a long-established history in the connected fields of Information Retrieval (IR) and Natural Language Processing (NLP). Since this thesis is centered around a research project in NLP, I will confine myself to the position of QA within the latter research area. Hence, the role of search as a crucial part in finding relevant documents, which mainly belongs to the realm of IR, will not be discussed here. The focus of the thesis lies in reading and not retrieving. There are different versions of QA, namely closed-domain and open-domain QA. Since I will exclusively focus on open-domain QA, I will briefly introduce the former and elaborate more thoroughly on the latter.

Closed-domain QA deals with questions that concern a specific domain \cite{aqua-closed,closed-domain-semantic,closed-domain-QA-2014}, e.g., chemistry, pharmacy or neurology. This can be useful if one aims at analyzing numerous research papers in one of the aforementioned fields to either write a review paper or conduct a meta-analysis, or would like to perform plagiarism detection concerning a certain domain. Closed-domain QA, however, is restricted to a confined domain, does not require common knowledge, and is forced to exploit notably smaller datasets than open-domain QA, as resources are sparse. Hence, closed-domain heavily relies on ontologies such as knowledge graphs which often contain a large amount of factual information.

Open-domain QA was initially defined as finding answer spans in collections of unstructured, raw text \cite{DBLP:conf/acl/ChenFWB17}. In open-domain QA, a Machine Comprehension (MC) system must retrieve the relevant documents to answer the respective question \cite{DBLP:conf/semweb/CabrioCAMLG12,DBLP:journals/ipm/RyuJK14,DBLP:conf/acl/ChenFWB17,DBLP:conf/aaai/WangYGWKZCTZJ18}. This is mainly performed through IR search methods. In so doing, the system is first required to understand which documents are decisive to perform the latter step. After finding those documents, the system has to search for the correct text span within the corresponding natural language paragraphs to correctly respond to the question. In earlier versions of open-domain QA, structured data such as ontologies, databases or Knowledge Bases (KBs) such as the popular Freebase KB \cite{DBLP:conf/sigmod/BollackerEPST08} were frequently exploited to develop and evaluate QA systems (e.g., \cite{DBLP:conf/naacl/YuLZZR18}). Due to their limitations and expensiveness, however, recent research in QA has shifted its attention again towards finding answers in pieces of raw text rather than retrieving information from KBs. Interestingly, this was the focus of QA in the first place. In this study, I will exclusively draw attention to documents of unstructured, raw natural language text without the utilization of KBs.

Span-selection QA, or RC, may be perceived as a sub-field of open-domain QA, and will be the focus of this thesis. As such RC is concerned with the development of Machine Reading (MR) systems that are capable of finding an answer span in the documents that were previously retrieved via IR methods. Recall that the latter represents the first step in any open-domain QA setting. The systems are meant to first read a short paragraph regarding a certain domain about which a question is asked. As a next step, they must find the correct answer span in the paragraph whenever a question is answerable. There are cases where a question cannot be answered given the corresponding context \cite{DBLP:journals/corr/abs-1806-03822}. The correct answer then simply evaluates to the empty string. The latter problem has been addressed only recently and remains open to investigation. To foster an examination in this area, similarly to SQuAD v2.0 \cite{DBLP:journals/corr/abs-1806-03822}, a significant number of questions in SubjQA is unanswerable \cite{subjqa2020}. 

\section{Transformers}
\label{section:transformers}

The recent advent of the Transformer \cite{DBLP:conf/nips/VaswaniSPUJGKP17} has yielded an enormous deluge of studies concerning neural architectures, particularly in NLP. Never before has NLP received as much attention as since the release of the Transformer. Among the best-performing Natural Language Understanding (NLU) systems are solely models that leverage this neural architecture in one way or another, as can be inferred from publicly available leaderboards such as GLUE \cite{DBLP:journals/corr/abs-1804-07461} \footnote{\url{https://gluebenchmark.com/leaderboard/}}, SuperGLUE \cite{DBLP:conf/nips/WangPNSMHLB19} \footnote{\url{https://super.gluebenchmark.com/}}, SQuAD v1.0 \cite{DBLP:journals/corr/RajpurkarZLL16} \footnote{\url{https://rajpurkar.github.io/SQuAD-explorer/explore/1.1/dev/}}, SQuAD v2.0 \cite{DBLP:journals/corr/abs-1806-03822} \footnote{\url{https://rajpurkar.github.io/SQuAD-explorer/explore/v2.0/dev/}}, or WikiSQL \cite{DBLP:journals/corr/abs-1709-00103} \footnote{\url{https://github.com/salesforce/WikiSQL}}. Owing to this recent development in NLP I will exclusively exploit architectures that are based on the Transformer. Hence, I will now provide a short introduction about the general concept and the mathematical details behind this neural model. 

Contrary to Recurrent Neural Networks (RNNs) \cite{DBLP:journals/cogsci/Elman90} and Convolutional Neural Networks (CNNs) \cite{DBLP:conf/shape/CunHBB99}, the Transformer \cite{DBLP:conf/nips/VaswaniSPUJGKP17} does neither leverage temporal dependencies between timesteps through recurrence nor spatial features through filters and sliding windows respectively. The architecture is exclusively based on attention mechanisms, first introduced by Bahdanau et al., 2014 \cite{DBLP:journals/corr/BahdanauCB14}, which makes it highly parallelizable and computationally efficient. As such, it does not suffer from the same memory constraints as sequential computation does. The main advantage of exploiting attention mechanisms is that dependencies between timesteps (e.g., tokens) can be modelled independent of their position in the input sequence or distance to a token in the output sequence such as in Machine Translation (MT) where learning dependencies between distant positions is indispensable to correctly map words from one language to another \cite{DBLP:journals/corr/BahdanauCB14,DBLP:conf/emnlp/LuongPM15,DBLP:conf/nips/VaswaniSPUJGKP17}.

In sequential computation an input sequence is summarized as the recursively computed hidden representation of which each element at timestep $t$ is dependent on both the hidden representation at the previous timestep $t-1$ and the current input at $t$ (this is explained in more detail in Section~\ref{method:birnns}). In so doing, all inputs are weighted equally. This bears the constraint that early positions in an input sequence may be overwritten by later positions or simply weighted less. In the worst case, this can lead to a phenomenon called catastrophic forgetting and hence result in an enormous performance decline for longer sequences \cite{DBLP:journals/corr/BahdanauCB14,DBLP:conf/emnlp/LuongPM15}. In contrast, self-attention or intra-attention mechanisms possess the ability to relate positions in an input sequence independent of their distance to each other, yielding a more informative and richer representation of an input sequence \cite{DBLP:journals/corr/BahdanauCB14,DBLP:conf/emnlp/LuongPM15,DBLP:conf/nips/VaswaniSPUJGKP17,devlin2018bert}. This alleviates the aforementioned constraint that models solely based on recurrence suffer from. The Transformer is the first neural architecture to date that exclusively leverages such self-attention mechanisms without utilizing any recurrence or convolutions in its computation of sequence representations. As such, it only consists of self-attention and point-wise, fully connected layers stacked on top of each other.

The Transformer's self-attention mechanisms employ a scaled version of the dot-product attention, first introduced in \cite{DBLP:conf/emnlp/LuongPM15}. The difference between dot-product and scaled dot-product attention is the scaling factor $\frac{1}{\sqrt{d_{k}}}$, where $d_{k}$ corresponds to the dimensions of query and keys, and was introduced to counteract the potential vanishing gradient problem the softmax function might suffer from when $d_{k}$ becomes large  \cite{DBLP:conf/nips/VaswaniSPUJGKP17}. 

\begin{figure}[t]
\centering
\captionsetup{justification=centering}
\includegraphics[width=1.1\textwidth]{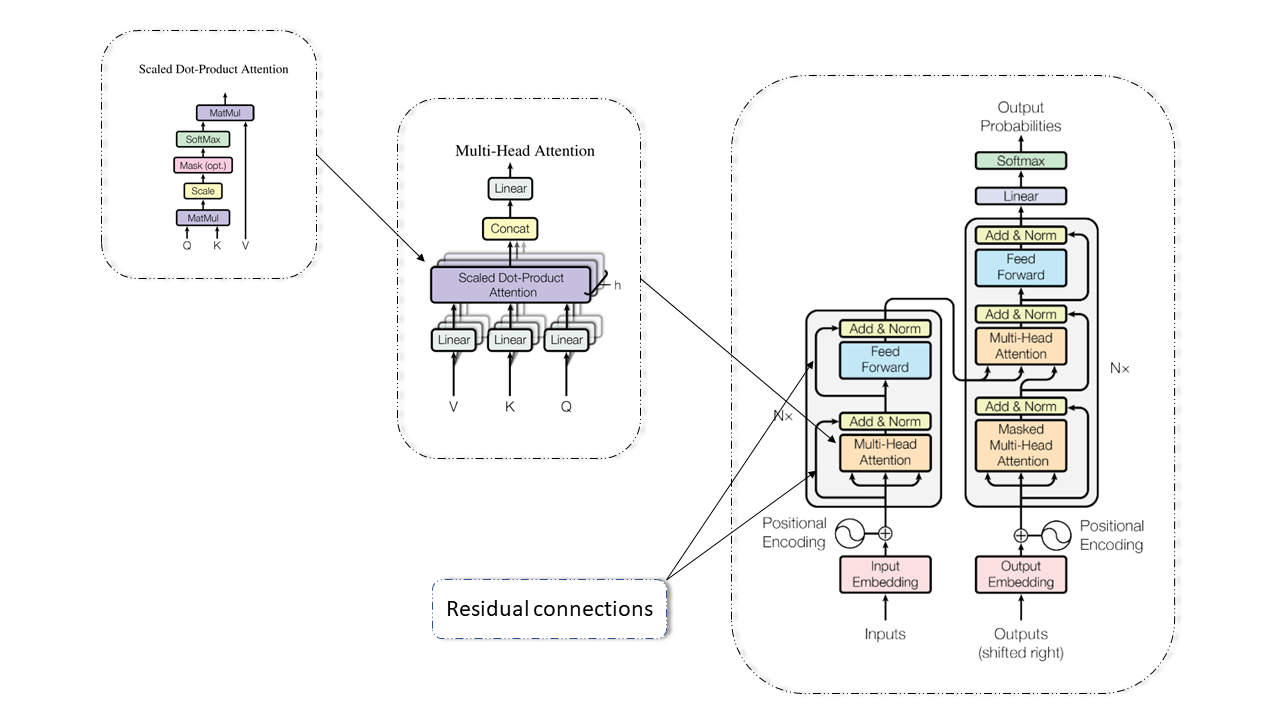}
\caption{Scaled Dot-Product Attention, Multi-Head Attention and the Transformer architecture \cite{DBLP:conf/nips/VaswaniSPUJGKP17}.}
\label{fig:transformer}
\end{figure}

The input to the attention function contains queries and keys of dimension $d_{k}$ and values of dimension $d_{v}$. The sets of query, keys, and values, are represented as the three matrices Q, K, and V, and simply evaluate to linear layers (see Figure~\ref{fig:transformer}). To obtain attention weights, a softmax function - the softmax function will be explained in more detail in Section~\ref{section:method} but basically maps a vector of continuous values to a discrete probability distribution - is applied to the inner product of Q and K, scaled by $\frac{1}{\sqrt{d_{k}}}$.

The matrix of attention weight vectors is then multiplied with V to obtain a weighted version of the values V (see leftmost rectangle in Figure~\ref{fig:transformer}). The computation can be summarized as follows,

\begin{equation}
    \textrm{Attention}(Q, K, V)=\textrm{softmax}\left(\frac{Q K^{T}}{\sqrt{d_{k}}}\right) V.
\label{equation:scaled_dot_product}
\end{equation}

This attention function is performed $h$ times in parallel (see center rectangle in Figure~\ref{fig:transformer}) instead of just once. The $h$ output values, each of dimension $d_{v}$, are then concatenated, and passed through another linear layer. The outputs of this self-attention layer serve as inputs to a point-wise, feed-forward neural network, together forming a single Transformer block (see rightmost rectangle in Figure~\ref{fig:transformer}). 

A Residual connection \cite{DBLP:conf/cvpr/HeZRS16} is applied to each sub-layer, followed by layer normalization \cite{DBLP:journals/corr/BaKH16}. Residual connections contribute to richer feature representations through adding function inputs to function outputs \cite{DBLP:conf/cvpr/HeZRS16}. Layer normalization significantly reduces the training time of deep neural architectures through stabilizing activities of individual neurons within layers, thus making a model computationally more efficient \cite{DBLP:journals/corr/BaKH16}.

Based on this architecture, a large number of neural Language Models (LMs) has been developed recently to tackle an array of problems in NLP, and in so doing outperforming more traditional approaches. Of those, BERT \cite{devlin2018bert} has been proven to be the most successful, and thus received ample attention lately. I will now explain the mechanisms behind BERT, and elaborate on why it is more successful than other architectures. 

\subsection{BERT}
\label{section:bert_intro}

\begin{figure}[h!]
\centering
\includegraphics[width=1\textwidth]{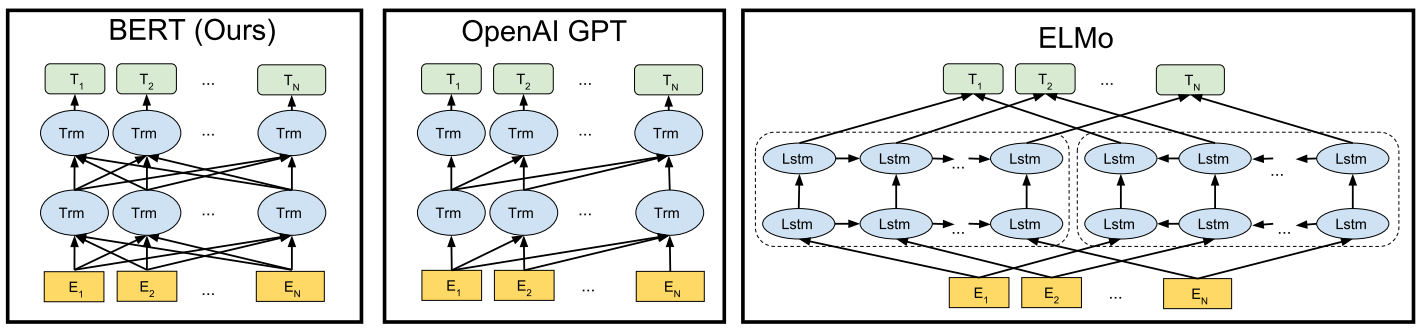}
\caption{Different pre-training techniques. BERT \cite{devlin2018bert} leverages a deep bidirectional Transformer. Open AI's GPT \cite{radford2018improving} exploits left-to-right Transformers. ELMO \cite{DBLP:conf/acl/PetersABP17,DBLP:conf/naacl/PetersNIGCLZ18} uses the concatenation of forward (left-to-right) and backward (right-to-left) LSTMs. As one can infer from the figures, solely BERT is jointly conditioned on the left and right context across layers. Figure copied from \cite{devlin2018bert}.}
\label{fig:bert_elmo_gpt}
\end{figure}

BERT refers to Bidirectional Encoder Representations from Transformers and as such is the first deeply bidirectional Language Model (LM) based on the Transformer architecture. Traditionally, the objective of pre-training LMs was to predict words given \textbf{either} the right or the left context of some window size (e.g., previous \textit{n}-gram) \cite{ratnaparkhi1997simple}. This is called left-to-right or right-to-left language modelling. Until the advent of BERT, all LMs based on the Transformer or Long Short-Term Memories (LSTMs) \cite{hochreiter1997long} were deployed either unidirectional \cite{radford2018improving} or shallowly bidirectional \cite{DBLP:conf/acl/PetersABP17,DBLP:conf/naacl/PetersNIGCLZ18}, and therefore not capable of contextualizing a word given the entire context the word appears in (i.e., right- and left-hand side of a token). BERT, however, has closed the gap and, as the name suggests, exploits the context both to the left and right of a word (see Figure~\ref{fig:bert_elmo_gpt} for a comparison between BERT and the aforementioned models with respect to their pre-training). BERT is thus the first unsupervised, deeply bidirectional LM for NLP that exclusively leverages fully connected linear layers and self-attention mechanisms that can easily relate tokens independent of their positions in an input sequence \cite{devlin2018bert}. This is particularly important for token-level tasks such as QA, where the context to both the left and right of an input token is decisive to find the correct answer span in a paragraph. Hence, BERT became indispensable in the disentanglement of a word's context on a variety of NLP tasks, as numerous recent studies have shown \cite{DBLP:journals/corr/abs-1909-03405,DBLP:journals/corr/abs-1909-04120,DBLP:journals/corr/abs-1909-02209}, and both the GLUE and SuperGLUE leaderboards indicate \cite{DBLP:journals/corr/abs-1804-07461,DBLP:conf/nips/WangPNSMHLB19}, where models that deploy BERT, or optimized versions of BERT (e.g., \cite{DBLP:journals/corr/abs-1907-11692,DBLP:conf/iclr/LanCGGSS20}), clearly outperform more traditional approaches. 

\begin{figure}[h!]
\centering
\begin{subfigure}{.41\textwidth}
    \centering
    \captionsetup{justification=centering}
    \includegraphics[width=.90\textwidth]{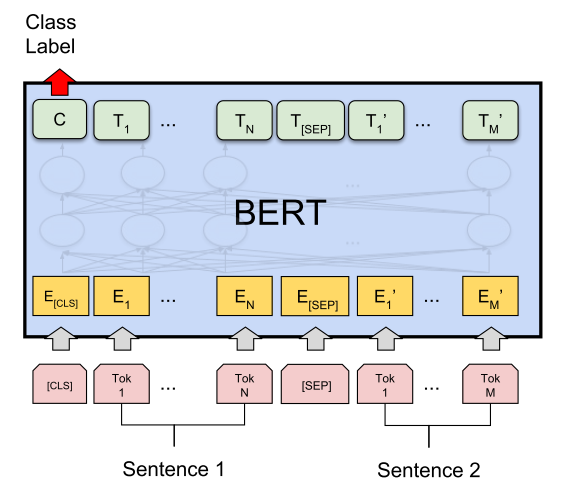}
    \caption{BERT fine-tuning for sentence pair classification.}
\end{subfigure}%
\begin{subfigure}{.44\textwidth}
    \centering
    \captionsetup{justification=centering}
    \includegraphics[width=.95\textwidth]{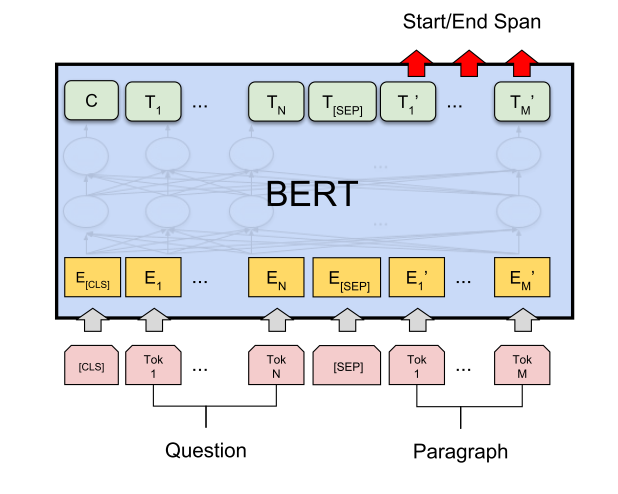}
    \caption{BERT fine-tuning for QA.}
\end{subfigure}
\caption{BERT fine-tuning \cite{devlin2018bert}.}
\label{fig:bert_finetuning}
\end{figure}

BERT's main pre-training objective is masked language modelling (MLM) \cite{devlin2018bert}. That is, some of the tokens of an input sequence are randomly masked, and the model is optimized to infer their vocabulary IDs based solely on their contexts. In contrast to standard left-to-right LM pre-training (e.g., \cite{radford2018improving}), BERT is optimized to jointly condition on both directions. As a result, fine-tuning for downstream application can easily be deployed, and requires nothing more than one additional task-specific output layer \cite{devlin2018bert} (see Figure~\ref{fig:bert_finetuning}). The inputs for such linear output layers are BERT's deeply bidirectional feature representations corresponding to an input token sequence, yielded through the pre-trained MLM objective. There is, however, the possibility to inform BERT about temporal dependencies through leveraging recurrence during fine-tuning, which has recently been explored with respect to token-level tasks \cite{hu2019question}. I will investigate further into this idea and scrutinize whether additional recurrent layers on-top of the pre-trained BERT model enhance performance concerning QA. This might fuse the best of both worlds for sequence modelling tasks: using a highly parallelizable and computationally efficient Transformer during pre-training, and exploiting recurrence during fine-tuning via (bidirectional) LSTMs. 

\newpage

\section{Recurrent Neural Networks}
\label{method:birnns}

I will now briefly explain the algorithm behind bidirectional RNNs as this is another neural architecture I exploit in the experiments. As their name suggests, bidirectional RNNs recursively process a word sequence $\mathbf{x}^{T}_i = [x^{1}_i, x^{2}_i,..., x^{T}_i]$ forward and backward in time \cite{birnns}. As such, both past-aware $\overrightarrow{\mathbf{h}_i} = [h^{1}_i, h^{2}_i,..., h^{T}_i]$ and future-aware $\overleftarrow{\mathbf{h}_i} = [h^{T}_i, h^{T\textbf{-}1}_i,..., h^{1}_i]$ hidden representations of an input sequence $\mathbf{x}^{T}_i$ are computed in every recurrent hidden layer $h_i$.

\begin{equation}
\overrightarrow{\mathbf{h}}_{i}^{(t)} =\mathrm{LSTM}\left(\overrightarrow{\mathbf{h}}_{t-1}, \mathbf{z}_{t}\right), t=1, \cdots,|x|
\end{equation}

\begin{equation}\overleftarrow{\mathbf{h}}_{i}^{(t)} =\mathrm{LSTM}\left(\overleftarrow{\mathbf{h}}_{t+1}, \mathbf{z}_{t}\right), t=|x|, \cdots, 1
\end{equation}

To compute the final hidden state $\mathbf{h}_i$, forward $\overrightarrow{\mathbf{h}_i}$ and backward $\overleftarrow{\mathbf{h}_i}$ hidden states are summed, before passing the sequence further to the next layer. The latter is done to keep the dimensionality of feature representations constant. Recall that $\mathbf{z} \in \mathbf{R}^{768}$. \footnote{For simplicity, for now $\theta(\mathbf{x}) = \mathbf{z}$.}
If we pass $\mathbf{z}$ through a BiLSTM and compute hidden representations both forward and backward in time of which both $\{\overrightarrow{\mathbf{h}}_{i}^{(t)}, \overleftarrow{\mathbf{h}}_{i}^{(t)}\} \in \mathbf{R}^{768}$, a concatenation would thus yield $\mathbf{H}_{i}^{(t)} \in \mathbf{R}^{768 \times 2}$ which we would like to avoid to both not overload the computational budget and keep the dimensionality of feature vectors in the same $\mathbf{R}$ space. Otherwise the comparison might not be equal as we would increase the dimensionality of feature representations twofold which could potentially result in more fine-grained word embeddings. Hence, $\mathbf{H}_{i}^{(t)}$ is computed as follows.

\begin{equation}
\mathbf{H}_{i}^{(t)} = \overrightarrow{\mathbf{h}}_{i}^{(t)} + \overleftarrow{\mathbf{h}}_{i}^{(t)}
\end{equation}

Since LSTMs in contrast to vanilla RNNs or GRUs \cite{chung2014empirical} consist of both hidden and cell states, the same computation as outlined above must be performed for cell states $\mathbf{c}_i$. LSTM denotes the LSTM function \cite{hochreiter1997long}, $\mathbf{h}_{i}^{(t)}$ is the hidden state at time step $t$, and $\mathbf{z}$ represents the contextual word embedding yielded by \textsc{bert}, that is $\theta(\mathbf{x}) \in \mathbf{R}^{768}$.

Before moving to the second \textsc{post-bert} encoding neural architecture, I would like to discuss a potential caveat of such recurrent encodings. Recall that the special [CLS] token in \textsc{BERT} reflects the semantic representation of an entire word sequence $\mathbf{x}_{i}$ (see Figure~\ref{fig:bert_finetuning}). Speaking with respect to temporal dependencies, the latter is the token at timestep $0$, or in other words the token at the $0^{th}$ index of the word vector in latent space. An RNN, however, recursively processes a sequence of words timestep by timestep, starting at $0$ and stopping at $T$, taking into account both each previous hidden state $h_{i}^{t-1}$ and each current input $z_{i}^{t}$. Hence, the semantic representation of a sentence is not encoded at the $0^{th}$ index as in \textsc{BERT} but at the last index, that is at timestep $T$. This does not come with any problems for a QA task, where one has to pass the entire hidden representation of an input sequence to the fully-connected QA output layer. It might even be beneficial, as one recent study has shown, due to the fact that temporal dependencies are additionally encoded after \textsc{BERT} \cite{hu2019question}. For simple classification tasks, however, this might indeed result in a problem. Usually, one is required to exclusively pass the hidden feature representation encoded in the [CLS] token to a classification output layer (see Figure~\ref{fig:bert_finetuning}). If one applies a recurrent neural module before employing the latter step, then the question of which latent representation must be used for the classification layer becomes non-trivial. By virtue of simplicity and according to common knowledge about BiLSTMs \cite{hochreiter1997long,birnns}, I will leverage the hidden representation at the last time step as the input vector for classification layers - when performing STL classification experiments or MTL.

\section{Highway Networks}
\label{method:highway}

Network depth comes with the cost of longer training cycles, the necessity of more training data, and the complexity of more sophisticated optimization and regulation techniques to exchange information between different layers \cite{glorot2010understanding,lecun2015deep}. Although powerful and highly promising with respect to machine learning tasks, such deep networks require careful training procedures. 
Highway networks were developed to facilitate this information flow through employing some form of regulation between layers in deep neural architectures \cite{DBLP:journals/corr/SrivastavaGS15}. In so doing, Highway networks employ gating units to regulate the flow through the network. This is deployed in a similar manner as the gating mechanisms in LSTMs \cite{hochreiter1997long}. Those gates enable paths along which information flows across layers, namely information highways. Hence, the name Highway networks. 

In general, a feedforward linear layer is deployed as follows,

\begin{equation}
\mathbf{y}=H\left(\mathbf{z}, \mathbf{W}_{\mathbf{H}} + \mathbf{b}_{\mathbf{H}}\right).
\end{equation}

This layer is called the projection layer and is denoted with the letter H. It employs an affine transformation on \textsc{bert}'s latent feature representation $\mathbf{z}$, followed by a rectified linear unit (ReLU) non-linearity which sets all negative values of an input vector or matrix to $0$\footnote{$f(z) \: = \: \max (0, z)$} to yield the projection layer's output $\mathbf{y}$. The weights of this layer $\mathbf{W}_{\mathbf{H}}$ are initialized according to the Xavier uniform initialization, also called Glorot initialization \cite{glorot2010understanding}, which generally yields better results than a simple random initialization of a layer's weights.
In a Highway network, two linear transforms are employed in addition to $H$, a transform gate $T\left(\mathbf{z}, \mathbf{W}_{\mathbf{H}}\right)$ and a carry gate $C\left(\mathbf{z}, \mathbf{W}_{\mathbf{H}}\right)$, thus resulting in the following equation,

\begin{equation}
\mathbf{y}=H\left(\mathbf{z}, \mathbf{W}_{\mathbf{H}} + \mathbf{b}_{\mathbf{H}}\right) \odot T\left(\mathbf{z}, \mathbf{W}_{\mathbf{T}} + \mathbf{b}_{\mathbf{T}}\right)+\mathbf{z} \odot C\left(\mathbf{z}, \mathbf{W}_{\mathbf{C}} +\mathbf{b}_{\mathbf{C}}\right).
\end{equation}

In contrast to the projection layer $H$, which is followed by a ReLU non-linearity, the transform gates $T$ and $C$ are both followed by a sigmoid function, that is

\begin{equation}
\sigma(\mathbf{z})=\frac{1}{1+e^{-\mathbf{z}}}, \mathbf{z} \in \mathbf{R},
\end{equation}

where 

\begin{equation}
    \mathbf{z} \in \{\left(\mathbf{z}, \mathbf{W}_{\mathbf{T}} +\mathbf{b}_{\mathbf{T}}\right); \left(\mathbf{z}, \mathbf{W}_{\mathbf{C}} +\mathbf{b}_{\mathbf{C}}\right)\},
\end{equation}

instead of

\begin{equation}
\mathbf{ReLU}\left(\mathbf{z}, \mathbf{W}_{\mathbf{H}} + \mathbf{b}_{\mathbf{H}}\right)=\max (0, \left(\mathbf{z}, \mathbf{W}_{\mathbf{H}} + \mathbf{b}_{\mathbf{H}}\right)).
\end{equation}

According to the original paper \cite{DBLP:journals/corr/SrivastavaGS15}, $C$ is set to $C = 1 - T$ and therefore also employed in the current set-up. Hence,

\begin{equation}
\mathbf{y}=H\left(\mathbf{z}, \mathbf{W}_{\mathbf{H}} + \mathbf{b}_{\mathbf{H}}\right) \odot T\left(\mathbf{z}, \mathbf{W}_{\mathbf{T}} + \mathbf{b}_{\mathbf{T}}\right)+\mathbf{z} \odot \left(1 - T\left(\mathbf{z}, \mathbf{W}_{\mathbf{T}} + \mathbf{b}_{\mathbf{T}}\right)\right).
\end{equation}

One can quickly see that this is nothing other a sum between an element-wise multiplication (i.e., Hadamard product) between two feedforward neural networks and an element-wise multiplication (i.e, Hadamard product) between an input sequence $\mathbf{z}$ and another, differently parametrized feedforward neural network. The fact that this transformation not only takes into account the linear transformations yielded by the different gates but also the input $\mathbf{z}$, similar to a residual block, makes a Highway network more flexible and thus better suited for networks with more depth than a simple linear transform \cite{DBLP:journals/corr/SrivastavaGS15}. Furthermore, a Highway network has thus the ability to adaptively transform or copy feature representations which might be beneficial as a bridging step between \textsc{BERT} and an output layer for the final classification. The dimensionality for both inputs, outputs, layers, and gates must be the same in a Highway block consisting of fully connected layers, which is what I employ. Hence, $\{\mathbf{z}, \mathbf{y}, H(\mathbf{z}, \mathbf{W_{H}} + \mathbf{b_{H}}), T(\mathbf{z}, \mathbf{Wf_{T}} + \mathbf{b_{T}})\} \: \in \: \mathbf{R}^{768}$.

\section{Multi-task Learning} 
\label{section:multitask}

In multi-task learning (MTL) a learner is required to perform several tasks in parallel. Hence the name MTL. This is in stark contrast to single task learning (STL), where a model is optimized to perform well on a single task only. In MTL, one of the tasks usually serves as the main task, which a learner is evaluated on at inference time, and the remaining tasks serve as auxiliary tasks to provide useful information that enrich a model's feature representations to enhance performance on the main task \cite{caruana1997multitask,PhDThesisBjerva}. In general, the learner, e.g., a neural network, is trained on a training set $D_{t}$, that is exploited across all tasks in the task set $\mathbf{T}$. Hence, a main task $T$ and an auxiliary task $T'$ are drawn from the same training set $D_{t}$ but leverage different signals to perform well on the respective tasks. 

As such, MTL may be considered an inductive transfer method that exploits the domain specific information in training signals across different tasks and therefore introduces noise that help a model to generalize better with respect to unseen data \cite{caruana1997multitask}. MTL has recently received ample attention in Deep Learning (DL) in general and NLP in particular owing to its success across a variety of machine learning tasks \cite{DBLP:journals/corr/Ruder17a,DBLP:journals/corr/abs-1901-11504}. However, the benefits of MTL unfold if and only if the tasks in the task set $\mathbf{T}$ are related \cite{DBLP:conf/nodalida/Bjerva17,DBLP:conf/eacl/PlankA17,DBLP:conf/eacl/SogaardB17}. That means, that a task $T$ and another task $T'$ must have a somewhat similar training objective, where the signals from $T'$ are beneficial to enhance a model's performance concerning $T$. If the latter is not guaranteed, a model will learn feature representations that are not useful for any of the tasks in $\mathbf{T}$ \cite{caruana1997multitask,DBLP:conf/nodalida/Bjerva17,DBLP:conf/eacl/SogaardB17}. What makes the latter set-up particularly computationally efficient is the fact that MTL is exclusively deployed during training. At inference time, the model is solely evaluated on the main task $T$ as this is the model's primary objective.

In MTL for neural networks, all feature representations $F$ in a model's set of hidden layers $\mathbf{H}$ are shared across tasks, whereas the model contains task-specific parameters for each task in $\mathbf{T}$ that are not shared between tasks and are part of the output layer corresponding to each task. This is called \textsc{hard parameter} sharing.  \textsc{hard parameter} sharing has proven to be more successful than \textsc{soft parameter} sharing for a variety of machine learning tasks, and is therefore considered common practice in MTL in general \cite{caruana1997multitask} and for NLP in particular \cite{PhDThesisBjerva,DBLP:journals/corr/Ruder17a,DBLP:journals/corr/abs-1901-11504}. Moreover, \textsc{hard parameter} sharing is computationally more efficient than \textsc{soft parameter} sharing, where each task in $\mathbf{T}$ has its own model and corresponding hidden feature representations $F$. The number of models is equal to the number of tasks, and thus the number of parameters in the shared parameter set $\theta$ is as many times higher than in \textsc{hard parameter} sharing as there are tasks in $\mathbf{T}$ and as such requires notably more computational budget. What is more, \textsc{hard parameter} sharing is easier to implement and vastly reduces the risk of overfitting on the main task $T$ as a model has to share its feature representations across tasks \cite{DBLP:conf/eacl/SogaardB17}. This might not hold for \textsc{soft parameter} sharing owing to the fact that the feature representations $F$ are shared across models but do not necessarily force a single model to produce feature representations that are useful for all tasks in $\mathbf{T}$. \textsc{soft parameter} sharing is employed during sequential knowledge transfer, where a model is sequentially optimized until convergence with respect to each task in $\mathbf{T}$ \cite{caruana1997multitask}. This is opposed to parallel transfer, where a learner is simultaneously trained on all tasks. I will, due to the aforementioned reasons, primarily draw attention to \textsc{hard parameter} in the experiments but also employ \textsc{soft parameter} sharing to contrast MTL techniques against each other.

%% file: method.tex
\section{Model}

\begin{figure}[h!]
    \centering
    \includegraphics[width=1.05\textwidth]{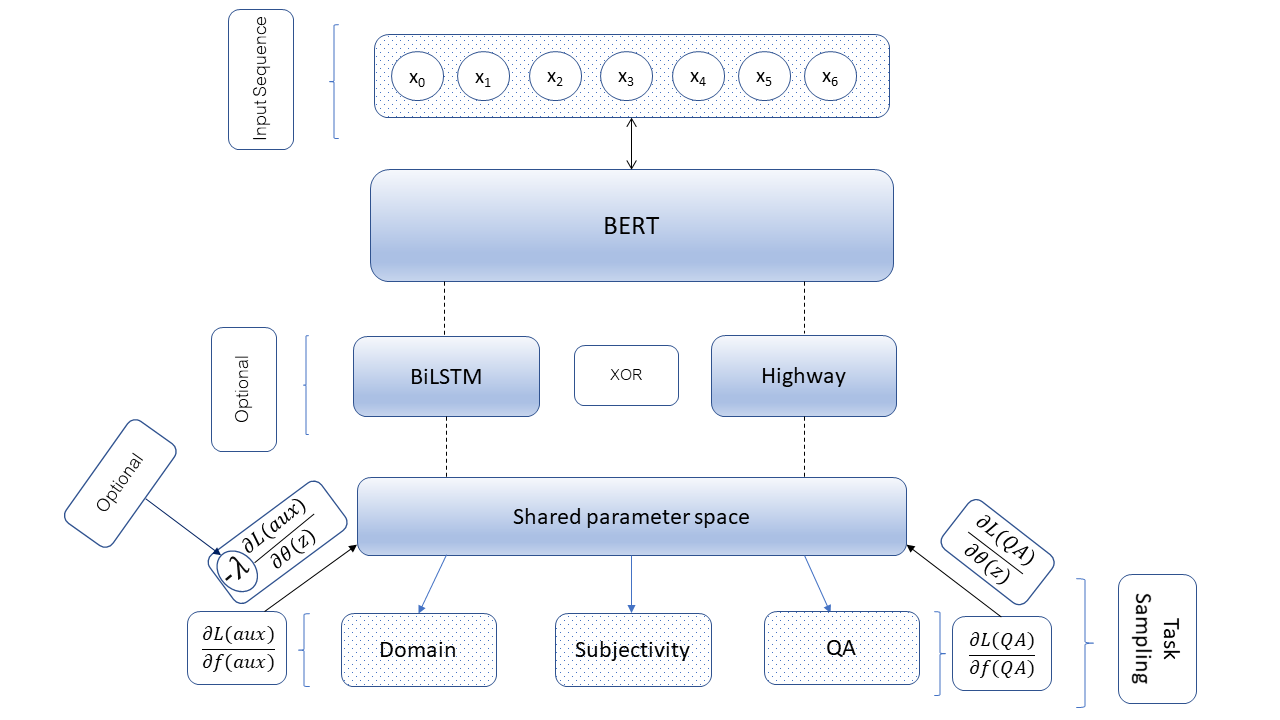}
    \caption{Multi-task learning for QA. Additional \textsc{post-bert} encoding layers are optional. Feature representations are shared across all tasks (i.e., \textsc{hard parameter} sharing). Each task in the task set $\mathbf{T}$ has its own task-specific output layer. The backprogated gradients w.r.t. the auxiliary tasks may be reversed. The latter depends on the employed MTL set-up. For simplicity, the input sequence $x_{i}^{T}$ is depicted in $\mathbf{R}^{7}$.}
    \label{fig:mtl_model}
\end{figure}

Figure~\ref{fig:mtl_model} illustrates the deployed MTL architecture, and serves as an introductory high-level overview. I will now go through each part of the model step-by-step from bottom to top, starting with extracting contextual features through BERT \cite{devlin2018bert}, a neural architecture based on the Transformer (see Section~\ref{section:transformers}). 

\subsection{BERT}
For every implemented QA model, a pre-trained \textsc{distilbert} Transformer \cite{sanh2019distilbert} serves as the feature extractor prior to any task-specific output or \textsc{post-bert} custom encoding layers \footnote{\url{https://huggingface.co/transformers/}}. Compared to BERT \cite{devlin2018bert}, of which the Base and Large model consist of 12 and 24 Transformer layers respectively, \textsc{distilbert} only contains 6 Transformer layers without showing a statistically significant deterioration in performance compared to BERT Base on a variety of NLP downstream tasks \cite{DBLP:journals/corr/abs-1804-07461,sanh2019distilbert}. This makes \textsc{distilbert} highly user-friendly and easy to deploy.

Updating the weights of a full BERT Transformer model is not feasible with the available computational budget. Hence, I am required to leverage a distilled version of BERT. For simplicity and to facilitate reading, I use the name BERT throughout the following sections when referring to a \textsc{distilbert} model.

\subsection{Notations}
 Table~\ref{tab:notations} depicts the notations that will be used throughout the following sections.

\begin{table*}[ht]
\centering
\begin{tabular} {@{}r|r@{}}
\toprule
Math notation & Natural Language reference \\
\midrule
$\mathbf{(q, c)}$ & question - context pair \\
$\mathbf{(q, a)}$ & question - answer pair \\
$\mathbf{x}$ & input sequence \\
$\tilde \mathbf{f}$ & a model \\
$\theta (\mathbf{x})$ & feature extraction through \textsc{bert} \\
$\phi$ & additional encoding/fine-tuning layer(s) \\
$f_{k}$ & task-specific output layer \\
%$k$ & number of auxiliary modules in MTL \\
$f_{qa}$ & QA head  \\
$f_{sbj}$ & subjectivity head \\
$f_{dom}$ &  context-domain head  \\
\bottomrule
\end{tabular}
\caption{Notations that will be used throughout the following section(s).}
\label{tab:notations}
\end{table*}

\subsection{Multi-task Learning} 
On top of the pre-trained \textsc{BERT} language model (LM) which I refer to as $\theta (\mathbf{x})$ throughout the following paragraph, I build my own task-specific output layers. Each model $\tilde \mathbf{f}$ consists of a fully-connected feed-forward linear output layer which I refer to as $f_{qa}(\theta (\mathbf{x}))$ to predict the answer span $\mathbf{a}$ within the context $\mathbf{c}$ given an input question-context pair $(\mathbf{q, c}) = \mathbf{x}$ . In addition to QA, which $\forall \tilde \mathbf{f}$ is implemented as the main task, neural networks are augmented with $\mathbf{k}$ auxiliary task modules, where either $\mathbf{k} = 1$ or $\mathbf{k} = 2$. I refer to the task-specific output layer corresponding to the classification of both a question and its respective context $(\mathbf{q, c})$ into a subjective opinion vs. an objective, measurable fact as the classifier $f_{sbj}(\theta (\mathbf{x}))$. Hence, the first auxiliary task is defined as a binary sequence classification task with respect to both the question $\mathbf{q}$ and its corresponding answer $\mathbf{a}$. 

On the other hand, the second auxiliary task $\forall \tilde \mathbf{f}$ with $\mathbf{k} = 2$ auxiliary modules is defined as a multi-way classification of the \textsc{context-domain} $\mathbf{c}^{d}$, where the number of classes is $6$ or $7$ respectively, dependent on whether $(\mathbf{x}, \mathbf{y}^{d}) \in D_{Subj}$ or $(\mathbf{x}, \mathbf{y}^{d}) \in D_{Subj} \cup D_{SQuAD}$\footnote{$D_{Subj}$ and $D_{SQuAD}$ refer to question-review pairs from \textsc{SubjQA} or \textsc{SQuAD} respectively.}. A domain, $\mathbf{y}^{d}$,  is part of the following set, $\{\textsc{books}, \textsc{electronics}, \textsc{grocery}, \textsc{movies}, \textsc{restaurants}, \textsc{tripadvisor}, \textsc{wikipedia}\}$. The latter domain, namely \textsc{Wikipedia}, is part of the class set, if and only if $(\mathbf{x}, \mathbf{y}^d) \in D_{Subj} \cup D_{SQuAD}$. I refer to task-specific output layers for the latter auxiliary task as $f_{dom}(\theta (\mathbf{x}))$.

There is, however, the possibility that the model hierarchy does not evaluate to the aforementioned $f_k(\theta (\mathbf{x}))$ structure but must rather be depicted as $f_k(\phi(\theta (\mathbf{x})))$, where $\phi$ summarizes the parameter set of custom NN encoders on top of $\theta (\mathbf{x})$. Experiments are run both in a setting where $\phi$ is a Recurrent Neural Network (RNN), and in a setting where it is a Highway layer. In every set-up, $\phi$ is placed between $\theta (\mathbf{x})$ and any task-specific layer $f_k$. If a custom feature encoder $\phi$ is implemented in-between $\theta (\mathbf{x})$ and $f_k$, and MTL is performed, then the parameters of $\phi$ are shared among all $f_k(\phi(\theta (\mathbf{x})))$, where only $f_k$ is task-specific, and therefore not shared among the full model parameter set. This is called \textsc{hard parameter} sharing which has proven to be more successful than \textsc{soft parameter} sharing for a variety of NLP tasks \cite{DBLP:journals/corr/Ruder17a} and is therefore considered common practice in MTL in general \cite{caruana1997multitask} and for NLP in particular \cite{PhDThesisBjerva,DBLP:conf/eacl/PlankA17,DBLP:conf/eacl/SogaardB17}. 

Due to the fact that all three tasks are classification tasks by nature, relatedness is given on a higher, more abstract level of machine learning tasks. Although this does not guarantee relatedness with respect to MTL in particular \cite{DBLP:conf/eacl/SogaardB17}, it is indispensable to stress that tasks should resemble each other not only on a lower, more task-specific but also on a higher, more task-nature related level \cite{caruana1997multitask}. Before I explain in more detail how $\phi$ is implemented, I would like to stress the importance of different task sampling strategies and give a short overview of the strategies employed in the experiments.

\subsection{Task Sampling}
\label{method:task_sampling} 
To decipher whether different task sampling regimes in MTL impact model performance on the main task differently, I compare two task sampling strategies against each other in each of the implemented MTL settings. In a \textsc{uniform sampling} setting, the main task, that is QA, and the auxiliary task(s), that is subjectivity or context-domain classification, are equally often sampled during a single training epoch. Hence, in an MTL setting with one auxiliary task, each of the tasks is optimized in $50 \%$ of all training steps, and in a setting with two auxiliary tasks, a model $\tilde \mathbf{f}$ is fine-tuned on each of the tasks $\frac{1}{3}$ of the time. 

In an \textsc{oversampling} setting, however, the main task, that is QA, is sampled $\frac{2}{3}$ per training epoch, and the remaining $\frac{1}{3}$ of training steps are equally distributed among fine-tuning on the auxiliary task(s). Thus, in an MTL setting with two auxiliary tasks, each of the tasks is sampled $\frac{1}{6}$ per epoch. Contrary to the former setting, this follows a skewed sampling distribution of machine learning tasks during training, where the main task is oversampled and the auxiliary tasks are optimized equally often. The latter strategy is employed to oversample the main task and examine potential differences in performance compared to uniformly sampling all tasks. 

\subsection{Modelling Subjectivity}
\label{method:subjectivity}
Before moving to the explanation of implementation details with respect to $\phi$, I will briefly discuss $f_{sbj}(\theta (\mathbf{x}))$ more thoroughly. The vast majority of character sequences, namely strings, in a review paragraph $\mathbf{r}$ or more general, a context $\mathbf{c}$, consists of tokens that are not reflecting subjective opinions. Thus, it might be easier for a model $\tilde \mathbf{f}$ to solely classify the answer $\mathbf{a}$, which is a sub-string of $\mathbf{c}$, into subjective opinions vs. objective, measurable facts, instead of learning to predict whether an entire review reflects subjectivity. 

Therefore, I will implement two different versions of $f_{sbj}(\theta (\mathbf{x}))$. In the first setting, which can be considered the standard sequence-pair classification setting for questions and contexts, $f_{sbj}(\theta (\mathbf{x}))$ is optimized to predict whether both the question $\mathbf{q}$ and its corresponding context $\mathbf{c}$ belong to the class of subjective opinions or measurable facts. In the second setting, which can be considered an exploratory modelling attempt, $f_{sbj}(\theta (\mathbf{x}))$ is trained to classify the answer $\mathbf{a}$ instead of the context $\mathbf{c}$ into subjective vs. objective. Here, one has to alternate between batches, where \textsc{batch}$_1$ consists of mini-batches of $(\mathbf{q, c})$ sequence pairs and \textsc{batch}$_2$ contains mini-batches of $(\mathbf{q, a})$ sequence pairs. The latter setting is called \textsc{batch alternation} as one is required to exploit sequences from \textsc{batch}$_2$ as input to $f_{sbj}(\theta (\mathbf{x}))$ and leverage sequences from \textsc{batch}$_1$ as inputs to $f_{qa}(\theta (\mathbf{x}))$ $\lor$ $f_{dom}(\theta (\mathbf{x}))$, of which both are generated for each training iteration.

When fine-tuning a model $\tilde \mathbf{f}$ on \textsc{subjectivity} classification only, this is employed during both training and test time. For QA-MTL, however, the latter is deployed exclusively during training time. At inference time, where the model does not perform any auxiliary task, the model has to find an answer span $\mathbf{a}$ in a context $\mathbf{c}$ to respond to the corresponding question $\mathbf{q}$. Thus, each input sequence $\mathbf{x} = (\mathbf{q, c})$.

\subsection{Recurrent Neural Networks}

Neural architectures based on the Transformer \cite{DBLP:conf/nips/VaswaniSPUJGKP17,rush-2018-annotated} such as \textsc{BERT} do not take into account temporal dependencies between tokens in a sequence of tokens, $x_{i}^{t} \in \mathbf{x}_{i}^{T}$. Hence, I equip a model $\tilde \mathbf{f}$ with the possibility to exploit an RNN based neural module, namely Long-Short-Term-Memory network \cite{hochreiter1997long}, on top of \textsc{BERT} prior to any task-specific linear output layer. A few recent studies have shown that further encoding the contextual feature representations yielded by \textsc{BERT} through LSTMs before performing the QA task enhances the learner's performance (e.g., \cite{hu2019question}). Since \textsc{BERT} is the first deep neural language model (LM) based on the Transformer \cite{devlin2018bert} which exploits bidirectional self-attention mechanisms, temporal dependencies must be computed forward and backward in time to retain the contextual features specific to \textsc{BERT}. See Section~\ref{method:birnns} for mathematical details of RNNs. 

 I refer to any RNN based \textsc{post-bert} encoding layer as $\phi$, which is applied to $\theta(\mathbf{x})$, $\Rightarrow \phi(\theta(x))$. I leverage PyTorch's LSTM implementation \cite{PyTorch}. For each recurrent \textsc{post-bert} encoder, 2 LSTM layers are deployed, the bidirectional flag of the LSTM class is set to true to employ a BiLSTM as discussed in Section~\ref{method:birnns}. Additionally, a dropout rate of $.25$ is applied to each layer. I call this an LSTM block. If a model is equipped with a \textsc{post-bert} recurrent encoding layer, then it consists of one and only one LSTM block. 

\subsection{Highway Networks}

Another way of further encoding \textsc{BERT}'s feature representations is to pass a word sequence's latent representation $\mathbf{z}_{i}$ through a Highway network \cite{DBLP:journals/corr/SrivastavaGS15}. Again, the same study as mentioned in the previous section \cite{hu2019question} showed that a Highway layer in-between \textsc{BERT}, that is $\phi(\theta(x))$ or $\phi(z)$, and a task-specific output layer, that is $f_{k}$, supports the information flow between \textsc{BERT} and the linear QA output layer, thus enhancing model performance. See Section~\ref{method:highway} for a thorough introduction about and mathematical details of Highway networks. 

I refer to any \textsc{post-bert} encoding layer that leverages a Highway block as $\phi$, which is applied to $\theta(\mathbf{x})$, $\Rightarrow \phi(\theta(x))$. As such, $\phi$ may be employed as an RNN module or a Highway block, depending on the set-up. The potential caveats for a recurrent \textsc{post-bert} encoding mentioned in Section~\ref{method:birnns} do not hold for Highway blocks since a Highway network, similar to \textsc{BERT}, does not leverage recurrence, thus making this set-up potentially better suited for STL classification tasks. If a model is equipped with a \textsc{post-bert} Highway layer, then it consists of one and only one Highway block. This is done to not overload the computational budget and keep the number of parameters in a model $\tilde \mathbf{f}$ similar across set-ups. 

\section{Fine-tuning}
\label{method:finetuning}

Each pre-trained model $\tilde \mathbf{f}$ is fine-tuned either on $D_s$, $D_o$ or $D_c$. $D_s$ refers to the dataset that only consists of question-review pairs from \textsc{SubjQA}, $D_o$ denotes the dataset that exclusively contains question-context pairs from \textsc{SQuAD}, and $D_c = D_s \cup D_o$. $\forall \tilde D \in \{D_s, D_o, D_c\}$ fine-tuning is performed for a predefined number of $T = 3$ epochs. This follows the fine-tuning regime as recommended in the original BERT paper \cite{devlin2018bert}. During each epoch $t$, a model $\tilde \mathbf{f}$ updates its weights for $\frac{N^{t}}{b}$  steps, where $N^{t}$ refers to the number of examples in a given train set $X^{t}$ and $b$ denotes the batch size. The latter is set to $16$ for all training procedures, which alongside batch sizes of $32$ and $64$ is considered standard practice \cite{devlin2018bert,sanh2019distilbert}. I chose $b = 16$ since in initial experiments mini-batches of size $b = 32$ did not fit into GPU memory for a Titan X with 12 GB given my training procedure and model set-up.

\subsection{Evaluation}
To inspect whether a model $\tilde \mathbf{f}$, that is exposed to the training set $X^{t}$, generalizes well to unseen validation data $X^{v}$, and does not overfit to $X^{t}$, a researcher is required to evaluate $\tilde \mathbf{f}$ on $X^{v}$ a predefined number of $k$ times during training. The most common set-up for performing this step is to either test $\tilde \mathbf{f}$ exactly once after the entire training ($k = 1$) or after an epoch $t$ ($k = T$) \cite{devlin2018bert,DBLP:journals/corr/abs-1811-01088}. However, one recent study has examined this evaluation regime in more detail \cite{dodge2020finetuning}. The authors showed that evaluating $\tilde \mathbf{f}$ on $X^{v}$ a predefined number of $10$ times during an epoch ($k = T \times 10$) leads to significantly better performance than exploiting either of the aforementioned standard evaluation set-ups. Thus, I implement both a set-up where $\tilde \mathbf{f}$ is evaluated after a training epoch $t$ ($k = T$), and an alternative version where $\tilde \mathbf{f}$ is tested $10$ times on $X^{v}$ during an epoch $t$ ($k = T \times 10$). 

For various reasons, I do not implement the third evaluation set-up, where $\tilde \mathbf{f}$ is evaluated once on $X^{v}$ after the entire training procedure ($k = 1$). Firstly, this is less common practice in machine learning research than evaluating $\tilde \mathbf{f}$ after each epoch $t$. Secondly, it seems less reasonable to implement a $k = 1$ regime, if one endeavours to stop the training procedure early when $\tilde \mathbf{f}$ either decreases or plateaus with respect to its performance. This performance, however, cannot be measured on $X^{t}$ since it is not a valid indicator for generalization. The model is trained on $X^{t}$ and hence has already seen all examples from $X^{t}$ as many items as was iterated over $X^{t}$. A researcher must therefore investigate whether performance drops or reaches a plateau with respect to $X^{v}$ to correctly implement an early stopping regime. One can see that the latter cannot be done, if $k = 1$ - at least not in the standard way of early stopping the training.

\subsection{Early Stopping} 
I implement early stopping for two reasons. Firstly, it is computationally inefficient to train a model $\tilde \mathbf{f}$ a total number of $T$ epochs on $X^{t}$, if model performance on $X^{v}$ does not increase or even plummets towards the end of training. To save time and computational budget, one could simply terminate the training when this happens. Secondly, one wants to save the weights of $\tilde \mathbf{f}$ at its peak performance on $X^{v}$ during training, and not after a performance drop. Early stopping during training is performed, whenever the cross-entropy loss with respect to QA evaluated on the entire validation set $X^{v} = \{(x_j^{v}, y_j^{v})\}^{n_s}_j$ does not increase for $k = 5$ evaluation steps, if $k = T \times 10$, or is higher than the loss at the $(k\:-\:1)^{th}$ evaluation step, if $k = T$, since $T = 3$ for all fine-tuning regimes, and hence $(k-2)^{th}$ at $k = 3$ for $k = T$ is validation performance after epoch 1, which we do not want to compare against. 

The latter early stopping regimes are executed after a model $\tilde \mathbf{f}$ is fine-tuned for a single training epoch since I do not want to stop training during the first epoch. The weights of a model $\tilde \mathbf{f}$ are saved only when the validation loss at the $k^{th}$ evaluation step is lower than the previous minimum loss, and not stored when an increase in validation loss is observed. 

\subsection{Optimization} 
Each model $\tilde \mathbf{f}$ is optimized through Adam \cite{kingma2014adam} with weight decay fix as recommended in \cite{devlin2018bert}. The learning rate $\eta$ is set to $5e-5$ $\forall \tilde \mathbf{f}$ as recommended for fine-tuning BERT on QA. Furthermore, a linear scheduler with a warm-up period is applied to the optimizer to control $\eta $ during training \cite{DBLP:journals/corr/abs-1801-06146}. This works as follows: $\eta$ linearly increases for a predefined number of steps, called \textsc{warm-up steps}, and after the warm-up period decreases linearly until model convergence or stopping of the training procedure. 

For the simplest training set-up, that is a single task learning (STL) setting where a model $\tilde \mathbf{f}$ performs the main task (i.e., QA) only, the empirical risk is minimized per iteration through mini-batch gradient descent optimization as follows,

\begin{equation}
\min _{\Theta} \frac{1}{2}\left(\frac{1}{n} \sum_{i=1}^{n} J\left( f_{qa}\left(\phi\left(\theta\left(\mathbf{x}_{i}\right)\right)\right)^{s}, y_{i}^{s}\right) + \frac{1}{n} \sum_{i=1}^{n} J\left( f_{qa}\left(\phi\left(\theta\left(\mathbf{x}_{i}\right)\right)\right)^{e}, y_{i}^{e}\right)\right)\footnote{The parameter set $\phi$ of the custom encoder structure is optional, but for simplicity depicted in the equation. This holds for other equations too.},
\end{equation}

where $f_{qa}(\phi(\theta(x_{(i)})))$ is the conditional probability that the QA model knows that the question-context sequences $x_{(i)}$ within a mini-batch of $n$ sequences correspond to the the correct start and end positions $y_{(i)}^{s}$ and $y_{(i)}^{e}$ of the respective answer spans $y_{(i)}^{s-e}$, and $J(\theta)$ is the cross-entropy loss function with respect to all model parameters $\theta$ which is computed for both $y_{(i)}^{s}$ and $y_{(i)}^{e}$ as follows,

\begin{equation}
J(\Theta) = -\sum^{n}_{y_i} 1(X, y_i) \log (P_{r}(y_i | f_{qa}(\phi(\theta(x_i))))),
\label{equation:cross_entropy}
\end{equation}

where $1(x_i, y_i)$ is the binary indicator function (0 or 1) if the start or end position respectively is correct for the question-context sequence $x_i$, and $P_r(y_i | f_{qa}(\phi(\theta(x_i)))$ is a given discrete probability distribution over all possible start and end positions respectively. The latter is computed by the \textsc{softmax} function as follows:

\begin{equation}
\sigma(\mathbf{z})_{i}=\frac{e^{z_{k}}}{\sum_{i=1}^{K} e^{z_{j}}},
\label{equation:softmax}
\end{equation}

where $(\mathbf{z})_i$ evaluates to $f_{qa}(\phi(\theta(x_i)))$ and denotes the non-normalized output of an NN model $\tilde \mathbf{f}$, in the literature referred to as \textsc{logits}. Logits are $k$-dimensional vectors, where $k$ denotes the possible number of classes (i.e., start and end positions respectively). For QA, this is an output matrix of size $k \times 2$. Hence, $\mathbf{z}_i \in \mathbf{R}^{k \times 2}$. Following the standard practice we compute start and end logits, however, separately and split the resulting raw output $\mathbf{z}_i \in \mathbf{R}^{k \times 2}$ into $z_i^{s} \in \mathbf{R}^{k}$ and $z_i^{e} \in \mathbf{R}^{k}$ respectively to compute the \textsc{softmax} over $z_i^{s}$ and $z_i^{e}$ to yield probability distributions $p_i^{s}$ and $p_i^{e}$. Since the \textsc{softmax} function is nothing other than a multinomial logistic regression over $k$ classes, we can simply write $\sigma$ to denote \textsc{softmax}. Thus, we must compute $\sigma(z^{s})_i$ and $\sigma(z^{s})_i$ separately. 

\section{Multi-task Learning}
In MTL, we minimize the following empirical risk for multi-way classification of \textsc{context-domains} $d$ or binary classification of \textsc{subjectivity} labels of questions and answers respectively,

\begin{equation}
\min _{\Theta} \frac{1}{N} \sum_{i=1}^{N} J\left( f_{aux}\left(\phi\left(\theta\left(\mathbf{x}_{i}\right)\right)\right)^{aux}, y_{i}^{aux}\right),
\label{equation:auxiliary_loss}
\end{equation}

where $J$ denotes the categorical cross-entropy as described in Equation~\ref{equation:cross_entropy} for multi-way classification and the binary cross-entropy for binary classification, $N$ refers to the total number of mini-batches and $f_{qa}(\phi(\theta(x_{(i)})))^{aux}$ is the conditional probability that the model classified the input sequence $x_{(i)}$ in a mini-batch of $n$ examples, where $n = 16$, as the corresponding true domains $y_{(i)}^{d}$ or the correct \textsc{subjectivity} labels $y_{(i)}^{sbj}$. Whenever label imbalance is observed, loss weighting is performed. That is, the empirical risk for classes that appear less frequently in the training data is weighted higher. To provide an example, imagine that positive examples account for $100$ of all training examples $N$, where $N = 900$, in a binary classification problem with a single class. Then, the loss for positive examples is multiplied by a factor of $\frac{800}{100} = 8$ such that the loss acts as if the dataset contains equally many positive and negative examples. Similarly, to account for label imbalance in the multi-class problem cross-entropy loss weights corresponding to each class are computed as follows,

\begin{equation}
    w^{k} = 1 - \frac{n^{k}}{N}, k=1, \cdots, \mathbf{k}
\label{equation:class_weights}
\end{equation}

Hence, cross-entropy loss for each class is weighted according to the class weights obtained from Computation~\ref{equation:class_weights}. For multi-way classification, the \textsc{softmax} function as depicted in Equation~\ref{equation:softmax} is computed over the model's raw output logits to yield a discrete probability distribution over all possible domains $k$. In binary classification, however, the \textsc{sigmoid} function is computed over the model's raw output logits to obtain a single scalar value between $0-1$ (i.e., probability value), according to the following formula,

\begin{equation}
\sigma(\mathbf{z})_{i}=\frac{e^{z_{i}}}{e^{z_{i}}+1}
\label{equation:sigmoid}
\end{equation}

where $(\mathbf{z})_{i}$ evaluates to $f_{sbj}(\phi(\theta(x_{i})))$ and denotes the model's \textsc{subjectivity} label prediction.

\section{Adversarial Training in MTL}
\label{method:adversarial}

To inspect whether models benefit from learning domain invariant features, which has proven to be useful in various machine learning problems where labels from the target domain were limited \cite{chopra2013dlid,glorot2011domain,ganin2014unsupervised,long2015learning,DBLP:journals/corr/MinSH17}, I implement two different adversarial training settings in MTL. In the first adversarial training setting, which I will refer to as \textsc{adversarial simple} throughout the following section, the loss is simply reversed after each comparison between the model's raw output logits  $f_{qa}(\phi(\theta(x_{(i)})))$ and the true labels $y_{(i)}^{aux}$.

In the second adversarial training setting, which I refer to as \textsc{adversarial grl} throughout the following section, a gradient reversal layer (GRL) following \cite{ganin2014unsupervised}, is placed in-between the shared feature encoding layers $\phi(\theta(\mathbf{x}))$ and the auxiliary task-specific output layers $f_{aux}$.

\subsection{Reversing Losses}
In an \textsc{adversarial simple} training setting, the sign of the loss is simply reversed to make the model not learn the auxiliary task(s) at all, whereas all other optimization parameters stay as in a \textsc{normal} MTL setting. Hence, goal of the optimization procedure is to maximize the loss for auxiliary task(s):

\begin{equation}
\max _{\Theta} \frac{1}{N} \sum_{i=1}^{N} J\left( f_{aux}\left(\phi\left(\theta\left(\mathbf{x}_{i}\right)\right)\right)^{aux}, y_{i}^{aux}\right),
\label{equation:adversarial_loss}
\end{equation}

where $J$ denotes the categorical or binary cross-entropy as described in Equation~\ref{equation:cross_entropy}, $N$ refers to the total number of mini-batches and $f_{aux}(\phi(\theta(x_{(i)})))$ is the conditional probability that the model classified an input sequence $x_{(i)}$ in a mini-batch of $n$ examples, where $n = 16$, according to the goal of the particular auxiliary task. 

\subsection{Reversing Gradients}
\begin{figure}[h!]
    \centering
    \captionsetup{justification=centering}
    \includegraphics[width=.68\textwidth]{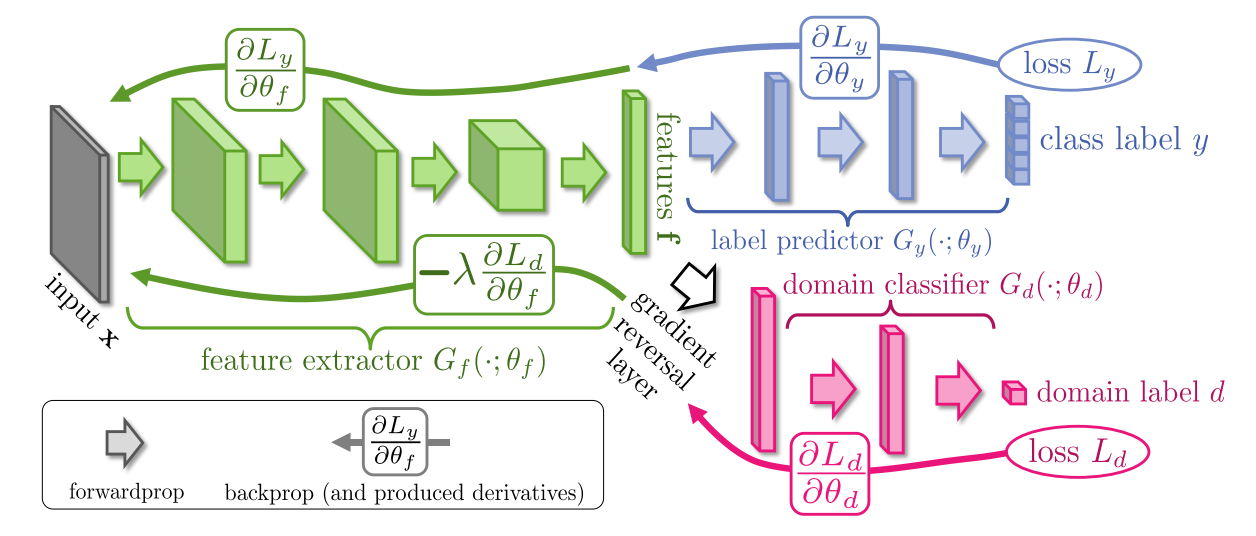}
    \caption{GRL architecture \cite{ganin2014unsupervised}.}
    \label{fig:grl_architecture}
\end{figure}

Following Ganin et al., 2014 \cite{ganin2014unsupervised}, a Gradient Reversal Layer (GRL) is placed in-between the shared feature encoding layers $\phi(\theta(\mathbf{x}))$ and the auxiliary task-specific output layers $f_{aux}$. The primary goal of \textsc{adversarial grl} is to produce domain-invariant feature representations while at the same time making the model learn the auxiliary task(s). This is opposed to \textsc{auxiliary simple} where the models are optimized in way that does not make them learn the auxiliary tasks at all. Hence, the optimization follows a normal training setting, as depicted in Equation~\ref{equation:auxiliary_loss}. In contrast to a normal training setting, here the gradients for the feature extractor(s), $\phi(\theta(x))$, are reversed. Thus, the partial derivative $\frac{\partial L_{(d)}}{\partial \theta_{(f)}}$ is scaled by $-\lambda$, which was set to $1$ for all experiments, such that the backpropagated gradient is simply reversed and therefore negative. See Figure~\ref{fig:grl_architecture} for an overview of the GRL model architecture and Figure~\ref{fig:mtl_model} for how gradients are reversed in my work.

\newpage

\section{Sequential Transfer}
\label{method:seq_transfer}

To investigate a model that is trained on both auxiliary tasks and the main task sequentially, I fine-tune BERT in its simplest set-up, that is without any additional custom encoding layer $\phi$, on all three tasks, namely context-domain classification, subjectivity classification and QA, sequentially. This can also be referred to as \textsc{soft parameter} sharing, where each task has both a separate feature encoder and task-specific output layer but information concerning the model's hidden representations is shared across tasks to ultimately help the model enhance performance on the main task.

To inspect whether information from the auxiliary tasks is useful to perform better on the main task, the model is evaluated on the train set in an additional epoch after convergence on each of the two auxiliary tasks. During this additional synthetic evaluation epoch, the model's raw output logits are stored for each input sequence $x_{i} \in {D}^{t}$. To yield smoother distributions and obtain actual probability scores, the logits are passed through a \textsc{sigmoid} (see Equation~\ref{equation:sigmoid}) and a \textsc{softmax} (see Equation~\ref{equation:softmax}) function for subjectivity and context-domain classification respectively. I refer to these vectors of concatenated probability scores as \textsc{soft targets}. During QA, the soft targets corresponding to both subjectivity and context-domain classification, $\mathbf{p}_{i} \in \mathbf{R}^{K}$, are concatenated with the matrix of hidden representations for each input sequence, $\mathbf{H}_{i}^{l} \in \mathbf{R}^{T \times D}$, at the last transformer layer before performing the classification, yielding a new matrix of hidden representations, $\mathbf{H}_{i}^{l} \in \mathbf{R}^{T \times (D + K)}$.\footnote{$K = 8$, $T = 512$, $D = 768$, $l = 6$.} 

In addition, I implement another set-up whose computations scarcely deviate from the aforementioned setting but leverage hard instead of soft targets in the concatenation part. I refer to this setting as \textsc{oracle}. The vector $\mathbf{p}_{i} \in \mathbf{R}^{K}$ consists of two parts, namely $\mathbf{p}_{i}^{s} \in \mathbf{R}^{K_{s}}$ and $\mathbf{p}_{i}^{d} \in \mathbf{R}^{K_{d}}$.\footnote{$K_{s} = 2$, $K_{d} = 6$.} Whereas $\mathbf{p}_{i}^{s} \in \mathbf{R}^{K_{s}}$ contains two $1$s, iff both the answer and the question are subjective, a single $1$ and one $0$, iff the answer or the question is subjective, and two $0$s, iff both are objective, $\mathbf{p}_{i}^{d} \in \mathbf{R}^{K_{d}}$ evaluates to a one-hot-encoded vector with a single $1$ at the index of the correct domain and $0$s otherwise. The concatenation with the matrix of hidden representations, $\mathbf{H}_{i}^{l} \in \mathbf{R}^{T \times D}$, is the same as above, thus yielding $\mathbf{H}_{i}^{l} \in \mathbf{R}^{T \times (D + K)}$.

In this set-up, no adversarial training is performed as the model is meant to learn each task separately. The architecture and the corresponding fine-tuning procedure is illustrated below in Figure~\ref{fig:seq_transfer_model}.

\begin{figure}[h!]
    \centering
    \includegraphics[width=.71\textwidth]{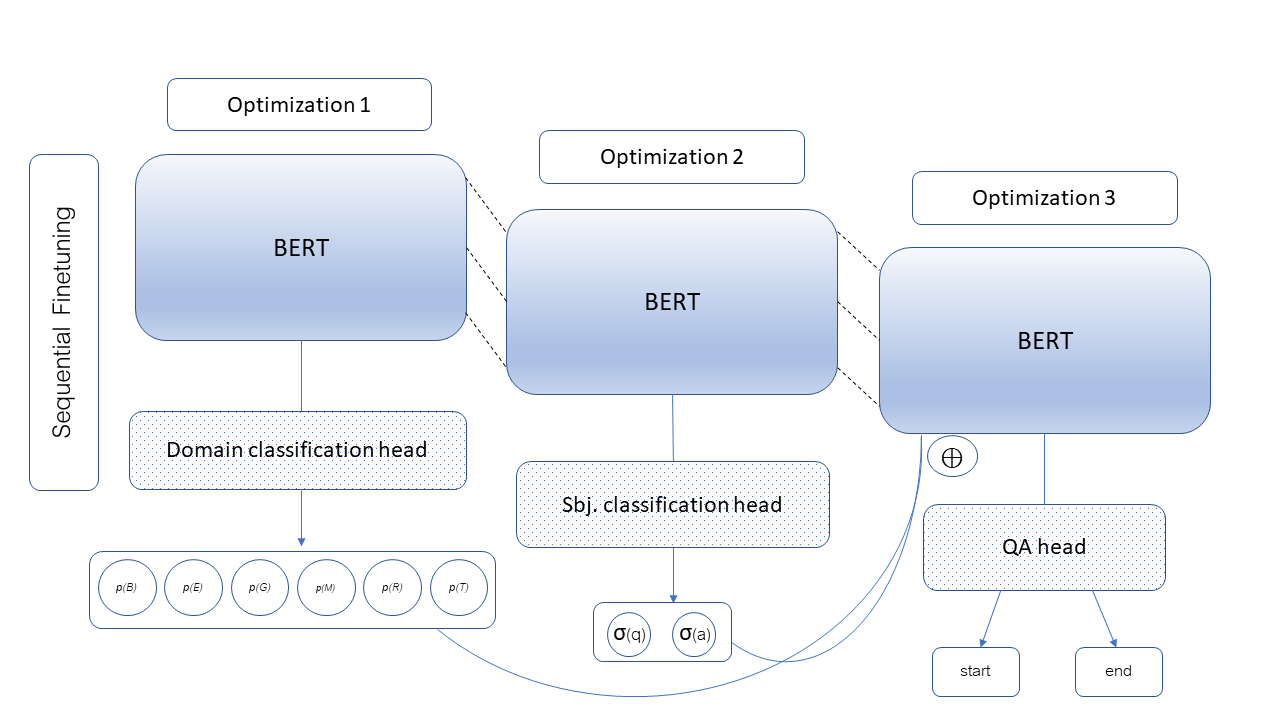}
    \caption{Sequential transfer. \textsc{bert} is sequentially fine-tuned on each task in the task set $\mathbf{T}$. In so doing, the model transfers knowledge from the previous tasks to the current task through \textsc{soft parameter} sharing, and ultimately injects information from the  auxiliary tasks into \textsc{bert} for QA through the concatenation of soft (auxiliary) targets with \textsc{bert}'s hidden representations for each token w.r.t. an input sequence $x_i$. The concatenation happens at the last Transformer layer only.}
    \label{fig:seq_transfer_model}
\end{figure}

%% file: data.tex
\section{SQuAD}

\begin{table}[h!]
\centering
\begin{tabular} {@{}l|c|c@{}}
\toprule
\textsc{QA type}&\multicolumn{1}{c}{\textsc{Question}}&\multicolumn{1}{c}{\textsc{Answer}}\\
\midrule
\midrule
\textsc{objective} & "How many awards did Beyoncé win at the 46th Grammy's Awards?" &  "five" \\
\textsc{objective} & "How many nights did Beyoncé play at the resort?" & "four"  \\
\textsc{objective} & "What instrument did Auguste Franchomme play?" & "cello"  \\
\textsc{objective} & "Many reviewers consider the second part of the book to be about what issue?" & "race relations"  \\
\bottomrule
\end{tabular}
\caption{Examples of answerable questions and their corresponding answers in SQuAD.}
\label{tab:qa_examples_squad}
\end{table}

At the time of writing, SQuAD is the most popular and largest span-based QA data set to train and evaluate machine reading systems on \cite{DBLP:journals/corr/abs-1806-03822,devlin2018bert}. There are two versions of SQuAD, namely SQuAD v1.0 \cite{DBLP:journals/corr/RajpurkarZLL16} and SQuAD v2.0 \cite{DBLP:journals/corr/abs-1806-03822}. Owing to the fact that SQuAD v2.0 is both the latest and more complex span-based QA data set of the two as well as more similar to SubjQA than SQuAD v1.0, I decided on exploiting SQuAD v2.0 only to compare against SubjQA. The main difference between SQuAD v1.0 and SQuAD v2.0 is that SQuAD v2.0 contains questions that are not answerable given the corresponding paragraph. Moreover, question-paragraph pair sequences are slightly longer in SQuAD v2.0 than in SQuAD v1.0 \cite{DBLP:journals/corr/abs-1806-03822}. This makes it similar to the nature of SubjQA which even consists of more unanswerable questions and longer context sequences than SQuAD v2.0 (see Table~\ref{tab:overview_squad_subjqa}). For simplicity and to avoid numbering, I will refer to SQuAD v2.0 with SQuAD throughout the following sections.

As depicted in Table~\ref{tab:qa_examples_squad}, answerable questions in SQuAD have a clear and objective answer whose string span is part of the corresponding Wikipedia paragraph. Questions and answers in SQuAD are both extracted from various Wikipedia articles, and thus contribute to highly accurate English grammar as most Wikipedia articles usually undergo proof-reading through independent reviewers. This is in stark contrast to the subjective QA dataset SubjQA as I will discuss in the following section.

\begin{table*}[ht]
\centering
\begin{tabular} {@{}l|rrrr|rrr@{}}
\toprule
\textsc{Source $\setminus$ Split}&\multicolumn{4}{c}{\textsc{Train}}&\multicolumn{1}{c}{\textsc{Dev}}&\multicolumn{1}{c}{\textsc{Test}}\\
& $n$ questions &\% answerable  &\% objective &\% subjective  &  $n$ questions &  $n$ questions \\ 
\midrule
SQuAD &  15,228 & 53.5 & 100.0 & 0.0  & 3,807 &  \_  \\
SubjQA & 14,630 & 44.0 & 17.3 & 82.7& 1,595  & 4,075  \\
\bottomrule
\end{tabular}
\caption{Overview of SQuAD and SubjQA across dataset splits.}
\label{tab:overview_squad_subjqa}
\end{table*}

\begin{figure*}[ht]
\centering
\includegraphics[width=1.0\textwidth]{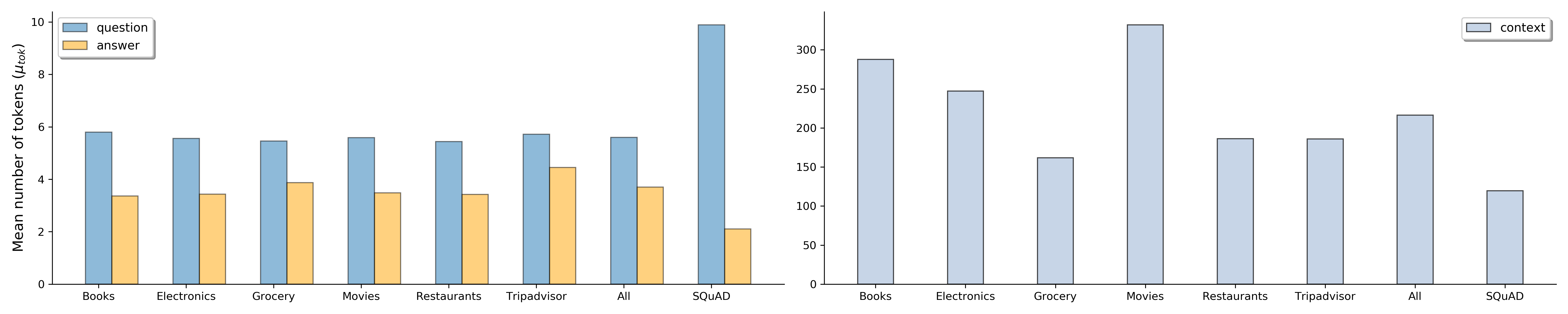}
\caption{Average number of tokens ($\mu$) per word sequence $x_i$ across all domains in SubjQA and SQuAD of which the latter was treated as its own domain. The bar graph on the left-hand side depicts the average number of tokens for questions and answers respectively. The right bar chart shows the average number of tokens for reviews (SubjQA) and Wikipedia paragraphs (SQuAD) respectively.}
\label{fig:squad_vs_subjqa_doc_lengths}
\end{figure*}

\section{SubjQA}

\begin{table}[h!]
\centering
\begin{tabular} {@{}l||rr|rrr@{}}
\toprule
\textsc{Domain $\setminus$ Source}&\multicolumn{2}{c}{\textsc{SQuAD}}&\multicolumn{3}{c}{\textsc{SubjQA}}\\
\midrule
& Train & Dev &  Train & Dev & Test \\ 
\midrule
& $n$ examples & $n$ examples &  $n$ examples  &  $n$ examples  &  $n$ examples \\ 
\midrule
books &  \_  &  \_ &  2,503 & 264 & 573 \\
electronics &  \_  & \_ & 2,382 & 267  & 675 \\
grocery &  \_  & \_ & 2,827 & 313 & 322\\
movies &  \_  & \_ & 2,456 & 273 & 632 \\
restaurants &  \_  & \_ & 2,349 & 231 & 799 \\
tripadvisor &  \_  & \_ & 2,113 & 247 & 1,074 \\
wikipedia &   15,228 & 3,807 & \_ & \_ & \_\\
\bottomrule
\end{tabular}
\caption{Distribution of paragraph or review domains across dataset splits in SQuAD and SubjQA respectively.}
\label{tab:domain_distribution}
\end{table}

SubjQA is a recently developed span-based QA data set that contains both objective and subjective questions \cite{subjqa2020}. Due to the fact that the dataset is meant to be subjective in nature the latter set of questions is with $82.7\%$ of the total number of questions highly overrepresented (see Table~\ref{tab:overview_squad_subjqa}). This makes it particularly difficult for QA models as they are required to learn about and understand the subjective features of a question $q_i$. Similarly to SQuAD, about half of the questions ($\sim 56.0\%$) are not answerable given the corresponding context. This is another detail of the dataset that makes it more difficult than other datasets that exclusively contain answerable questions. In contrast to SQuAD, SubjQA does not target common knowledge (e.g., "What is the birth place of Barack Obama?") which is likely to have occurred in the data used for pre-training of deep LMs such as \textsc{BERT} \cite{devlin2018bert,downeygetting}. The lack of targeting common knowledge, frequently contained in encyclopedias such as Wikipedia, enhances the difficulty of answering questions w.r.t. this dataset and makes fine-tuning indispensable.

\begin{table}[h!]
\centering
\begin{tabular} {@{}l|c|c@{}}
\toprule
\textsc{QA type}&\multicolumn{1}{c}{\textsc{Question}}&\multicolumn{1}{c}{\textsc{Answer}}\\
\midrule
\midrule
\textsc{objective} & "How is the read?" & "I thoroughly enjoyed reading about America"  \\
\textsc{subjective} & "Do you think the audio is very strong?" & "the sound is decent" \\
\midrule

\textsc{objective} & "How good is the camera of the nook?" & "the camera is excellent"   \\
\textsc{subjective} & "Do you want some tea?" & "I drink Lipton iced tea" \\
\midrule

\textsc{objective} & "Which flavor was there?"  & "salad"  \\
\textsc{subjective} & "What is the quality of the product?" & "I didn't think it was bad" \\

\midrule

\textsc{objective} &  "How helpful is the front desk?"  & "staff were pleasant and helpful" \\
\textsc{subjective} & "How good are the actors in this film?" & "the actors are brilliant"  \\

\bottomrule
\end{tabular}
\caption{Examples of answerable questions and their corresponding answers in SubjQA.}
\label{tab:qa_examples_subjqa}
\end{table}

In SubjQA, the context corresponding to a question $q_i$ is a review paragraph $r_i$ belonging to one of six domains, where $r_i \in$ \{\textsc{books},\textsc{electronics}, \textsc{grocery}, \textsc{movies}, \textsc{restaurants}, \textsc{tripadvisor}\}. A review $r_i$ never belongs to more than a single domain. As depicted in Table~\ref{tab:domain_distribution}, the data set is fairly balanced with respect to the different review domains. This is different to SQuAD, which does only consist of paragraphs extracted from Wikipedia (single domain). Hence, when fine-tuning on SubjQA it is crucial to inform a QA model about linguistic domain variances and shifts to not end up with an architecture that performs well on one or few domains but poorly on the rest.

\begin{figure}[h!]
\centering
\includegraphics[width=1.0\textwidth]{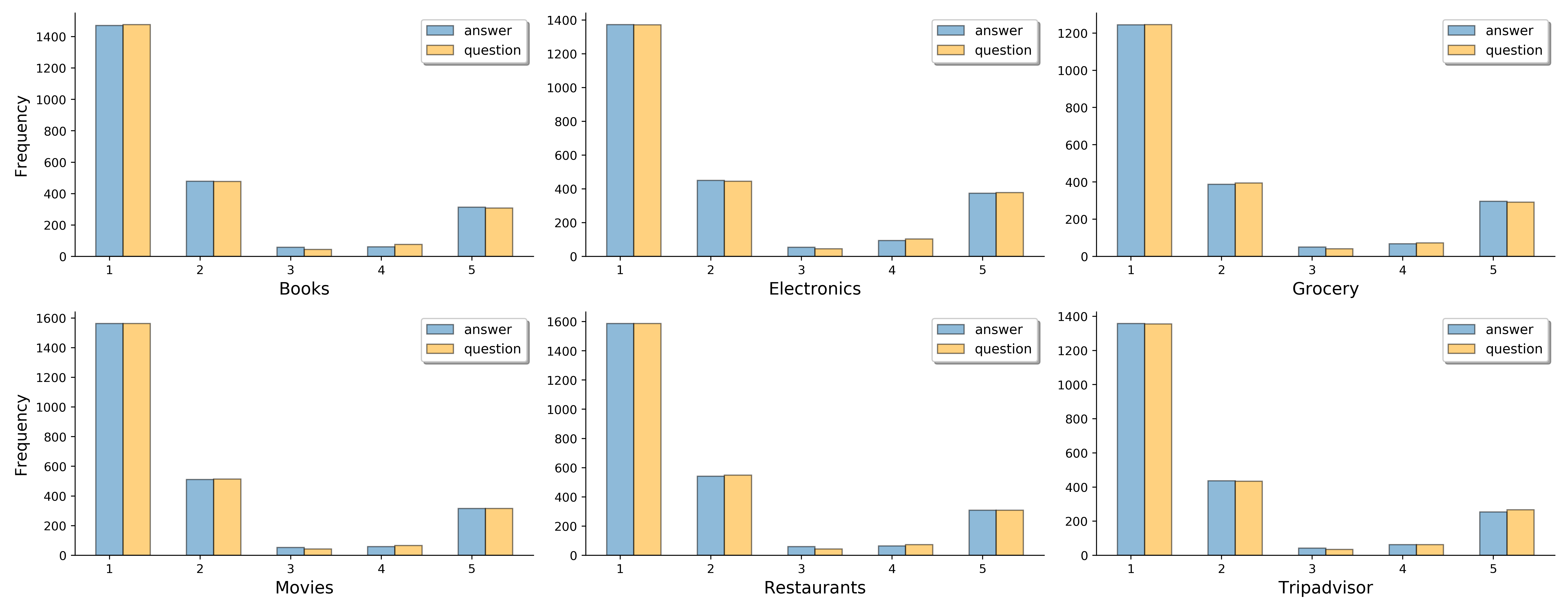}
\caption{Frequencies of objectivity vs. subjectivity levels assigned to questions and answers respectively by human crowdworkers across the different domains in SubjQA. Possible objectivity vs. subjectivity levels were integer values between 1 - 5 on a classic Likert scale. The lower the assigned value ($< 3$), the more likely the question or answer expressed a subjective opinion opposed to an objective, measurable fact.}
\label{fig:subj_levels_domains}
\end{figure}

\begin{table}[h!]
\centering
\begin{tabular} {@{}c|cc@{}}
\toprule
\multicolumn{1}{c}{\textsc{SQuAD}}&\multicolumn{2}{c}{\textsc{SubjQA}}\\
\midrule
\textsc{objective} &  \textsc{objective} & \textsc{subjective} \\
\midrule
\midrule

\texttt{what:} \: 43.97\% & \texttt{how:} \: 42.32\% & \texttt{how:} \: 61.41\% \\

\texttt{who:} \: 9.92\% & \texttt{what:} \: 23.80\% &  \texttt{what:} \: 16.53\% \\

\texttt{how:} \: 8.20\% & \texttt{is:} \: 10.49\% & \texttt{is:} \: 7.95\%\\

\texttt{when:} \: 5.30\% & \texttt{where:} \: 6.62\% & \texttt{does:} \: 2.93\% \\

\texttt{in:} \: 4.77\% & \texttt{does:} \: 4.68\% &  \texttt{do:} \: 2.23\%\\
\bottomrule
\end{tabular}
\caption{Top $5$ interrogative words for the train datasets of SQuAD and SubjQA respectively. For this purpose, SubjQA was split into objective and subjective questions according to the annotations human crowdworkers provided (see Figure~\ref{fig:subj_levels_domains}). Percentage to the right of each token denotes \% of questions that started with this prefix in the respective dataset split.}
\label{tab:question_prefix_distrib}
\end{table}

What's compelling about the data set is that question-answer types (i.e., subjective vs. objective) were manually annotated by human workers. This ensures the reliability of the labels. Crowdworkers were asked whether the respective question is asking about the subjective opinion of the reviewer or about an objective, measurable fact. Moreover, they were asked whether the selected answer span - crowdworkers had to select the correct answer span in the review - expresses a subjective opinion or an objective measurable fact. In so doing, workers had to to assign an integer value between $1 - 5$ according to a Likert scale \cite{allen2007likert} to both a question and the provided answer (see Figure~\ref{fig:subj_levels_domains}). The higher the assigned value, the more objective the question or answer respectively appeared to the crowdworker.

What can be inferred from Table~\ref{tab:question_prefix_distrib} is the fact the distributions of prefixes of questions between objective and subjective questions in the train set of $D_{SubjQA}$ do not notably differ from one another. It seems as if the vast majority of questions starts with the same prefix no matter whether the question was labelled objective or subjective by the human crowdworkers. The difference in the prefix distribution between $D_{SubjQA}$ and $D_{SQuAD}$, however, appears to be more apparent (see Table~\ref{tab:question_prefix_distrib}). This might reflect a potential caveat for the task of classifying a "question - context" $(\mathbf{q}, \mathbf{c})$ pair sequence into subjective vs. objective when exploiting $D_{SubjQA}$ only.

%% file: quantitative_analysis.tex
\section{Question Answering}
\label{section:qa_results}

\subsection{Single-task Learning}
\label{section:qa_results_stl}

\begin{table}[h!]
\begin{tabularx}{\textwidth} {@{}l||XX|XXr@{}}
\toprule
 \textsc{Model  $\setminus$ Fine-tuning} &\multicolumn{2}{c}{\textsc{SubjQA}} &\multicolumn{2}{c}{\textsc{Combined}}\\ &  Exact-match &   $F$1 &  Exact-match & $F$1\\ 
\midrule
BERT   & 76.04 &  76.49 &  \textbf{75.37} &  \textbf{76.13} & \\
BERT + Highway  &  75.95  & 76.57 &  73.86 &  75.37 &  \\
BERT + BiLSTM  &  \textbf{76.06} &  \textbf{76.93} &  74.63 &  75.79 &  \\
\midrule
$\bar \Theta_{\mathbf{QA}}$ & 76.02  & 76.66 &  74.62 &  75.76 &  \\
\bottomrule
\end{tabularx}
\caption{Single Task Learning (STL) - Question Answering (QA).  Models were either fine-tuned on \textsc{SubjQA} or both \textsc{SQuAD} and \textsc{SubjQA} which I refer to as \textsc{combined}, and evaluated on \textsc{SubjQA} only. Each model consisted of a pre-trained \textsc{distilbert} feature extractor, custom \textsc{post-BERT} encoding layers and a task-specific (QA) output layer that were all jointly fine-tuned on either of the two $D_{i \: \in \:  \{subj, \: comb\}}$ versions. Best results are depicted in bold face.}
\label{tab:stl_main_results}
\end{table}

In a STL setting, all implemented models were fine-tuned on $D_{i \: \in \:  \{subj, \: comb\}}$ to exclusively perform QA. Each model consisted of a pre-trained \textsc{distilbert base} feature extractor, custom \textsc{post-BERT} encoding layers (see Section~\ref{section:method} for further details) and a task-specific output layer for QA that were all jointly fine-tuned on either of the two $D_{i \: \in \:  \{subj, \: comb\}}$ versions. Inference was performed on the test set of $D_{subj}$ only to evaluate which fine-tuning regime yields better model performance in respect of SubjQA. 

To validate the necessity of fine-tuning the models on $D_{i \: \in \:  \{subj, \: comb\}}$ before performing inference on $D_{subj}$, I tested a \textsc{distilbert base} model that was previously fine-tuned on SQuAD.\footnote{\url{https://huggingface.co/transformers/model_doc/distilbert.html\#distilbertforquestionanswering}} SQuAD is a span-selection QA dataset that consists of only objective questions from a single domain (i.e., \textsc{wikipedia}) \cite{DBLP:journals/corr/RajpurkarZLL16}. Hence, a model that was fine-tuned exclusively on SQuAD is likely to not perform well on $D_{subj}$. As expected, the performance of the pre-trained \textsc{distilbert base} model was with an exact-match and an $F1$-score of $34.59 \%$ and $39.66 \%$ rather moderate. The higher an $F$1-score, the better did a learner perform. To create conditions that allow for fair comparisons between models, I fine-tuned my \textsc{distilbert base} implementation on 80\% of the official SQuAD train set, and evaluated it on the test set of SubjQA. This model yielded an exact-match accuracy and an $F1$-score of $59.58 \%$ and $61.70\%$ respectively, which is significantly higher than the publicly available pre-trained version. This is most likely due to the fact that the publicly available model was fine-tuned on SQuAD v1.0, whereas I have exploited SQuAD v2.0 for all experiments. This model served as the \textsc{baseline} model. Hence, the performances of all subsequently implemented models were compared against its exact-match and $F1$-scores with respect to $D_{subj}$. This experiment was conducted to inspect \hyperref[section:rq]{\textbf{Research Question (RQ)}} 1. Recall that \hyperref[section:rq]{\textbf{RQ}} 1 aimed at investigating whether it is necessary to fine-tune BERT on SubjQA for achieving a high score or whether it is sufficient to use a model fine-tuned on SQuAD.

Fine-tuning a model on $D_{comb}$ yielded worse performance than fine-tuning a model on $D_{sbj}$ as shown in Table~\ref{tab:stl_main_results}. The results are, however, not considerably different. This indicates that fine-tuning exclusively on SubjQA appears crucial to achieve the highest possible performance. However, fine-tuning a model on $D_{comb}$ might let a learner perform well with respect to both SubjQA and SQuAD without a significant deterioration in performance compared to models trained solely on the task-specific datasets. 

\begin{figure}[h!]
\centering
\begin{subfigure}{.47\textwidth}
    \centering
    \captionsetup{justification=centering}
    \includegraphics[width=.95\textwidth]{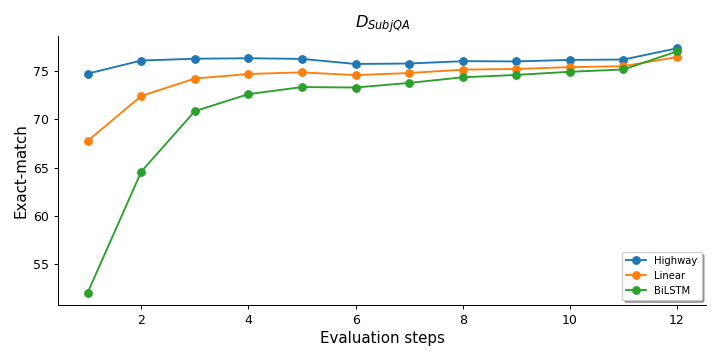}
    \caption{Exact-match (\textit{train})}
\end{subfigure}%
\begin{subfigure}{.47\textwidth}
    \centering
    \captionsetup{justification=centering}
    \includegraphics[width=0.9\textwidth]{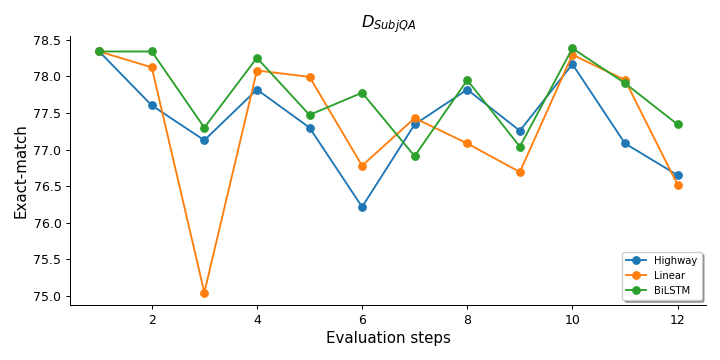}
    \caption{Exact-match (\textit{dev})}
\end{subfigure}
\begin{subfigure}{.47\textwidth}
    \centering
    \captionsetup{justification=centering}
    \includegraphics[width=.95\textwidth]{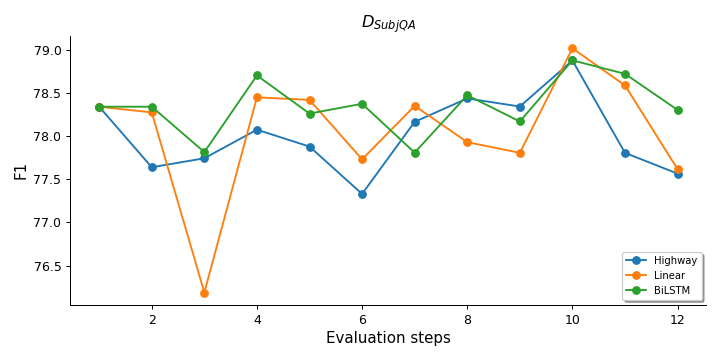}
    \caption{$F1$ (\textit{dev})}
\end{subfigure}%
\begin{subfigure}{.47\textwidth}
    \centering
    \captionsetup{justification=centering}
    \includegraphics[width=0.95\textwidth]{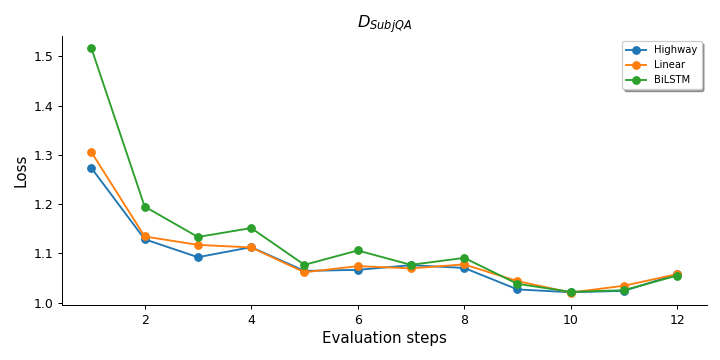}
    \caption{Evaluation loss (\textit{dev})}
\end{subfigure}
\caption[short]{STL training w.r.t. Question Answering (QA). Models were fine-tuned and evaluated on \textsc{SubjQA}. Depicted are exact-match accuracies, $F$1-scores and cross-entropy losses as a function of evaluation steps for both train and development sets of $D_{sbj}$ across all implemented STL QA models.}
\label{fig:qa_stl}
\end{figure}

The following analyses were performed to examine \hyperref[section:rq]{\textbf{RQ}} 2, which sought insight about the benefits of additional LSTM or Highway layers on top of BERT with respect to downstream performance. Indeed, additional \textsc{post-BERT} encoding layers, namely a Highway network or a BiLSTM, both increased $F$1-scores. The best STL model, BERT$_{\mathbf{QA}}$ + BiLSTM, achieved an $F$1-score of 76.93\% which is a relative improvement of .58\% over BERT$_{\mathbf{QA}}$ which scored 76.49\% $F$1. The improvement of models with additional \textsc{post-BERT} encoding layers, $\phi$, is also reflected in the $F$1-scores and exact-match accuracies with respect to the development set as a function of evaluation steps (see Figure~\ref{fig:qa_stl}). Performance on the validation set varied notably less for models with an additional Highway layer or a BiLSTM in-between \textsc{bert} and the linear QA output layer. Fewer fluctuations are in general an indicator of both more stable learning and less randomness involved in the performances. 

\newpage

\subsection{Multi-task Learning}
\label{section:qa_results_mtl}

 Experiments in this subsection were performed to investigate into \hyperref[section:rq]{\textbf{RQ}} 3. The goal of  \hyperref[section:rq]{\textbf{RQ}} 3 was to scrutinize whether MTL and adversarial training regimes improve upon single-task learning. 

\subsection{Parallel Transfer: QA and Subjectivity classification}

In the current MTL setting, mini-batches were alternated between all tasks in the task set $\mathbf{T} = \{T, T'\}$, namely QA (main), and subjectivity classification (auxiliary). In a \textsc{uniform sampling} setting, a model $\tilde \mathbf{f}$ was fine-tuned on each of the tasks $\frac{1}{2}$ of the time according to the strategy of uniformly sampling tasks outlined in Section~\ref{method:task_sampling}. In an \textsc{oversampling} setting, however, QA was sampled $\frac{2}{3}$ per epoch, and the remaining $\frac{1}{3}$ of training steps were allocated to subjectivity classification (see Section~\ref{method:task_sampling} for further details).

As can be inferred from Table~\ref{tab:mtl_sbj_main_results}, on average \textsc{oversampling} yielded better results than \textsc{uniform sampling}. The difference between the mean performances was statistically significant at $\alpha = .05$ according to an independent \textit{t}-test. Hence, oversampling the main task considerably increased performance over uniformly sampling both tasks. Additional \textsc{post-BERT} recurrent encoding layers (i.e., BiLSTMs) did not enhance performance over set-ups without such additional recurrent layers. This holds for the \textsc{uniform sampling} train version. Although slight improvements can be reported for an \textsc{oversampling} set-up, there is no statistical difference between the two implementations. In MTL, I did not perform experiments where learners leveraged (shared) Highway layers to keep a reasonable number of experiments and not exceed the constrained computational budget. In an \textsc{oversampling} setting, learners that were trained adversarially with respect to the auxiliary task improved over models that were not. 

The overall best model was a learner that was trained adversarially on subjectivity classification with respect to $(\mathbf{q, a})$ input sequences, namely BERT$_{\mathbf{QA} + \mathbf{Sbj}(\mathbf{q}, \mathbf{a})}$ + adversarial (simple), with an observed exact-match accuracy of 76.56\% and an $F$1-score of 76.94\%. This was the highest exact-match accuracy across training set-ups and model implementations. One other model, however, performed slightly better with respect to $F$1. This was another adversarially trained model, namely BERT$_{\mathbf{QA} + \mathbf{Sbj}(\mathbf{q}, \mathbf{c})}$ + BiLSTM + adversarial (GRL), which leveraged a \textsc{post-BERT} shared recurrent encoding layer and a gradient reversal layer (GRL) with respect to the auxiliary task. It achieved an $F1$-score of 76.98\% which was the highest reported $F1$-score overall.

\begin{table}
%\begin{footnotesize}
\begin{tabularx}{\textwidth} {@{}l||XX|XXr@{}}
\toprule
 \textsc{Model  $\setminus$ Fine-tuning} &\multicolumn{2}{c}{\textsc{SubjQA}} &\multicolumn{2}{c}{\textsc{Combined}}\\ &  Exact-match &   $F$1 &  Exact-match & $F$1 &\\
\midrule
\midrule
Auxiliary $\mathbf{1}$ - \textsc{uniform sampling} &   &  &  &  &  \\
\midrule
\midrule
BERT$_{\mathbf{QA} + \mathbf{Sbj}(\mathbf{q}, \mathbf{c})}$  &  76.24  &  76.94 &  74.29 &  75.69 &  \\
BERT$_{\mathbf{QA} + \mathbf{Sbj}(\mathbf{q}, \mathbf{c})}$ + adversarial (simple)  &  76.03  &  76.33 &  72.91 &  74.64 &  \\

BERT$_{\mathbf{QA} + \mathbf{Sbj}(\mathbf{q}, \mathbf{c})}$ + adversarial (GRL) &  75.69  &  76.60 &  76.36 &  76.57 &  \\
\midrule
BERT$_{\mathbf{QA} +  \mathbf{Sbj}(\mathbf{q}, \mathbf{a})}$  &   76.26  &  76.29 &  73.54 &  74.75 &  \\
BERT$_{\mathbf{QA} + \mathbf{Sbj}(\mathbf{q}, \mathbf{a})}$ + adversarial (simple)  &  75.67  &  76.43 &  72.97 &  74.20 &  \\

BERT$_{\mathbf{QA} + \mathbf{Sbj}(\mathbf{q}, \mathbf{a})}$ + adversarial (GRL)  &  76.26 &  76.47 &  72.85 &  74.34 &  \\
\midrule
BERT$_{\mathbf{QA} + \mathbf{Sbj}(\mathbf{q}, \mathbf{c})}$ + BiLSTM  &   76.30  &  76.40 &  74.55 &  75.68 &  \\
BERT$_{\mathbf{QA} + \mathbf{Sbj}(\mathbf{q}, \mathbf{c})}$ + BiLSTM + adversarial (simple)  &  75.34  &  76.31 & 74.13 &  75.43 &  \\
BERT$_{\mathbf{QA} + \mathbf{Sbj}(\mathbf{q}, \mathbf{c})}$ + BiLSTM + adversarial (GRL) &   74.57  &  75.83 &  74.15 &  75.45 &  \\
\midrule
BERT$_{\mathbf{QA} + \mathbf{Sbj}(\mathbf{q}, \mathbf{a})}$ + BiLSTM &    \textbf{76.38}  &  76.43 &  73.34 &  74.72 &  \\
BERT$_{\mathbf{QA} + \mathbf{Sbj}(\mathbf{q}, \mathbf{a})}$ + BiLSTM + adversarial (simple)  &   75.91 & 76.76 &  73.34 &  74.72 &  \\

BERT$_{\mathbf{QA} + \mathbf{Sbj}(\mathbf{q}, \mathbf{a})}$ + BiLSTM + adversarial (GRL)  &   75.91  & 76.76 &  74.78 &   75.87 &  \\
\midrule
$\bar \Theta_{\mathbf{QA} + \mathbf{Sbj}(\mathbf{q},\: \mathbf{a} \: \lor \: \mathbf{c})}$ &   75.88 & 76.46 & 73.93 & 75.17 & \\
\midrule
\midrule
Auxiliary $\mathbf{1}$ - \textsc{oversampling}   &   &  &  &  &  \\
\midrule
\midrule
BERT$_{\mathbf{QA} + \mathbf{Sbj}(\mathbf{q}, \mathbf{c})}$  &  75.95  &  76.23 &  74.17 &  75.34 &  \\
BERT$_{\mathbf{QA} + \mathbf{Sbj}(\mathbf{q}, \mathbf{c})}$ + adversarial (simple)  &  76.18  &  76.58 &  75.22 &  75.59 &  \\
BERT$_{\mathbf{QA} + \mathbf{Sbj}(\mathbf{q}, \mathbf{c})}$ + adversarial (GRL) &  76.22  &  76.55 &  76.14 &  \textbf{76.65} &  \\
\midrule
BERT$_{\mathbf{QA} + \mathbf{Sbj}(\mathbf{q}, \mathbf{a})}$  &   76.14  &  76.54 &  73.76 &  75.15 &  \\
BERT$_{\mathbf{QA} + \mathbf{Sbj}(\mathbf{q}, \mathbf{a})}$ + adversarial (simple)  &  \textbf{76.56}  &  76.94 &  75.91 & 76.51 &  \\

BERT$_{\mathbf{QA} + \mathbf{Sbj}(\mathbf{q}, \mathbf{a})}$ + adversarial (GRL)  &  76.03  &  76.32 &  72.99 &  74.63 &  \\
\midrule
BERT$_{\mathbf{QA} + \mathbf{Sbj}(\mathbf{q}, \mathbf{c})}$ + BiLSTM  &  76.18  &  76.90 &  75.97 &  76.44 &  \\
BERT$_{\mathbf{QA} + \mathbf{Sbj}(\mathbf{q}, \mathbf{c})}$ + BiLSTM + adversarial (simple)  & 76.36  & 76.58 &  \textbf{76.38} &  76.60 &  \\

BERT$_{\mathbf{QA} + \mathbf{Sbj}(\mathbf{q}, \mathbf{c})}$ + BiLSTM + adversarial (GRL) &   76.26  & \textbf{76.98} &  75.97 &  76.44 &  \\
\midrule
BERT$_{\mathbf{QA} + \mathbf{Sbj}(\mathbf{q}, \mathbf{a})}$ + BiLSTM &  76.12  &  76.93 &  74.82 &  76.21 &  \\
BERT$_{\mathbf{QA} + \mathbf{Sbj}(\mathbf{q}, \mathbf{a})}$ + BiLSTM + adversarial (simple)  &  76.18  & 76.86 & 73.62 &  74.98 &  \\
BERT$_{\mathbf{QA} + \mathbf{Sbj}(\mathbf{q}, \mathbf{a})}$ + BiLSTM + adversarial (GRL)  &   75.97  &  76.87 &  74.21 &  75.52 &  \\
\midrule
$\bar \Theta_{\mathbf{QA} + \mathbf{Sbj}(\mathbf{q},\: \mathbf{a} \: \lor \: \mathbf{c})}$ &  76.18 & 76.69 \textbf{*} & 75.00 \textbf{*} & 75.95 \textbf{*} & \\
\bottomrule
\end{tabularx}
%\end{footnotesize}
\caption{Multi-task learning (MTL) - Question Answering (QA) with one auxiliary task, namely subjectivity classification. In a \textsc{uniform sampling} task setting, all tasks - main and auxiliary task - were randomly sampled according to a uniform distribution, whereas in an \textsc{oversampling} setting, the main task (i.e, QA) was sampled $\frac{2}{3}$ per epoch. Models were either fine-tuned on \textsc{SubjQA} or both \textsc{SQuAD} and \textsc{SubjQA} which we call \textsc{combined}, and evaluated on \textsc{SubjQA} only. Each model consisted of a shared pre-trained \textsc{distilbert} feature extractor, optional shared \textsc{post-BERT} recurrent encoding layers (i.e., BiLSTMs) and task-specific output layers that were jointly fine-tuned on either of the two $D_{i \: \in \:  \{subj, \: comb\}}$ versions. \textsc{Adversarial simple} refers to adversarial training were the sign of the loss was simply reversed to make the model not learn the auxiliary task at all. \textsc{Adversarial GRL} refers to a more sophisticated adversarial strategy, namely a Gradient Reversal Layer (GRL) between the shared encoding layers and the task-specific output layers. \textbf{*} indicates a statistically significant difference between \textsc{oversampling} and \textsc{uniform sampling} according to an independent \textit{t}-test at $\alpha = .05$.}
\label{tab:mtl_sbj_main_results}
\end{table}
\newpage

\subsection{Parallel Transfer: QA, Subjectivity and Context-domain classification}

In the following MTL setting, mini-batches were alternated between all tasks in the task set $\mathbf{T} = \{T, T', T''\}$, namely QA, subjectivity classification, and context-domain classification.

When fine-tuning a model $\tilde \mathbf{f}$ on $D_{Subj}$, the learner is required to find a correct answer span $a_{i}$ that either reflects a subjective opinion or an objective, measurable fact. Thus, subjectivity classification (\textsc{aux}$_1$) appears to be a useful auxiliary task, as we have seen in the section above. Moreover, the answer span must be extracted from a review that belongs to different linguistic domains. Hence, context-domain classification (\textsc{aux}$_2$) might help a learner to better understand the review it must find an answer span $a$ in.

In a \textsc{uniform sampling} setting, a model $\tilde \mathbf{f}$ was fine-tuned on each of the tasks $\frac{1}{3}$ of the time according to the strategy of uniformly sampling tasks outlined in Section~\ref{method:task_sampling}. In an \textsc{oversampling} setting, however, QA was sampled $\frac{2}{3}$ per epoch, and the remaining $\frac{1}{3}$ of training steps were equally distributed among subjectivity and context-domain classification respectively, that is each auxiliary task was sampled $\frac{1}{6}$ per epoch. Since additional recurrent encoding layers did not yield a significant rise in performance in MTL with \textsc{aux}$_{1}$ (see Table~\ref{tab:mtl_sbj_main_results}), no experiments were performed where models leveraged a shared BiLSTM encoder.  

What becomes apparent from the results depicted in Table~\ref{tab:mtl_aux_2_main_results}, is the fact that \textsc{oversampling} clearly outperformed \textsc{uniform sampling} in this MTL setting. I performed an independent \textit{t}-test with respect to the results to test for statistical significance between the two task-sampling strategies. On average, models yielded significantly better exact-match accuracies and $F$1 scores in an \textsc{oversampling} setting compared to a \textsc{uniform sampling} setting with $p < .05$. MTL concerning all three tasks, however, seems to be less helpful for the main task compared to sampling solely between QA and subjectivity classification. Note that none of the models could outperform the best model, BERT$_{\mathbf{QA} + \mathbf{Sbj}(\mathbf{q}, \mathbf{a})}$ + adversarial (simple), and that on average performance was worse for this compared to the previous MTL set-up with just \textsc{aux}$_1$.
\newpage

\begin{table}
\begin{tabularx}{\textwidth} {@{}l||XXr@{}}
\toprule
 \textsc{Model  $\setminus$ Fine-tuning} &\multicolumn{2}{c}{\textsc{SubjQA}}\\ &  Exact-match &   $F$1 & \\
\midrule
\midrule
Auxiliary $\mathbf{1 \: \& \: 2}$ - \textsc{uniform sampling} &   &  &  \\
\midrule
\midrule
BERT$_{\mathbf{QA} + \mathbf{Dom}(\mathbf{q}, \mathbf{c}) + \mathbf{Sbj}(\mathbf{q}, \mathbf{c})}$ &  75.24  &  75.85 &  \\
BERT$_{\mathbf{QA} + \mathbf{Dom}(\mathbf{q}, \mathbf{c}) + \mathbf{Sbj}(\mathbf{q}, \mathbf{c})}$ + adversarial (simple) &  75.24  & 76.22 & \\
BERT$_{\mathbf{QA} + \mathbf{Dom}(\mathbf{q}, \mathbf{c}) + \mathbf{Sbj}(\mathbf{q}, \mathbf{c})}$ + adversarial (GRL) &  74.78 &  74.99 &  \\
\midrule
BERT$_{\mathbf{QA} + \mathbf{Dom}(\mathbf{q}, \mathbf{c}) + \mathbf{Sbj}(\mathbf{q}, \mathbf{a})}$ &  75.45  &  75.75 &  \\

BERT$_{\mathbf{QA} + \mathbf{Dom}(\mathbf{q}, \mathbf{c}) + \mathbf{Sbj}(\mathbf{q}, \mathbf{a})}$ + adversarial (simple) &  75.73 &  75.73 &  \\

BERT$_{\mathbf{QA} + \mathbf{Dom}(\mathbf{q}, \mathbf{c}) + \mathbf{Sbj}(\mathbf{q}, \mathbf{a})}$ + adversarial (GRL)  &  75.45  &  75.75 &  \\
\midrule
$\bar \Theta_{\mathbf{QA} + \mathbf{Dom}(\mathbf{q}, \mathbf{c}) + \mathbf{Sbj}(\mathbf{q},\: \mathbf{a} \: \lor \: \mathbf{c})}$ &  75.32 &  75.72  &  \\
\midrule
\midrule
Auxiliary $\mathbf{1 \: \& \: 2}$ - \textsc{oversampling}   &   &  &  \\
\midrule
\midrule
BERT$_{\mathbf{QA} + \mathbf{Dom}(\mathbf{q}, \mathbf{c}) + \mathbf{Sbj}(\mathbf{q}, \mathbf{c})}$ &  \textbf{76.16}  &  \textbf{76.49} & \\
BERT$_{\mathbf{QA} + \mathbf{Dom}(\mathbf{q}, \mathbf{c}) + \mathbf{Sbj}(\mathbf{q}, \mathbf{c})}$ + adversarial (simple) &  76.10  & 76.46 & \\

BERT$_{\mathbf{QA} + \mathbf{Dom}(\mathbf{q}, \mathbf{c}) + \mathbf{Sbj}(\mathbf{q}, \mathbf{c})}$ + adversarial (GRL) &  76.01  &  76.44 &  \\
\midrule
BERT$_{\mathbf{QA} + \mathbf{Dom}(\mathbf{q}, \mathbf{c}) + \mathbf{Sbj}(\mathbf{q}, \mathbf{a})}$ &  76.01 &  76.44 &  \\

BERT$_{\mathbf{QA} + \mathbf{Dom}(\mathbf{q}, \mathbf{c}) + \mathbf{Sbj}(\mathbf{q}, \mathbf{a})}$ + adversarial (simple) &  75.75  & 76.18 &  \\

BERT$_{\mathbf{QA} + \mathbf{Dom}(\mathbf{q}, \mathbf{c}) + \mathbf{Sbj}(\mathbf{q}, \mathbf{a})}$ + adversarial (GRL)  &  75.41  &  75.83 &  \\
\midrule
$\bar \Theta_{\mathbf{QA} + \mathbf{Dom}(\mathbf{q}, \mathbf{c}) + \mathbf{Sbj}(\mathbf{q},\: \mathbf{a} \: \lor \: \mathbf{c})}$ &  75.91 \textbf{*} &  76.31 \textbf{*} &  \\
\bottomrule
\end{tabularx}
%\end{footnotesize}
\caption{Multi-task learning (MTL) - Question Answering (QA) with two auxiliary tasks, namely subjectivity and context-domain classification. In a \textsc{uniform sampling} task setting, all tasks - main and auxiliary tasks - were randomly sampled according to a uniform distribution, whereas in an \textsc{oversampling} setting, the main task (i.e, QA) was sampled $\frac{2}{3}$ per epoch and $\frac{1}{3}$ was equally distributed among the \textsc{aux} tasks. Models were fine-tuned and evaluated on \textsc{SubjQA} or both \textsc{SQuAD}. Each model consisted of a shared pre-trained \textsc{distilbert} feature extractor and task-specific fully-connected output layers that were jointly fine-tuned on $D_{subj}$. \textsc{Adversarial simple} refers to adversarial training were the sign of the loss was simple reversed to make the model not learn the auxiliary task at all. \textsc{Adversarial GRL} refers to a more sophisticated adversarial strategy, namely a Gradient Reversal Layer (GRL) between the shared encoding layers and the task-specific output layers. \textbf{*} indicates a statistically significant difference between \textsc{oversampling} and \textsc{uniform sampling} according to an independent \textit{t}-test a $\alpha < .05$.}
\label{tab:mtl_aux_2_main_results}
\end{table}
\newpage

\newpage
\subsection{Parallel Transfer: QA and Context-domain classification}

In this MTL setting, mini-batches were alternated between QA (main) and context-domain classification (auxiliary). Hence, in a \textsc{uniform sampling} setting a model $\tilde \mathbf{f}$ was fine-tuned on QA in $50 \%$ of all training steps and on context-domain classification in the other half of training iterations. In an \textsc{oversampling} setting, QA was sampled $\frac{2}{3}$ per epoch and context-domain classification $\frac{1}{3}$ to make sure the model is exposed to the main task more frequently.

Interestingly, when compared against an MTL setting where mini-batches were alternated between QA and subjectivity classification, there was no statistical difference between the two MTL versions in a \textsc{uniform sampling} setting with respect to both the models' exact-match accuracies and $F1$-scores. However, with \textsc{oversampling} of the main task, MTL with \textsc{aux}$_{1}$ performed significantly better than MTL with \textsc{aux}$_{2}$ according to an independent \textit{t}-test at $\alpha = .05$ where $p < .05$. Moreover, the current MTL set-up was the only MTL version where there was no difference between uniformly sampling tasks and oversampling the main task with respect to model performance (see Tables~\ref{tab:mtl_sbj_main_results}, ~\ref{tab:mtl_aux_2_main_results}, ~\ref{tab:mtl_domain_main_results}).

It seems as if in a \textsc{uniform sampling} setting, none of the two auxiliary tasks helped the model to enhance its performance on the main task. In contrast, in an \textsc{oversampling} setting, subjectivity classification contributed to better QA performance on SubjQA, whereas context-domain classification did not help to answer subjective questions more accurately than STL. Models performed significantly worse in the latter MTL setting when compared to the former.

\begin{table}[t]
\begin{tabularx}{\textwidth} {@{}l||XXr@{}}
\toprule
 \textsc{Model  $\setminus$ Fine-tuning} &\multicolumn{2}{c}{\textsc{SubjQA}}\\ &  Exact-match &   $F$1 & \\
\midrule
\midrule
Auxiliary $\mathbf{2}$ - \textsc{uniform sampling} &   &  &  \\
\midrule
\midrule
BERT$_{\mathbf{QA} + \mathbf{Dom}(\mathbf{q}, \mathbf{c})}$ &  75.85  &  76.17 &  \\
BERT$_{\mathbf{QA} + \mathbf{Dom}(\mathbf{q}, \mathbf{c})}$ + adversarial (simple) &  \textbf{76.34}  &  76.48 & \\

BERT$_{\mathbf{QA} + \mathbf{Dom}(\mathbf{q}, \mathbf{c})}$ + adversarial (GRL) &  75.85  &  76.17 &  \\
\midrule
$\bar \Theta_{\mathbf{QA} + \mathbf{Dom}(\mathbf{q}, \mathbf{c})}$ &  76.01  &  76.27 &  \\
\midrule
\midrule
Auxiliary $\mathbf{2}$ - \textsc{oversampling}   &   &  &  \\
\midrule
\midrule
BERT$_{\mathbf{QA} + \mathbf{Dom}(\mathbf{q}, \mathbf{c})}$ &  75.65  &  76.35 &  \\
BERT$_{\mathbf{QA} + \mathbf{Dom}(\mathbf{q}, \mathbf{c})}$ + adversarial (simple) &  75.77  &  \textbf{76.49} &  \\

BERT$_{\mathbf{QA} + \mathbf{Dom}(\mathbf{q}, \mathbf{c})}$ + adversarial (GRL) &  75.65  &  76.35 &  \\
\midrule
$\bar \Theta_{\mathbf{QA} + \mathbf{Dom}(\mathbf{q}, \mathbf{c})}$ & 75.69  &  76.40 &  \\
\bottomrule
\end{tabularx}
\caption{Multi-task learning (MTL) - Question Answering (QA) with context-domain classification as the only auxiliary task. In a \textsc{uniform sampling} task setting, all tasks - main and auxiliary task - were randomly sampled according to a uniform distribution, whereas in an \textsc{oversampling} setting, the main task (i.e, QA) was sampled $\frac{2}{3}$ per epoch and $\frac{1}{3}$ was equally distributed among the \textsc{aux} tasks. Models were fine-tuned and evaluated on \textsc{SubjQA}. Each model consisted of a shared pre-trained \textsc{distilbert} feature extractor and task-specific fully-connected output layers that were jointly fine-tuned on $D_{subj}$. \textsc{Adversarial simple} refers to adversarial training were the sign of the loss was simple reversed to make the model not learn the auxiliary task at all. \textsc{Adversarial GRL} refers to a more sophisticated adversarial strategy, namely a Gradient Reversal Layer (GRL) between the shared encoding layers and the task-specific output layers.}
\label{tab:mtl_domain_main_results}
\end{table}
\newpage

\section{Sequential Transfer}

Furthermore, I've trained and evaluated the baseline model, BERT$_{\mathbf{QA}}$, fine-tuned on all tasks in the task set $\mathbf{T} = \{T, T', T''\}$ sequentially until model convergence. Here, opposed to the MTL with parallel transfer setting, where \textsc{hard parameter} sharing is employed, knowledge transfer follows the rules of \textsc{soft parameter} sharing (see Section~\ref{method:seq_transfer}).

As depicted in Table~\ref{tab:seq_transfer_main_results}, training a model sequentially on all tasks did not enhance performance on the main task for most of the deployed training set-ups. It even deteriorated performance significantly, when the model received information trough an oracle, that is the concatenation of \texttt{hard} targets with the model's hidden representations for each input token at the last transformer layer. In the version where subjectivity classification was performed with respect to $(\mathbf{q, c})$ input sequences performance did also decrease, but this time slightly rather than catastrophically. The only setting that yielded an increase in exact-match accuracy was the set-up, where \texttt{soft} targets, $\mathbf{p}_{i} \in \mathbf{R} ^ {k^{s}k^{d}}$, were concatenated with the hidden representations for each input token at the last transformer layer, $\mathbf{H}_{i}^{6} \in \mathbf{R}^{T \times 768}$, and subjectivity classification was performed with respect to $(\mathbf{q, a})$ input sequences (see Section~\ref{method:seq_transfer} for methodological details w.r.t. the concatenation). None of the implemented sequential transfer models could contribute to an increase in $F$1.  

One hypothesis to consider, is that the concatenation of \texttt{soft} targets - which encode probabilistic information about the auxiliary tasks - with the contextual hidden representations at the last transformer layer injected useful information about the natural language utterances in question and context into the model. Results have shown, however, that this additional information increased performance rather marginally.
\begin{table}[h!]
\begin{tabularx}{\textwidth} {@{}l||XXr@{}}
\toprule
 \textsc{Model  $\setminus$ Fine-tuning} &\multicolumn{2}{c}{\textsc{SubjQA}}\\ &  Exact-match &   $F$1 & \\
\midrule
\midrule
Auxiliary $\mathbf{1 \: \& \: 2}$ &   &  &  \\
\midrule
\midrule
BERT$_{\mathbf{QA}}$ (\texttt{hard}) &  63.25  &  65.58 &  \\
BERT$_{\mathbf{QA}}$ (\texttt{soft}) & 74.53  & 74.67 & \\
BERT$_{\mathbf{QA} + \mathbf{Sbj}(\mathbf{q}, \mathbf{a})}$ (\texttt{soft}) &  \textbf{76.40}  &  {76.40} & \\
\bottomrule
\end{tabularx}
\caption{Sequential transfer across tasks. BERT$_{\mathbf{QA}}$ was fine-tuned sequentially on context-domain classification, subjectivity classification and QA respectively (in this order). For the main task, QA, the model received information either through an oracle, namely \texttt{hard} targets, or the other already converged learners, namely \texttt{soft} targets, about the previous tasks.}
\label{tab:seq_transfer_main_results}
\end{table}
\newpage

\section{Fine-grained QA Results}

 The following investigations aimed at answering \hyperref[section:rq]{\textbf{RQ}} 4 \& 5. \hyperref[section:rq]{\textbf{RQ}} 4 motivated the analysis of the difference in QA performance between review domains, whereas \hyperref[section:rq]{\textbf{RQ}} 5 sought to infer the difficulty of subjective questions from interrogative words. 

\subsubsection{Domains} One crucial way to decipher a model's understanding of questions and their corresponding contexts is to examine its domain-specific performance if the dataset contains sentence pairs regarding various domains. What is interesting, is the observation that all evaluated models show a similar pattern concerning their domain-specific performance (see Table~\ref{tab:qa_results_per_domain}). Questions with respect to the domains \texttt{movies} and \texttt{books} were by far the easiest across the board. All evaluated models correctly predicted the answer span for $> 80\%$ and $> 79\%$ of questions regarding \texttt{movies} and \texttt{books} respectively. The difference in exact-match accuracies compared to the other four domains is more notable for a model that was fine-tuned on SQuAD. Recall that SQuAD consists of questions and paragraphs coming from a single domain only, namely \texttt{wikipedia} (see Table~\ref{tab:domain_distribution}). It is fair to assume that Wikipedia contains more paragraphs about \texttt{movies} and \texttt{books} than it does about \texttt{restaurants}, \texttt{grocery} or \texttt{tripadvisor}. Hence, a model fine-tuned on SQuAD might have encountered more questions concerning \texttt{movies} and \texttt{books} than regarding the other domains which could partly explain the difference in results.

Questions concerning reviews about \texttt{tripadvisor} were clearly the most difficult for all evaluated models (see Table~\ref{tab:qa_results_per_domain}). This can in part be explained through the fact that despite reviews regarding \texttt{tripadvisor} appearing the least often in the train set, most test examples belonged to this domain - almost twice as many as from other domains (see Table~\ref{tab:domain_distribution}).

The highest absolute and relative improvements of models fine-tuned on SubjQA over the baseline fine-tuned on SQuAD can be reported for the domains \texttt{grocery}, \texttt{restaurants} and \texttt{tripadvisor}. This might have a similar explanation as why the differences with respect to domain-specific performances are larger for a model fine-tuned on SQuAD (see above). The improvements of our best model, BERT$_{\mathbf{QA} + \mathbf{Sbj}(\mathbf{q}, \mathbf{a})}$ + adversarial (simple), over the baseline, BERT$_{\mathbf{QA}}$, do not appear to be vast but are highest for the domains \texttt{grocery} and \texttt{tripadvisor}. Both domains can be considered to be among the set of the more difficult domains.

\begin{table}[h!]
\centering
\begin{tabular} {@{}c|cc|c@{}}
\toprule
\textsc{domain  $\setminus$ fine-tuning} &
\multicolumn{2}{c}{\textsc{SubjQA}}&\multicolumn{1}{c}{\textsc{SQuAD}}\\
\midrule
& \textsc{best} & \textsc{baseline} & \textsc{baseline} \\
\midrule
\midrule
\texttt{movies} & \textbf{87.59}\% \: ($+7.13\%$) & \textbf{87.59}\% \: ($+7.13\%$) & 81.76\% \\
\texttt{books} & \textbf{84.23}\% \: ($+6.34\%$) & 84.00\% \: ($+6.05\%$) & 79.21\% \\
\texttt{electronics} & \textbf{81.46}\% \: ($+24.46\%$) & 81.01\% \: ($+23.77\%$) & 65.45\% \\
\texttt{grocery} & \textbf{75.59}\% \: ($+44.89\%$) & 74.58\% \: ($+42.96\%$) & 52.17\% \\
\texttt{restaurants} & \textbf{72.16}\% \: ($+68.13\%$) &  71.81\% \: ($+67.31\%$) & 42.92\% \\
\texttt{tripadvisor} & \textbf{57.25}\% \: ($+98.65\%$) &  55.88\% \: ($+93.89\%$) & 28.82\% \\
\midrule
\textsc{mean} & \textbf{76.38}\% \: ($+30.81\%$) &  75.81\% \: ($+29.83\%$) & 58.39\% \\
\bottomrule
\end{tabular}
\caption{Detailed exact-match accuracies for sentence pairs with respect to their review domain. Domains are sorted according to the respective model performance in descending order. Relative improvements over the baseline model fine-tuned on SQuAD (rightmost column) are depicted in parentheses. Best results are shown in bold face.}
\label{tab:qa_results_per_domain}
\end{table}

\subsubsection{Interrogative words} To disentangle the source of the errors with respect to interrogative words, I've computed the exact-match accuracy for each question starting with one of the question words that are depicted in Table~\ref{tab:question_prefix_distrib}.

Table~\ref{tab:results_per_q_word} shows that both models that were fine-tuned on SubjQA - BERT$_{\mathbf{QA} + \mathbf{Sbj}(\mathbf{q}, \mathbf{a})}$ + adversarial (simple) and BERT$_{\mathbf{QA}}$ -  achieved the highest relative improvements over BERT$_{\mathbf{QA}}$ fine-tuned on SQuAD for questions that started with \texttt{where} or \texttt{does}. This appears reasonable given the fact that both \texttt{where} and \texttt{does} are among the top-$k$ interrogative words for the train set of SubjQA but not for SQuAD (see Table~\ref{tab:question_prefix_distrib}). Thus, a model that was fine-tuned on SubjQA has seen questions starting with \texttt{where} or \texttt{does} considerably more often than a model fine-tuned on SQuAD. What is surprising, however, is the fact that questions starting with \texttt{how} report the least performance gains, although such questions amount to $\approx 50\%$ of all questions in SubjQA, and represent only $8\%$ of questions in SQuAD.

On the other hand, greater improvements can be reported for questions starting with \texttt{what} despite their significantly higher appearance in SQuAD compared to SubjQA. This might hint towards the post-hoc hypothesis of questions starting with \texttt{how} being "more" subjective than questions starting with \texttt{what} (according to Table~\ref{tab:question_prefix_distrib}), and thus are generally more difficult to answer than questions that start with \texttt{what}, as can be inferred from the poor scores for \texttt{how} and high scores for \texttt{what} questions across the board.

The best model, BERT$_{\mathbf{QA} + \mathbf{Sbj}(\mathbf{q}, \mathbf{a})}$ + adversarial (simple), improved over the baseline, BERT$_{\mathbf{QA}}$, mainly due to its enhanced performance regarding questions that start with \texttt{where}. For questions that start with one of the other top-$k$ interrogative words the performance was not significantly different between the two model versions (see Table~\ref{tab:results_per_q_word}). 

\begin{table}[h!]
\centering
\begin{tabular} {@{}c|cc|c@{}}
\toprule
\textsc{question word  $\setminus$ fine-tuning} &
\multicolumn{2}{c}{\textsc{SubjQA}}&\multicolumn{1}{c}{\textsc{SQuAD}}\\
\midrule
& \textsc{best} & \textsc{baseline} & \textsc{baseline} \\
\midrule
\midrule
\texttt{how} & \textbf{72.52}\% \: ($+28.01\%$) & 72.27\% \: ($+27.57\%$) & 56.65\% \\
\texttt{what} & \textbf{79.38}\% \: ($+23.70\%$) & 78.58\% \: ($+22.46\%$) & 64.17\% \\
\texttt{is} & \textbf{77.57}\% \: ($+29.37\%$) & 77.15\% \: ($+28.67\%$) & 59.96\% \\
\texttt{where} & \textbf{79.55}\% \: ($+52.19\%$) & 75.0\% \: ($+43.49\%$) & 52.27\% \\
\texttt{does} & \textbf{79.45}\% \: ($+39.19\%$) &  78.99\% \: ($+38.38\%$) & 57.08\% \\
\texttt{do} & 84.34\% \: ($+21.00\%$) &  \textbf{85.35}\% \: ($+22.45\%$) & 69.70\% \\
\midrule
\textsc{mean} & \textbf{78.80}\% \: ($+31.40\%$) & 77.89\%  \: ($+29.88\%$) & 59.97\% \\ 
\bottomrule
\end{tabular}
\caption{Detailed exact-match accuracies for sentence pairs whose questions start with one of the top-$k$ interrogative words across both objective and subjective questions in \textsc{SubjQA} (as depicted in Table~\ref{tab:question_prefix_distrib}). Relative improvements over the baseline model fine-tuned on \textsc{SQuAD} (rightmost column) are depicted in parentheses. Best results are shown in bold face.}
\label{tab:results_per_q_word}
\end{table}

\section{Subjectivity Classification}
\label{section:sbj_class_results}

\subsection{Binary}
To both better understand whether the different models show the ability to distinguish between subjective opinions and objective, measurable facts, and examine how difficult this particular auxiliary task (i.e., \textsc{aux}$_1$) is in general, a learner must be optimized exclusively to classify question-context ($\mathbf{q, c}$) or question-answer ($\mathbf{q, a}$) pair sequences into subjective vs. objective. Therefore, \textsc{BERT} was additionally fine-tuned solely on sequence classification as the main task $T$ without any other interfering task. Similarly to STL for QA, I fine-tuned \textsc{BERT} either on the train set of $D_{Subj}$ or $D_{Comb}$, and evaluated the models on the test set of $D_{Subj}$ only.

As in every other setting, each model consisted of a \textsc{distilbert} feature-extractor, optional custom encoding layers, and two task-specific fully-connected linear output layers for binary sequence classification, one for $\mathbf{q}$ and another for $\mathbf{c}$ or $\mathbf{a}$ depending on the set-up. Hence, the model had to classify both the question ($\mathbf{q}$) and its corresponding context ($\mathbf{c}$) or answer ($\mathbf{a}$) respectively into either a subjective opinion or an objective, measurable fact. Results are depicted in Table~\ref{tab:sbjclass_main_results}. Exclusively \textsc{macro} $F1$ scores are reported due to label imbalance (see Table~\ref{tab:overview_squad_subjqa}). The accuracy score is not an appropriate metric to measure a classifier's performance on an imbalanced dataset since if a learner was to predict the majority class for every sample (e.g., exclusively subjective question-answer pairs), it would yield a high score without having the metric reflect whether the model did understand anything about the data whatsoever. I have leveraged \textsc{macro} instead of \textsc{micro} averaging for $F$1 as the latter takes class imbalance into account and thus does not deviate much from the accuracy score. In contrast, \textsc{macro} averaging reveals insights about a learner's performance with respect to all classes independent of label (im-)balance.

\begin{figure}[h!]
\centering
\begin{subfigure}{.47\textwidth}
    \centering
    \captionsetup{justification=centering}
    \includegraphics[width=.95\textwidth]{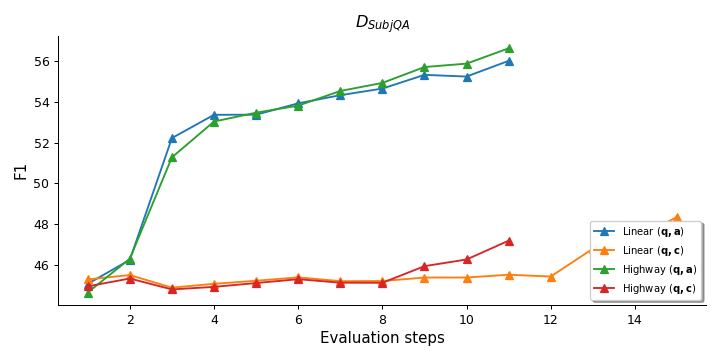}
    \caption{$F1$ (\textit{train})}
\end{subfigure}%
\begin{subfigure}{.47\textwidth}
    \centering
    \captionsetup{justification=centering}
    \includegraphics[width=0.9\textwidth]{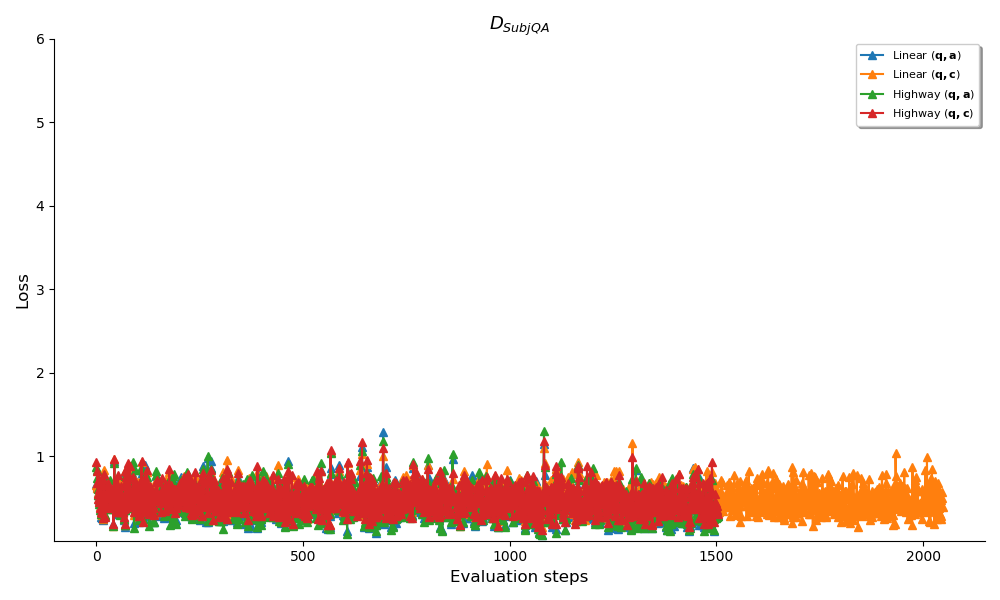}
    \caption{Batch loss (\textit{train})}
\end{subfigure}
\begin{subfigure}{.47\textwidth}
    \centering
    \captionsetup{justification=centering}
    \includegraphics[width=.95\textwidth]{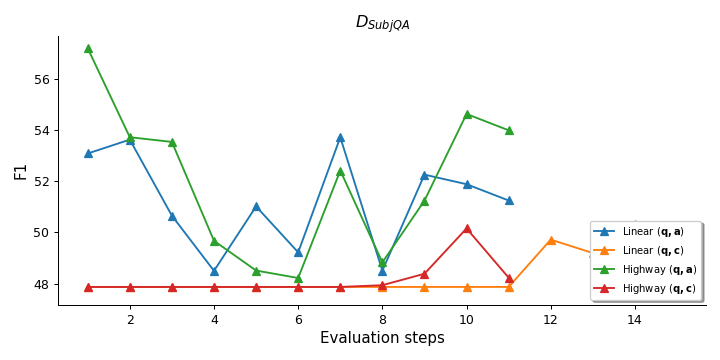}
    \caption{$F1$ (\textit{dev})}
\end{subfigure}%
\begin{subfigure}{.47\textwidth}
    \centering
    \captionsetup{justification=centering}
    \includegraphics[width=0.95\textwidth]{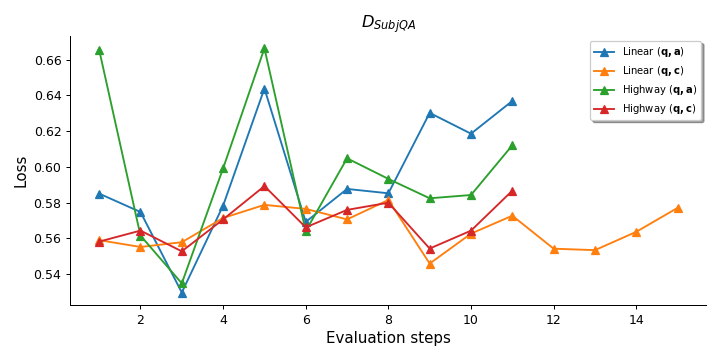}
    \caption{Evaluation loss (\textit{dev})}
\end{subfigure}
\caption[short]{Subjectivity classification. Models were fine-tuned and evaluated on \textsc{SubjQA}. Depicted are $F$1 scores and cross-entropy losses as a function of evaluation steps for both train and development sets of $D_{Subj}$ across all implemented STL models.}
\label{fig:sbj_class_subj}
\end{figure}

As can be inferred from Table~\ref{tab:sbjclass_main_results}, this task appeared to be difficult across the board. Not a single model yielded an $F1$-score $> 54.2\%$, which is not a particularly good results with respect to the general task of binary classification. The overall best model on this task, \textsc{BERT}$_{\mathbf{Sbj}(\mathbf{q}, \mathbf{a})}$, achieved an $F1$-score of $54.17 \%$, when fine-tuned on $D_{subj}$ and $52.80 \%$ when fine-tuned on $D_{comb}$ respectively. There is, however, no statistically significant difference with respect to the $F1$-scores between \textsc{BERT}$_{\mathbf{Sbj}(\mathbf{q}, \mathbf{a})}$ and \textsc{BERT}$_{\mathbf{Sbj}(\mathbf{q}, \mathbf{a})}$\textsc{+ Highway} according to an independent \textit{t}-test at $\alpha = .05$. Whether the model was optimized to classify ($\mathbf{q, c}$) or ($\mathbf{q, a}$), however, made a notable impact on model performance. $\bar \Theta_{\mathbf{Sbj}(\mathbf{q}, \mathbf{a})}$ performed better than $\bar \Theta_{\mathbf{Sbj}(\mathbf{q}, \mathbf{c})}$ in both fine-tuning set-ups. The difference between the average models was even statistically significant when trained on $D_{comb}$ at  $\alpha = .05$ (see Table~\ref{tab:sbjclass_main_results}). This is in line with training and evaluation curves displayed in Figure~\ref{fig:sbj_class_subj}, where both $F$1 scores on the train and development set respectively are higher and the minimum evaluation loss lower for models that were to classify ($\mathbf{q}, \mathbf{a}$) sequence pairs. For models that were optimized to classify ($\mathbf{q}, \mathbf{c}$) sequences it seems as if the models were not learning anything at all during training (see  Figure~\ref{fig:sbj_class_subj}). 

Interestingly, an additional recurrent neural model, namely BiLSTM, on top of \textsc{BERT} that takes into account temporal dependencies between timesteps in a sequence of tokens $\mathbf{x}$, did not help on this task. It neither deteriorated nor enhanced the model's performance. Therefore, I did not report \textsc{BERT} + BiLSTM results in Figures and Tables respectively.

\begin{table}[h!]
\begin{tabularx}{\textwidth} {@{}l||X|Xr@{}}
\toprule
 \textsc{Model  $\setminus$ Fine-tuning} &\multicolumn{1}{c}{\textsc{SubjQA}} &\multicolumn{1}{c}{\textsc{Combined}}\\
& $F$1 & $F$1\\ 
%\midrule
%BERT (baseline)   &  \_  &  \_ &  \_ &  \_ &  \\
\midrule
BERT$_{\mathbf{Sbj}(\mathbf{q}, \mathbf{c})}$  &  \textbf{54.17} & 51.92 & \\
BERT$_{\mathbf{Sbj}(\mathbf{q}, \mathbf{c})}$ + Highway  & 51.92 & 51.92 & \\
%BERT$_{\mathbf{Sbj}(\mathbf{q}, \mathbf{c})}$ + BiLSTM  & 51.92 & 51.92 &  \\
\midrule
$\bar \Theta_{\mathbf{Sbj}(\mathbf{q}, \mathbf{c})}$  &  53.05 &  51.92 &  \\
\midrule
BERT$_{\mathbf{Sbj}(\mathbf{q}, \mathbf{a})}$ &  \textbf{54.17} & 52.80 &  \\

BERT$_{\mathbf{Sbj}(\mathbf{q}, \mathbf{a})}$ + Highway & 52.79 & \textbf{53.07} & \\

%BERT$_{\mathbf{Sbj}(\mathbf{q}, \mathbf{a})}$ + BiLSTM  &  51.92 &  51.92 &  \\
\midrule
$\bar \Theta_{\mathbf{Sbj}(\mathbf{q}, \mathbf{a})}$ &  53.47  & 52.94 \textbf{*} &  \\
\bottomrule
\end{tabularx}
\caption{Subjectivity classification. Exclusively macro $F1$ scores reported due to class imbalance. Models were either fine-tuned on \textsc{SubjQA} or both \textsc{SQuAD} and \textsc{SubjQA} which I refer to as \textsc{combined}, and evaluated on \textsc{SubjQA} only. Each model consisted of a pre-trained \textsc{distilbert} feature extractor and task-specific output layers for sequence classification that were fine-tuned on either of the two $D_{i \: \in \:  \{subj, \: comb\}}$ versions. The abbreviation ($\mathbf{q}, \mathbf{c}$) refers to input sequences that consisted of question - context (i.e., question - review) sequence pairs, whereas ($\mathbf{q}, \mathbf{a}$) denotes question - answer pair input sequences.\textbf{*} indicates a statistically significant difference according to an independent \textit{t}-test with $p < .05$.}
\label{tab:sbjclass_main_results}
\end{table}

\newpage

%%% GRID PLOT OF TRAIN RESULTS FOR SUBJECTIVITY CLASSIFICATION %%%

%%% GRID PLOT OF TRAIN RESULTS FOR SUBJECTIVITY CLASSIFICATION %%%

\begin{figure}[h!]
\centering
\begin{subfigure}{.47\textwidth}
    \centering
    \captionsetup{justification=centering}
    \includegraphics[width=.95\textwidth]{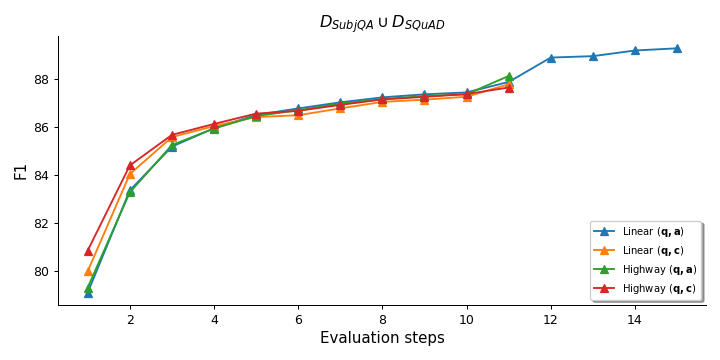}
    \caption{$F1$ (\textit{train})}
\end{subfigure}%
\begin{subfigure}{.47\textwidth}
    \centering
    \captionsetup{justification=centering}
    \includegraphics[width=0.9\textwidth]{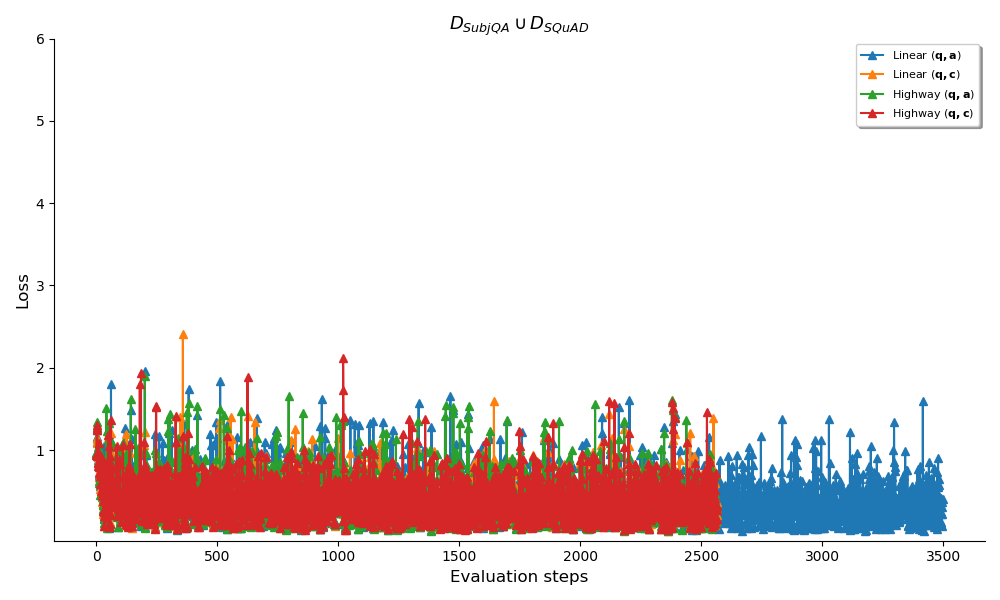}
    \caption{Batch loss (\textit{train})}
\end{subfigure}
\begin{subfigure}{.47\textwidth}
    \centering
    \captionsetup{justification=centering}
    \includegraphics[width=.95\textwidth]{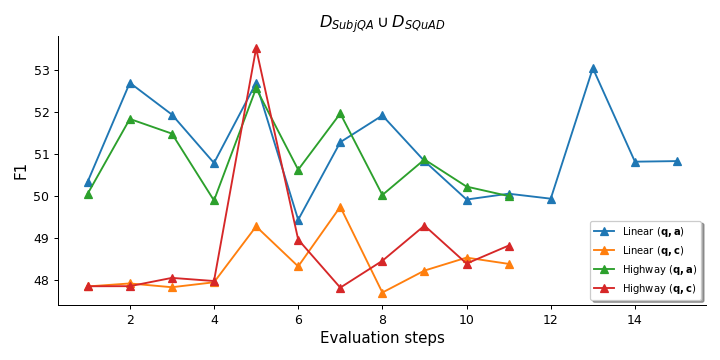}
    \caption{$F1$ (\textit{dev})}
\end{subfigure}%
\begin{subfigure}{.47\textwidth}
    \centering
    \captionsetup{justification=centering}
    \includegraphics[width=0.95\textwidth]{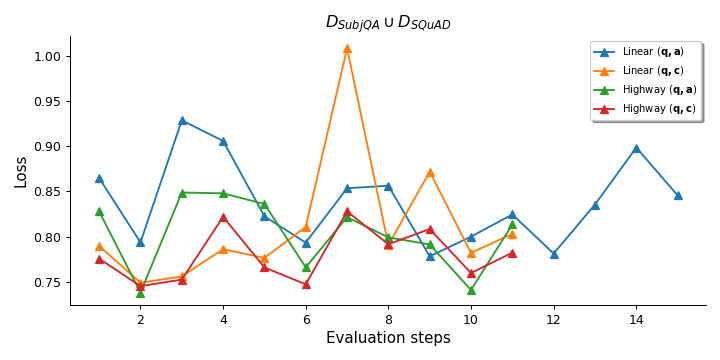}
    \caption{Evaluation loss (\textit{dev})}
\end{subfigure}
\caption[short]{Subjectivity classification. Models were fine-tuned on both \textsc{SQuAD} and \textsc{SubjQA} which we call \textsc{combined}, and evaluated on \textsc{SubjQA} only. Depicted are $F$1 scores and cross-entropy losses as a function of evaluation steps for train and development sets of $D_{Comb}$ and $D_{Subj}$ respectively across all  STL models.}
\label{fig:sbj_class_comb}
\end{figure}
\newpage

The results indicate that it is highly difficult for a model to distinguish between word sequences that contain subjective linguistic signals $\mathbf{x_{subj}}$ and word sequences that include objective linguistic cues $\mathbf{x_{obj}}$. This is most probably since question-context pair sequences $(\mathbf{q, c})$ on average consist of $100-300$ tokens per word sequence (see Figure~\ref{fig:squad_vs_subjqa_doc_lengths}) of which most are independent of subjective opinions or objective, measurable facts, and might therefore rather be domain- than opinion- or fact-specific. This is further reflected in the models' better classification performance on question-answer pair sequences $(\mathbf{q, a})$ compared to question-context pair sequences $(\mathbf{q, c})$ (see Table~\ref{tab:sbjclass_main_results}). It appears, however, that an answer's subjectivity might be reflected as part of the context and not in the context-free answer alone, which could explain the rather marginal improvements of classifying $(\mathbf{q, a})$ over classifying $(\mathbf{q, c})$, at least when both fine-tuned and evaluated on $D_{subj}$. That is another reason why I fine-tuned each model exclusively on context-domain classification as results with respect to subjectivity classification indicated that stronger signals might be reflected in the performance on the former task.

The following investigation was performed to conduct a more thorough error analysis. In so doing, I have examined whether the latter explanation is simply dataset specific, that is due to the linguistic nature of domain-variant reviews in $D_{subj}$. Hence, I fine-tuned the best model according to Table~\ref{tab:sbjclass_main_results} exclusively on the train set of $D_{comb}$ and evaluated the model on a synthetic test set $\in D_{comb}$ that consisted of the entire test set of $D_{subj}$ and $10\%$ of SQuAD's train set (see Table~\ref{tab:overview_squad_subjqa}). The results of this analysis are displayed in Table~\ref{tab:sbj_class_detailed}.

\begin{table}[h!]
\begin{footnotesize}
\begin{tabularx}{\textwidth} {@{}l||XXXr@{}}
\toprule
 \textsc{Model  $\setminus$ Fine-tuning} &\multicolumn{3}{c}{\textsc{Combined}}\\ &  $(\mathbf{q, c})_{\mathbf{SubjQA}}^{\mathbf{sbj}}$ &  $(\mathbf{q, c})_{\mathbf{SubjQA}}^{\mathbf{obj}}$ & $(\mathbf{q, c})_{\mathbf{SQuAD}}$ \\ 
\midrule
BERT$_{\mathbf{Sbj}(\mathbf{q}, \mathbf{c})}$  & 99.90\% & 0.00\% & 99.97\% & \\
\bottomrule
\end{tabularx}
\end{footnotesize}
\caption[short]{Fine-grained analysis of binary subjectivity classification. Depicted are accuracy scores per individual class. For this analysis, the objectivity class was split into question - context sentence pairs, $(\mathbf{q, c})$, belonging to \textsc{SubjQA} or \textsc{SQuAD} respectively.}
\label{tab:sbj_class_detailed}
\end{table}

The results in Table~\ref{tab:sbj_class_detailed} show that the model did not understand the objective class $\in  D_{Sbj}$ at all, and hence never made a correct classification with respect to sequences that belong to this class. The model most likely classified questions and answers respectively that belong to objective, measurable facts $\in  D_{Sbj}$  into subjective opinions $\in  D_{Sbj}$ as both sequence categories are part of the same dataset. To inspect whether the latter was the case or the model alternatively believes objective $(\mathbf{q, a \: \lor \: c})$ sequences $\in D_{Sbj}$ belong to $D_{Obj}$ which is entirely objective, one must conduct a multi-way sequence classification experiment, and in so doing perform a thorough analysis of both the model's predictions and hidden representations in latent space with respect to each class. 

The latter step is crucial to equip the model with the possibility to learn individual representations for each of the three categories, which is not possible in the binary classification task, where objective questions are encoded with the same label no matter whether they belong to \textsc{SubjQA} or \textsc{SQuAD}, and therefore optimized to learn feature representations for two different classes. Moreover, $F$1 scores concerning the binary classification task, where evaluation was performed on the synthetically created dataset, are not reliable since the model generally performed well on the objective class. Recall that all objective questions belong to the same class. Thus, the high $F1$ score of 87.24\% is not an adequate reflection of the model's comprehension of the three classes.
\newpage

\subsection{Multi-way} \label{section:multi_way_sbj_classification}To decipher the complexity of distinguishing between subjective and objective questions in \textsc{SubjQA}, I transformed the aforementioned subjectivity classification task from a binary into a multi-way classification problem. This time, the model was required to not only classify whether a question $\mathbf{q}_{i}$ was subjective or objective but had to differentiate between both subjective questions $\in$ \textsc{SubjQA}, objective questions $\in$ \textsc{SubjQA}, and objective questions $\in$ \textsc{SQuAD}. Hence, the model was trained to learn three classes instead of two, that is $(\mathbf{q, a \: \lor \: c}) \in \{D_{Sbj}^{obj}, D_{Sbj}^{sbj}, D_{Obj}\}$. 

This experiment was conducted to investigate whether the difficulty of distinguishing between subjective and objective questions is dataset-specific, that is owing to the linguistic nature of \textsc{SubjQA}, or a general one, that is due to the model's inability of understanding the semantic differences between subjective and objective questions. As mentioned in the previous section, \textsc{SQuAD} is a dataset that consists entirely of objective questions and answers extracted from Wikipedia paragraphs \cite{DBLP:journals/corr/RajpurkarZLL16,DBLP:journals/corr/abs-1806-03822}. Hence, if the model is indeed not capable of differentiating between subjective and objective questions, it will not perform well on the synthetically added \textsc{SQuAD} class either. Else, the poor performance concerning the binary classification task lies in the dataset and not in the capacity of the model. As in every training set-up, weighting of the loss with respect to each class was applied accordingly to account for label imbalance. 

\begin{figure}[h!]
\centering
\begin{subfigure}{.5\textwidth}
    \centering
    \captionsetup{justification=centering}
    \includegraphics[width=.90\textwidth]{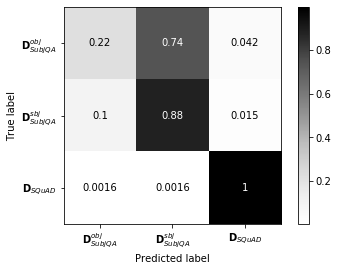}
    \caption{Multi-way $(\mathbf{q, a})$ classification}
\end{subfigure}%
\begin{subfigure}{.5\textwidth}
    \centering
    \captionsetup{justification=centering}
    \includegraphics[width=0.90\textwidth]{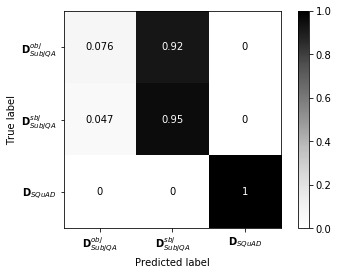}
    \caption{Multi-way $(\mathbf{q, c})$ classification}
\end{subfigure}
\caption{Normalized confusion matrices. The principal diagonal of the matrix depicts class-specific recall scores. The higher the score in the diagonal, the better did the model perform with respect to the corresponding class.}
\label{fig:sbj_multi_conf_mats}
\end{figure}

As can be inferred from the task-specific confusion matrices (see Figure~\ref{fig:sbj_multi_conf_mats}), the model was lacking the ability to distinguish between subjective and objective questions in \textsc{SubjQA}. The poor performances with respect to objective questions $\in$ \textsc{SubjQA} are in line with the binary classification results depicted in Table~\ref{tab:sbj_class_detailed}. The model could, however, perfectly differentiate between questions from \textsc{SubjQA} and \textsc{SQuAD} respectively. The model almost always predicted the subjective class for objective questions from \textsc{SubjQA} when classifying question - context, $(\mathbf{q, c})$, pair sequences (see Figure~\ref{fig:sbj_multi_conf_mats}b). It did a slightly better job in differentiating subjective from objective questions in \textsc{SubjQA} when classifying question - answer, $(\mathbf{q, a})$, pair sequences (see Figure~\ref{fig:sbj_multi_conf_mats}a). The latter task yielded with 68.26 macro $F$1 a 3.57\% relative improvement over 65.91 macro $F$1 in the former task, which is a considerable enhancement but still not an incredibly high performance. The difference between the two classification set-ups becomes more apparent in the 2D t-SNE plots of each sentence pair's semantic representation. The latter is visually depicted in Section ~\ref{section:qualitative_analysis}.

\newpage

\section{Context-domain Classification}

To inspect whether the second auxiliary task \textsc{aux}$_2$, namely context-domain classification, is useful to enhance QA with respect to reviews that belong to different domains, each of the STL models was exclusively fine-tuned on context-domain classification. As is depicted in Table~\ref{tab:domainclass_main_results}, it was fairly easy for the STL models, \textsc{BERT} and \textsc{BERT + HIGHWAY} in particular, to classify the questions and their corresponding reviews $(\mathbf{q}, \mathbf{c}$) into their respective domains $(\mathbf{y}^{d})$. 

In contrast to subjectivity classification, this time both accuracy and (macro) $F1$-scores are reported since both train and development sets were fairly balanced for domain labels (see Table~\ref{tab:domain_distribution}). If a model $\tilde \mathbf{f}$ was trained on $D_{comb}$, domains were weighted accordingly (see Section~\ref{method:finetuning} for further details). \textsc{BERT} and \textsc{BERT + HIGHWAY} achieved a macro $F1$ and an accuracy score of $> 98\%$  and $>99\%$ respectively in any fine-tuning setting (see Table~\ref{tab:domainclass_main_results}) - no matter whether the model was fine-tuned on $D_{subj}$ or $D_{comb}$. This means, that context-domains $(\mathbf{y}^{d})$ contain insightful linguistic signals that could lead to different results when answering questions about reviews from different domains, and hence might serve as a relevant auxiliary task in an MTL setting with respect to QA

What's interesting, however, is the fact that, in contrast to QA, where an additional RNN model between \textsc{BERT} and the fully-connected QA output layer helped (see Table~\ref{tab:stl_main_results}), a BiLSTM between \textsc{BERT} and the linear classification layer deteriorated the model's performance by an order of magnitude. I suspect this is because the special [CLS] token in any \textsc{BERT} model encodes semantic information of the entire sentence ($\mathbf{x}$) or sentence pair sequence  ($\mathbf{q}, \mathbf{c}$) respectively \cite{devlin2018bert}, but is not used to classify a sequence ($\mathbf{x}$) in an RNN based model. Any model based on recurrencies - opposed to linear connections that do not take the order of each token $x_i$ in a sequence $\mathbf{x}$ into account - encodes a sequence $\mathbf{x}$ timestep by timestep while taking into account each of the previous timesteps ($t-1, t-2, ..., t-i$). Therefore, in an RNN based model the vector at the last timestep $\mathbf{t}$ (i.e., last index in a matrix of continuous-valued vectors), is used as the input to the linear output layer to perform the classification task. This might result in a vector that does not encode the semantic information of the entire sequence thoroughly when used on top of \textsc{BERT}, since the special [CLS] token alone does already exploit this linguistic information. It is yet interesting to inspect why the latter computation is not feasible and decipher which timestep in an RNN module encodes the information contained in [CLS]. I leave this investigation for future work and encourage others to look deeper into this conundrum.

\begin{table}[h!]
\begin{tabularx}{\textwidth} {@{}l|XX||XXr@{}}
\toprule
 \textsc{Model  $\setminus$ Fine-tuning} &\multicolumn{2}{c}{\textsc{SubjQA}} &\multicolumn{2}{c}{\textsc{Combined}}\\
& Accuracy &   $F$1 &  Accuracy & $F$1\\ 
\midrule
BERT$_{\mathbf{Dom}(\mathbf{q}, \mathbf{c})}$   &  99.23  &  \textbf{98.65} &  \textbf{99.49} &  \textbf{98.89} &  \\
BERT$_{\mathbf{Dom}(\mathbf{q}, \mathbf{c})}$ + Highway  &  \textbf{99.25} &  98.61 &  99.37 & 98.75 &  \\
BERT$_{\mathbf{Dom}(\mathbf{q}, \mathbf{c})}$ + BiLSTM   &  45.58  &  28.62 &  40.73 & 24.52 &  \\
\bottomrule
\end{tabularx}
\caption{Context-domain classification. Models were either fine-tuned on \textsc{SubjQA} or both \textsc{SQuAD} and \textsc{SubjQA} which I refert to as \textsc{combined}, and evaluated on \textsc{SubjQA} only. Each model consisted of a pre-trained \textsc{distilbert} feature extractor, custom encoding layers on top of \textsc{BERT} and task-specific output layers for multi-class sequence classification that were fine-tuned on either of the two $D_{i \: \in \:  \{subj, \: comb\}}$ versions. The abbreviation ($\mathbf{q}, \mathbf{c}$) refers to input sequences that consisted of question - context (i.e., question - review) sequence pairs.}
\label{tab:domainclass_main_results}
\end{table}
\newpage

%%% GRID PLOTS OF TRAIN RESULTS FOR DOMAIN CLASSIFICATION %%%

\begin{figure}[h!]
\centering
\begin{subfigure}{.52\textwidth}
    \centering
    \captionsetup{justification=centering}
    \includegraphics[width=.99\textwidth]{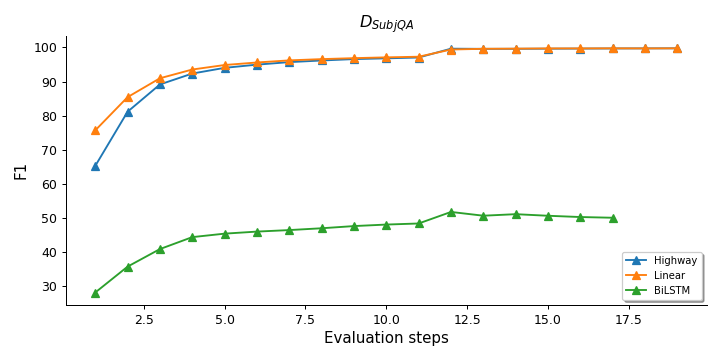}
    \caption{$F1$ (\textit{train})}
\end{subfigure}%
\begin{subfigure}{.52\textwidth}
    \centering
    \captionsetup{justification=centering}
    \includegraphics[width=0.95\textwidth]{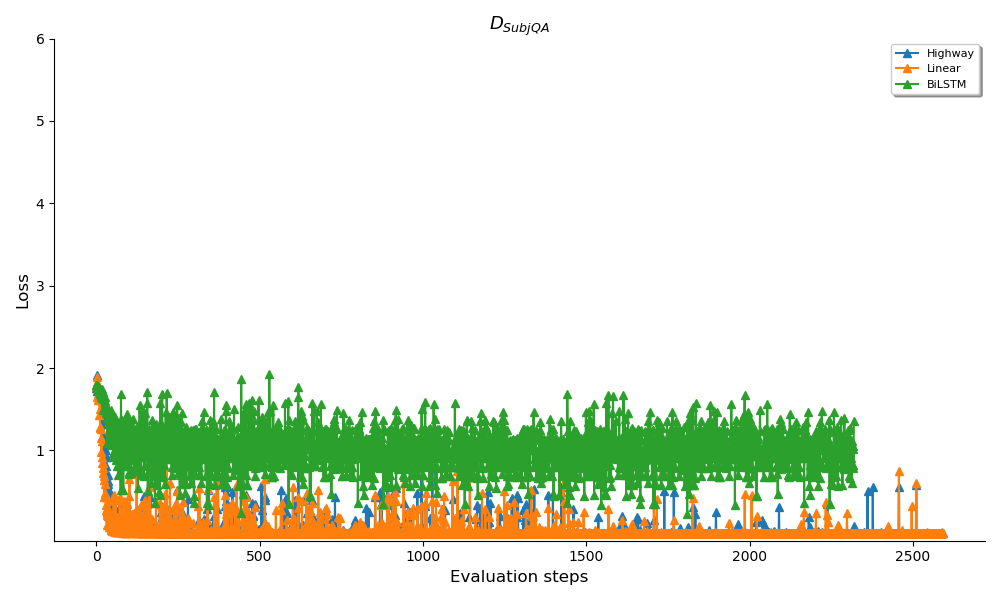}
    \caption{Batch loss (\textit{train})}
\end{subfigure}
\begin{subfigure}{.52\textwidth}
    \centering
    \captionsetup{justification=centering}
    \includegraphics[width=.99\textwidth]{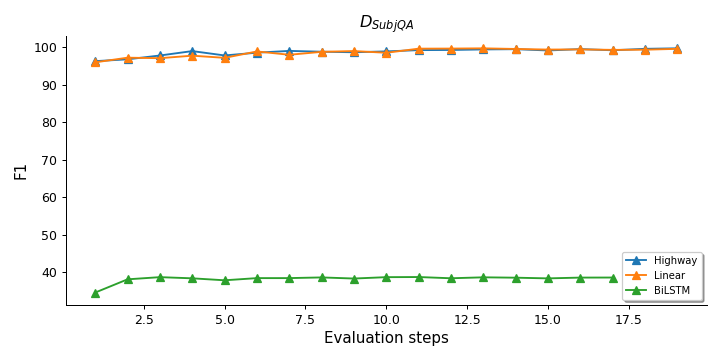}
    \caption{$F1$ (\textit{dev})}
\end{subfigure}%
\begin{subfigure}{.52\textwidth}
    \centering
    \captionsetup{justification=centering}
    \includegraphics[width=0.99\textwidth]{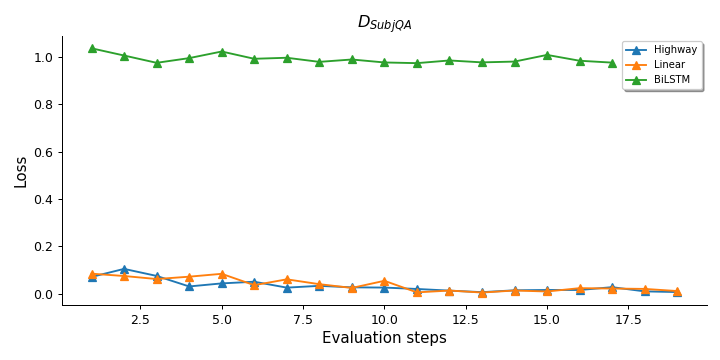}
    \caption{Evaluation loss (\textit{dev})}
\end{subfigure}
\caption{Context-domain classification. Models were fine-tuned on \textsc{SubjQA}, and evaluated on \textsc{SubjQA}. Depicted are $F$1 scores and cross-entropy losses as a function of evaluation steps for both train and development sets of $D_{Subj}$ across all implemented models.}
\label{fig:domain_class_comb}
\end{figure}

\iffalse
\begin{figure}[h!]
\centering
\begin{subfigure}{.52\textwidth}
    \centering
    \captionsetup{justification=centering}
    \includegraphics[width=.99\textwidth]{Figures/tasks/domain_classification/D_combined//batch_f1.png}
    \caption{$F1$ (\textit{train})}
\end{subfigure}%
\begin{subfigure}{.52\textwidth}
    \centering
    \captionsetup{justification=centering}
    \includegraphics[width=0.95\textwidth]{Figures/tasks/domain_classification/D_combined/batch_loss.png}
    \caption{Batch loss (\textit{train})}
\end{subfigure}
\begin{subfigure}{.52\textwidth}
    \centering
    \captionsetup{justification=centering}
    \includegraphics[width=.99\textwidth]{Figures/tasks/domain_classification/D_combined/val_f1.png}
    \caption{$F1$ (\textit{dev})}
\end{subfigure}%
\begin{subfigure}{.52\textwidth}
    \centering
    \captionsetup{justification=centering}
    \includegraphics[width=0.99\textwidth]{Figures/tasks/domain_classification/D_combined/val_loss.png}
    \caption{Evaluation loss (\textit{dev})}
\end{subfigure}
\caption{Context-domain classification. Models were fine-tuned on \textsc{SQuAD} and \textsc{SubjQA} which we call \textsc{combined}, and evaluated on \textsc{SubjQA} only. Depicted are $F$1 scores and cross-entropy losses as a function of evaluation steps for both train and development sets of $D_{Comb}$ across all implemented models.}
\label{fig:domain_class_comb}
\end{figure}
\fi
%\newpage

%% file: qualitative_analysis.tex
\section{Hidden Representations in Latent Space}
\label{section:hidden_reps_in_vec_space}

To decipher the performance, or more generally the behavior, of any neural network architecture, one is required to inspect a model's hidden representations at each of its spatial dimensions (opposed to a model's temporal dimensions such as in sequence modelling tasks with respect to temporal data). In neural networks, spatial dimensions or stages, are represented through layers. The more layers a neural network consists of, the deeper it is in space \cite{lecun2015deep}. Each layer refers to a different stage within the model's representational hierarchy. Usually, in earlier stages of a neural network, that is in the bottom layers, the model's representations in latent space draw attention to low-level features such as syntax in NLP models  (e.g., Part-of-Speech tags) \cite{PhDThesisBjerva} or edges in Computer Vision (CV) models \cite{DBLP:conf/icann/HintonKW11,DBLP:conf/nips/KrizhevskySH12,DBLP:conf/icpr/SermanetCL12,DBLP:conf/cvpr/DosovitskiySB15}. Contrary, later stages in the network, that is the top layers, focus on the representation of high-level features such as meaning in NLP models (e.g., relations between entities or the meaning of a word dependent on the context it appears in) \cite{DBLP:conf/emnlp/IrsoyC14,DBLP:conf/emnlp/LiuJM15,DBLP:conf/cncl/LiJH15,devlin2018bert} or abstract visual features in CV models (e.g., eyes, nose, mouth in faces) \cite{DBLP:conf/icml/JiXYY10,DBLP:conf/nips/KrizhevskySH12,lecun2015deep}.  

According to this general idea of a representational hierarchy in space, the notions of \texttt{question} and \texttt{answer} may be reflected in the top layers rather than in the early layers of the neural model. In this section, I will look deeper into the hidden representations of selected model architectures and examine what qualitative insights may be gained from those, why some classifications did not work as expected and most importantly at which stages in the hierarchy the model made mistakes. This investigation is dedicated to answer \hyperref[section:rq]{\textbf{RQ}} 6.

\section{Multi-way Subjectivity Classification}

Firstly, I will shed light on the task of multi-way \hyperref[section:multi_way_sbj_classification]{\textbf{subjectivity classification}} to better understand how subjectivity is reflected in the model's hidden representations. According to the \hyperref[section:multi_way_sbj_classification]{\textbf{quantitative results}} with respect to this task, the model seemingly could not distinguish between subjective and objective questions within $\mathbf{D}_{SubjQA}$, particularly when trained and evaluated on question - context pair sequences (see Figure~\ref{fig:sbj_multi_conf_mats}). 

\begin{figure}[h!]
\centering
\begin{subfigure}{.52\textwidth}
    \centering
    \captionsetup{justification=centering}
    \includegraphics[width=.99\textwidth]{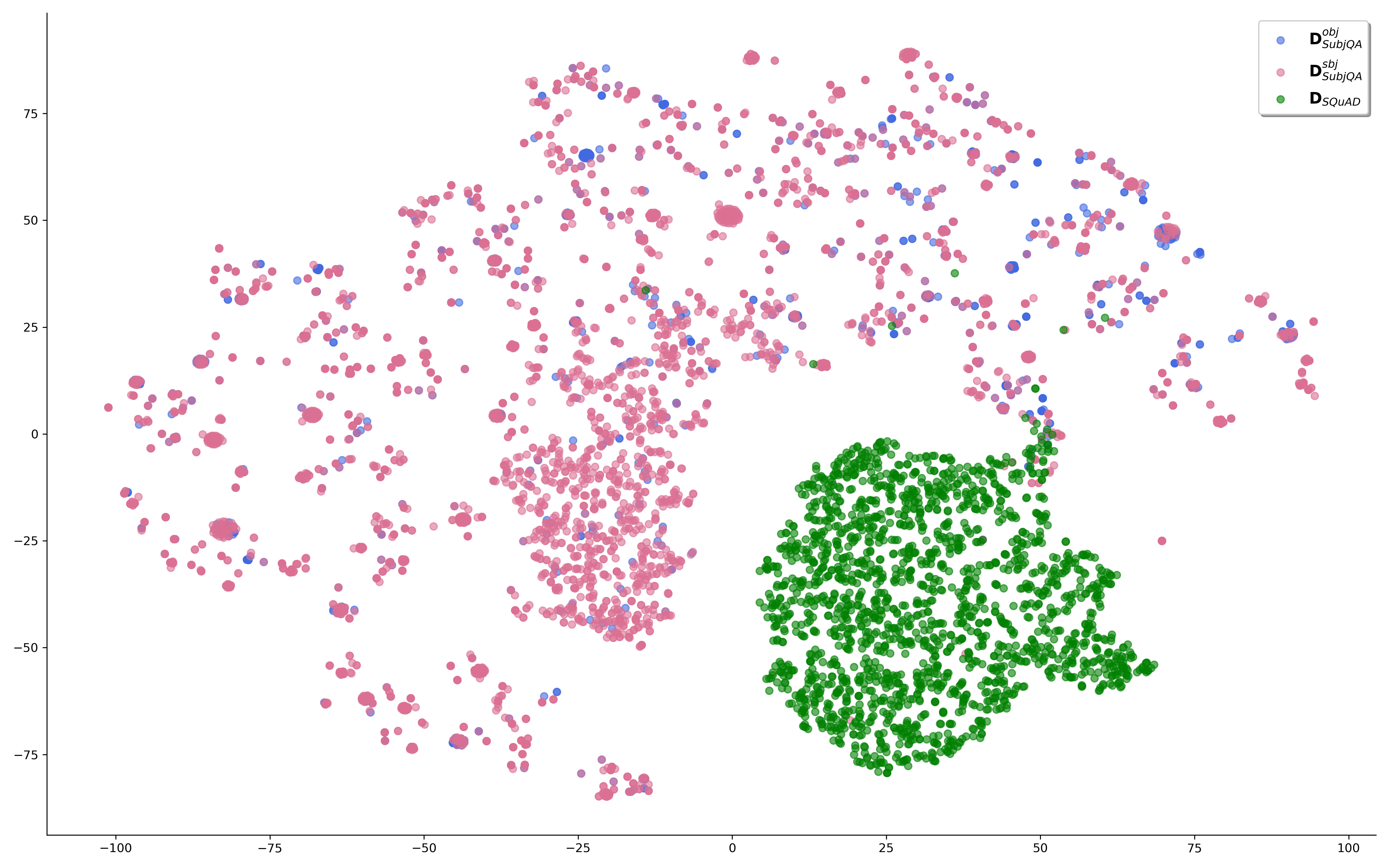}
    \caption{$(\mathbf{q, a})$ - $95\% \: \sigma^{2}$ retained in PCA}
\end{subfigure}%
\begin{subfigure}{.52\textwidth}
    \centering
    \captionsetup{justification=centering}
    \includegraphics[width=0.95\textwidth]{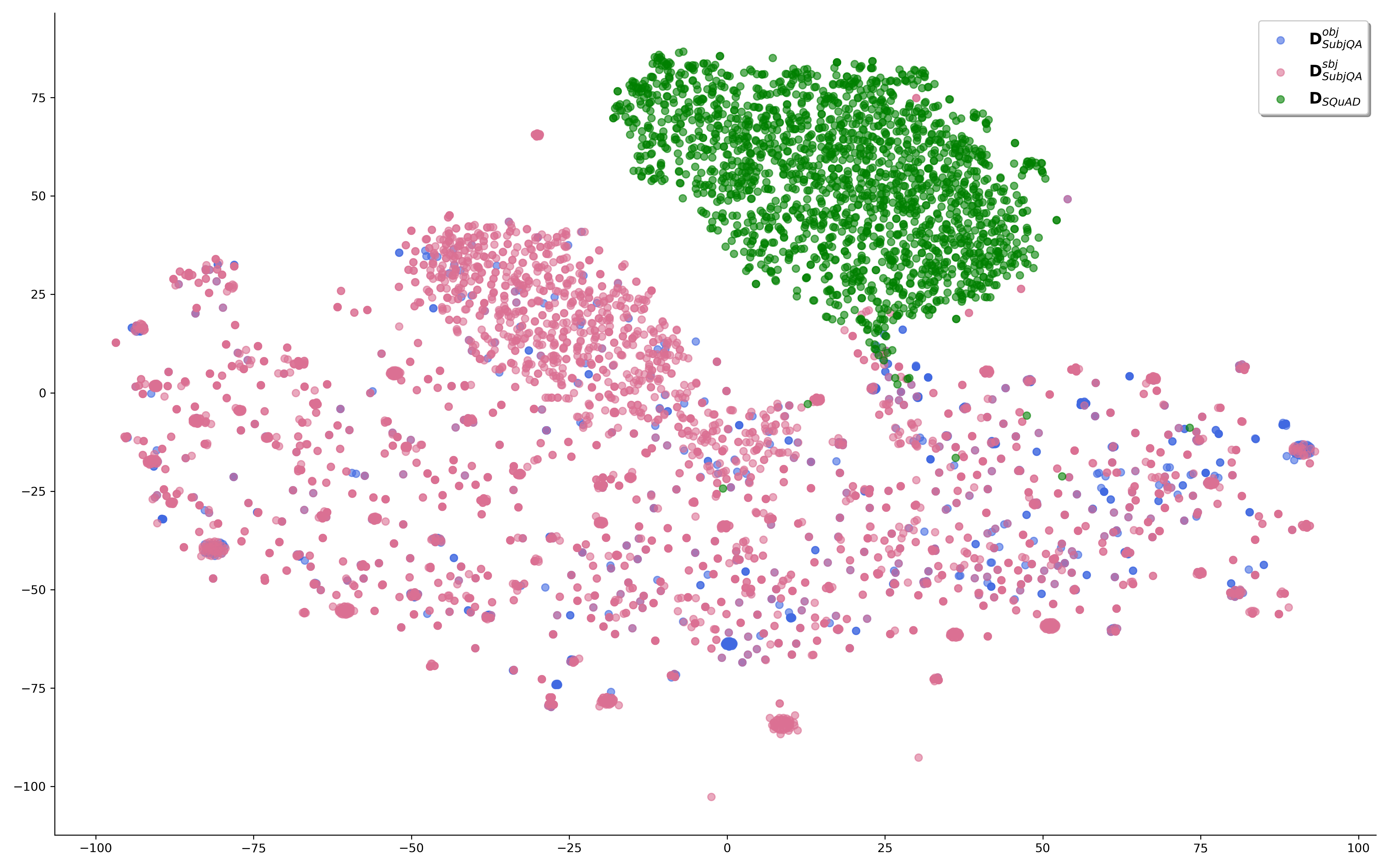}
    \caption{$(\mathbf{q, a})$ - $99\% \: \sigma^{2}$ retained in PCA}
\end{subfigure}
\begin{subfigure}{.52\textwidth}
    \centering
    \captionsetup{justification=centering}
    \includegraphics[width=.99\textwidth]{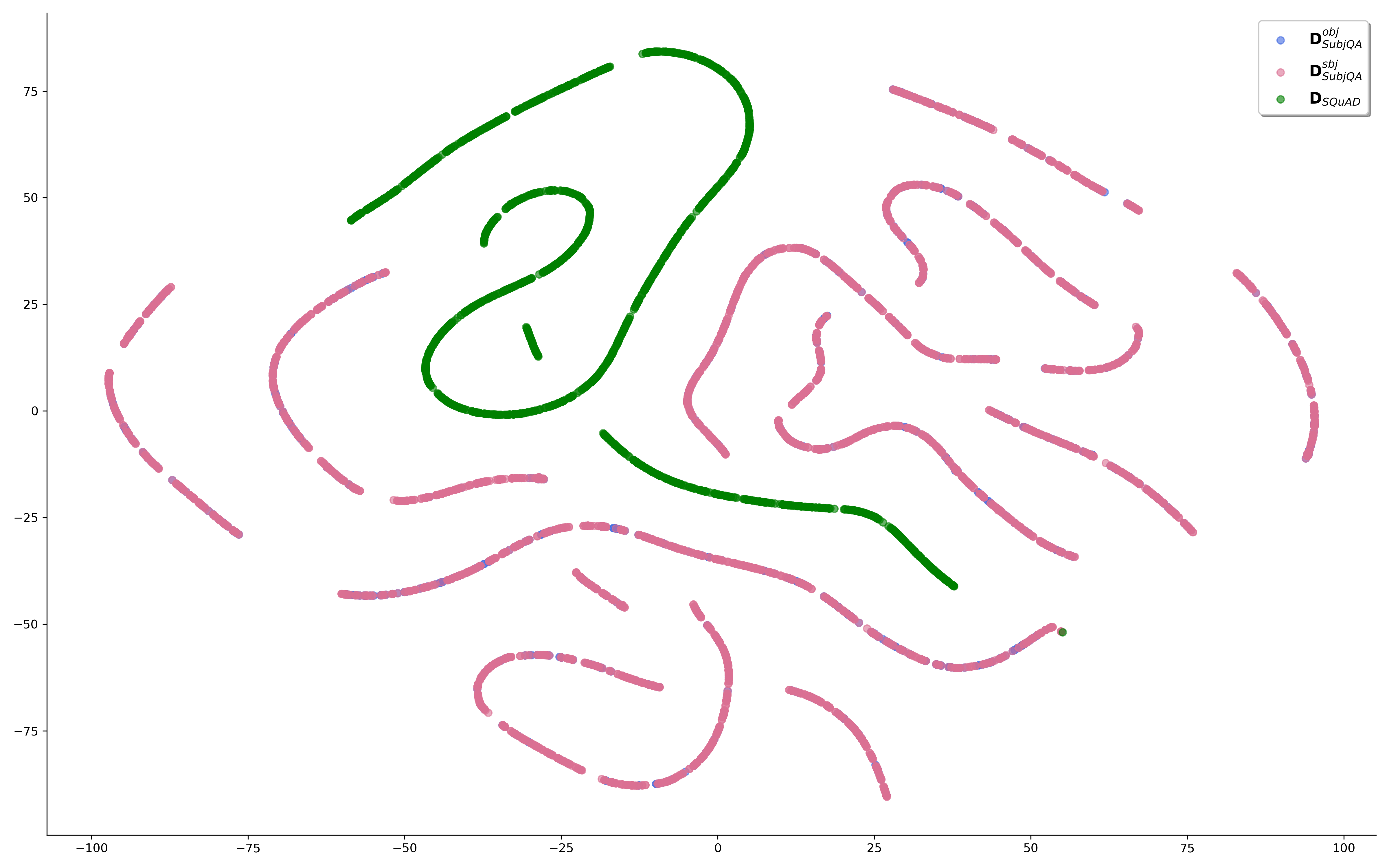}
    \caption{$(\mathbf{q, c})$ - $95\% \: \sigma^{2}$ retained in PCA}
\end{subfigure}%
\begin{subfigure}{.52\textwidth}
    \centering
    \captionsetup{justification=centering}
    \includegraphics[width=0.99\textwidth]{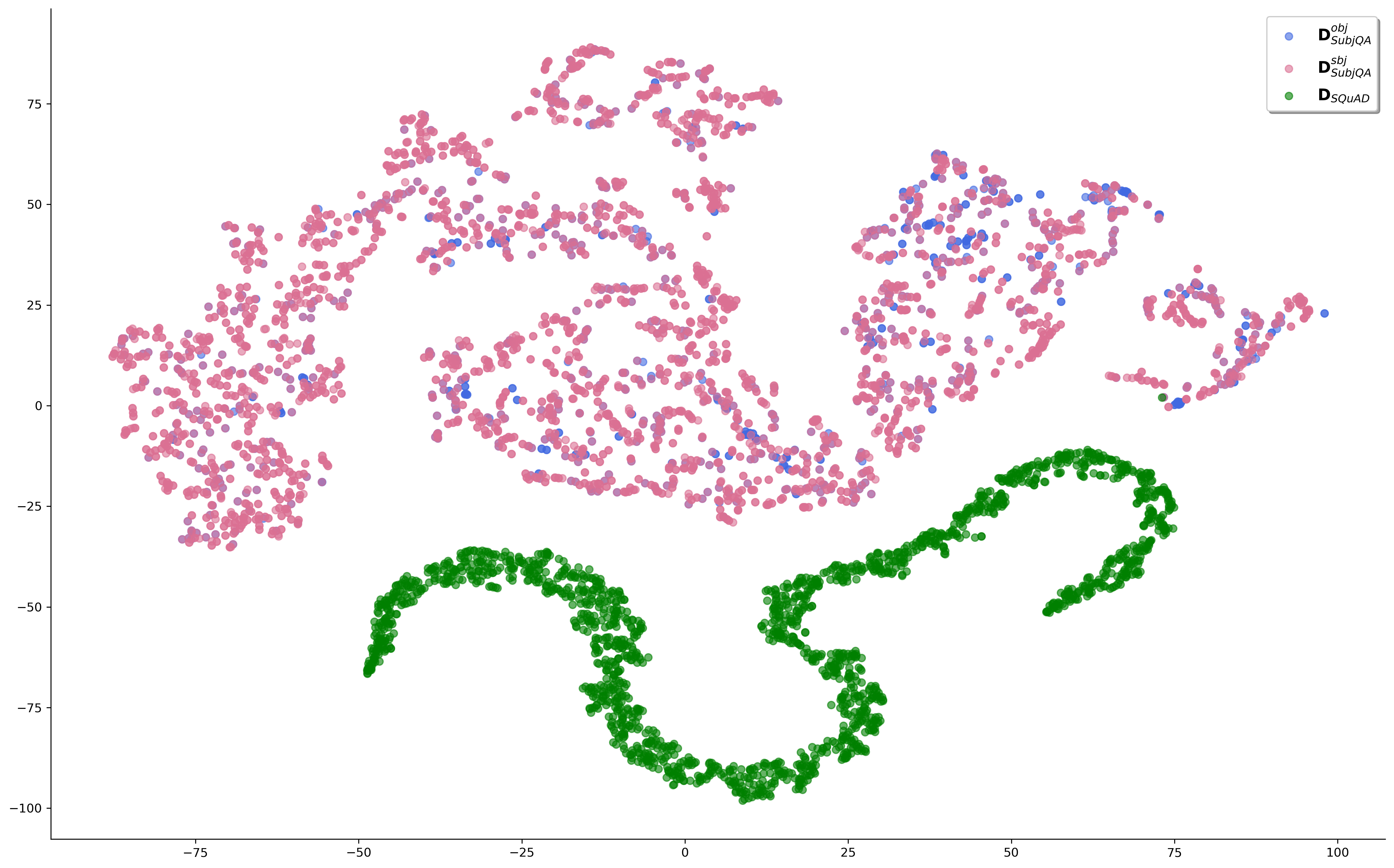}
    \caption{$(\mathbf{q, c})$ - $99\% \: \sigma^{2}$ retained in PCA}
\end{subfigure}
\caption{\textsc{BERT}'s feature representations of $(\mathbf{q, a})$ and $(\mathbf{q, c})$ sequences in latent space - iteratively optimized during multi-way classification - projected onto 2D via PCA and t-SNE respectively. The upper and lower row display question - answer sentence pairs, that is $(\mathbf{q, a})$ sequences, and question - context sentence pairs, that is $(\mathbf{q, c})$ sequences, respectively. Each sentence pair was represented through \textsc{BERT}'s semantic representation of the entire sequence reflected in the special \texttt{[CLS]} token. Feature representations were first transformed via PCA into $d$-dimensional space (depending on $\sigma^{2}$) and then projected onto 2D via t-SNE.  Pink: \textcolor{pink}{subjective} questions $\in \mathbf{D}_{SubjQA}$. Blue: \textcolor{blue}{objective} questions $\in \mathbf{D}_{SubjQA}$. Green: \textcolor{green}{objective} questions $\in \mathbf{D}_{SQuAD}$. Depicted are hidden representations from the model's last (6\textsuperscript{th}) layer.}
\label{fig:sbj_class_multi_tsne}
\end{figure}

To better understand why this is the case, I first projected each sentence pair's semantic representation in vector space - which is reflected in the hidden representation w.r.t. \textsc{BERT}'s special \texttt{[CLS]} token - for each sentence pair with Principal Components Analysis (PCA) \cite{DBLP:journals/corr/Shlens14} onto $n$ principal components that either retained 95\% or 99\% of the variance prior to transforming the sentence embeddings into 2D space \footnote{Using the t-Distributed Stochastic Neighbor Embedding (t-SNE) implementation provided by scikit-learn \cite{scikit-learn}}. The former step was performed to both save computational time - t-SNE is an expensive algorithm that leverages stochastic gradient descent (SGD) to iteratively search for a low-dimensional feature space until convergence \cite{vanDerMaaten2008} - and examine differences in the projections dependent on the retained variance. As clearly reflected in \textsc{BERT}'s low dimensional sentence pair projections, questions from \textsc{SQuAD} are clustered in a space that is highly distinguishable from the two other classes, both of which belong to \textsc{SubjQA} (see Figure~\ref{fig:sbj_class_multi_tsne}).

This holds even more so when classifying $(\mathbf{q, c})$ as indicated by the quantitative results (see Figure~\ref{fig:sbj_multi_conf_mats}). The model's feature representations for $(\mathbf{q, c})$ sentence pairs that belong to \textsc{SQuAD} are embedded in a space that is perfectly distinguishable from the two other classes. In those projections, the data points do not even touch one another as opposed to t-SNE plots with respect to $(\mathbf{q, a})$ sentence pairs. However, the model does not appear to differentiate \textsc{SubjQA}'s subjective from \textsc{SubjQA}'s objective questions in latent space whatsoever. This is in line with the class-specific results depicted in the confusion matrices~\ref{fig:sbj_multi_conf_mats}, and hints towards the potential post hoc hypotheses that either none of the questions and corresponding answers in \textsc{SubjQA} is fully objective or subjective and thus cannot be modeled with discrete values, or \textsc{SubjQA} is in general a relatively subjective dataset. I will elaborate this in more detail in Section~\ref{section:discussion}.

\newpage

\section{Multi-task Learning for Question Answering}

\paragraph{Knowing about subjectivity but being dataset agnostic} One could argue that in the (multi-way) subjectivity classification task the model somehow learned to distinguish SubjQA from SQuAD. Thus, subjective and objective questions that belong to \textsc{SubjQA} were projected into the same latent space but examples that belong to \textsc{SQuAD} were separated from the latter two. To account for this potential objection, I've implemented a model that was simultaneously trained on three tasks, namely QA, subjectivity, and dataset classification, of which the latter classification task was performed in an adversarial manner with a Gradient Reversal Layer (GRL) \cite{ganin2014unsupervised} in between \textsc{BERT} and the respective task-specific output layer(s) (see Section~\ref{method:adversarial} for further details). Hence, the model was optimized to correctly classify questions into subjective vs. objective as was done in the previous setting - although not in a multi-way but binary fashion with $1$ representing the subjective and $0$ the objective class - but at the same time trained to not know anything about the source of the sentence pair example, that is being dataset agnostic.
%\newpage

\begin{figure}[h!]
\centering
\begin{subfigure}{.42\textwidth}
    \centering
    \captionsetup{justification=centering}
    \includegraphics[width=0.95\textwidth]{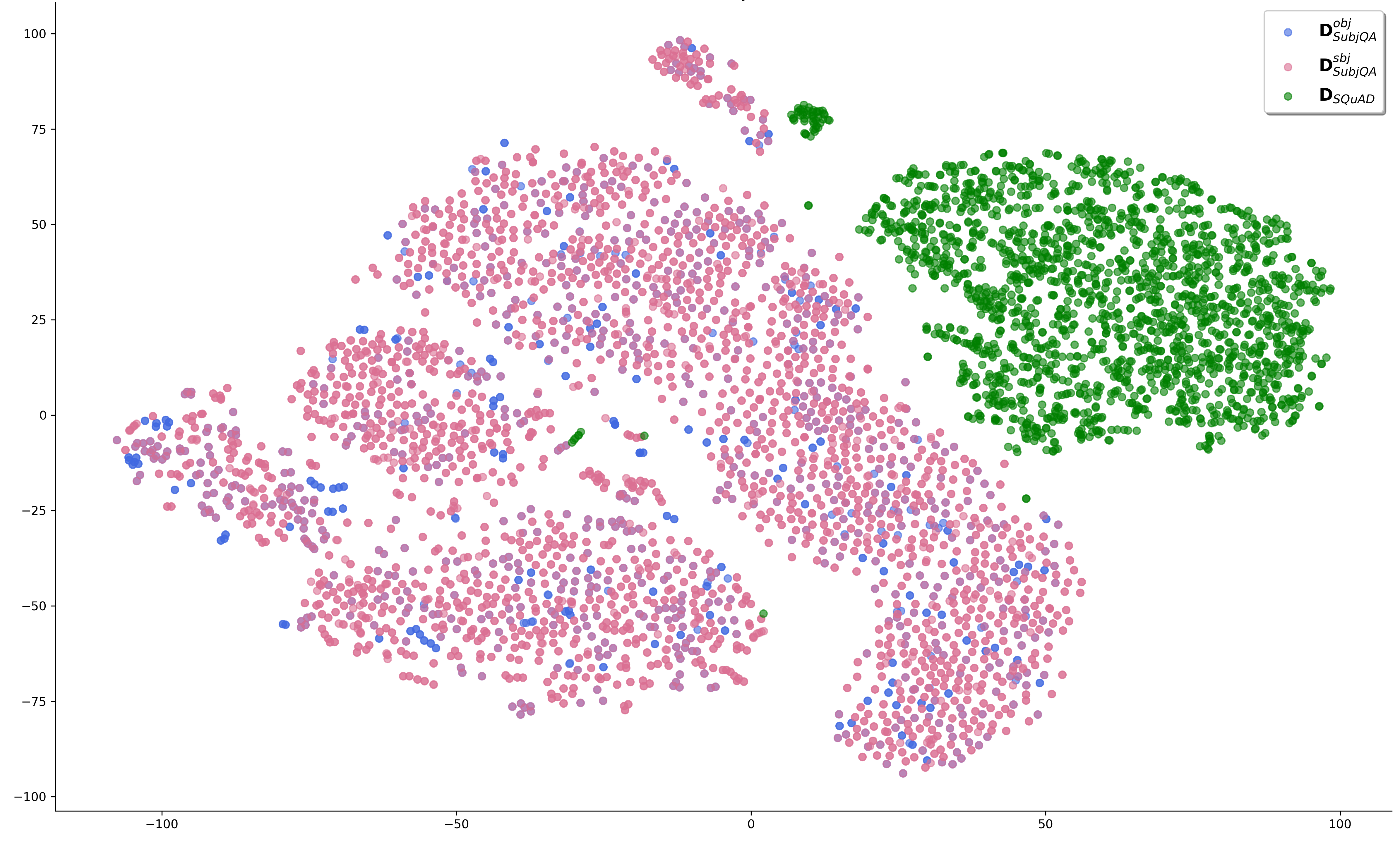}
    \caption[short]{Layer 1}
\end{subfigure}%
\begin{subfigure}{.42\textwidth}
    \centering
    \captionsetup{justification=centering}
    \includegraphics[width=0.95\textwidth]{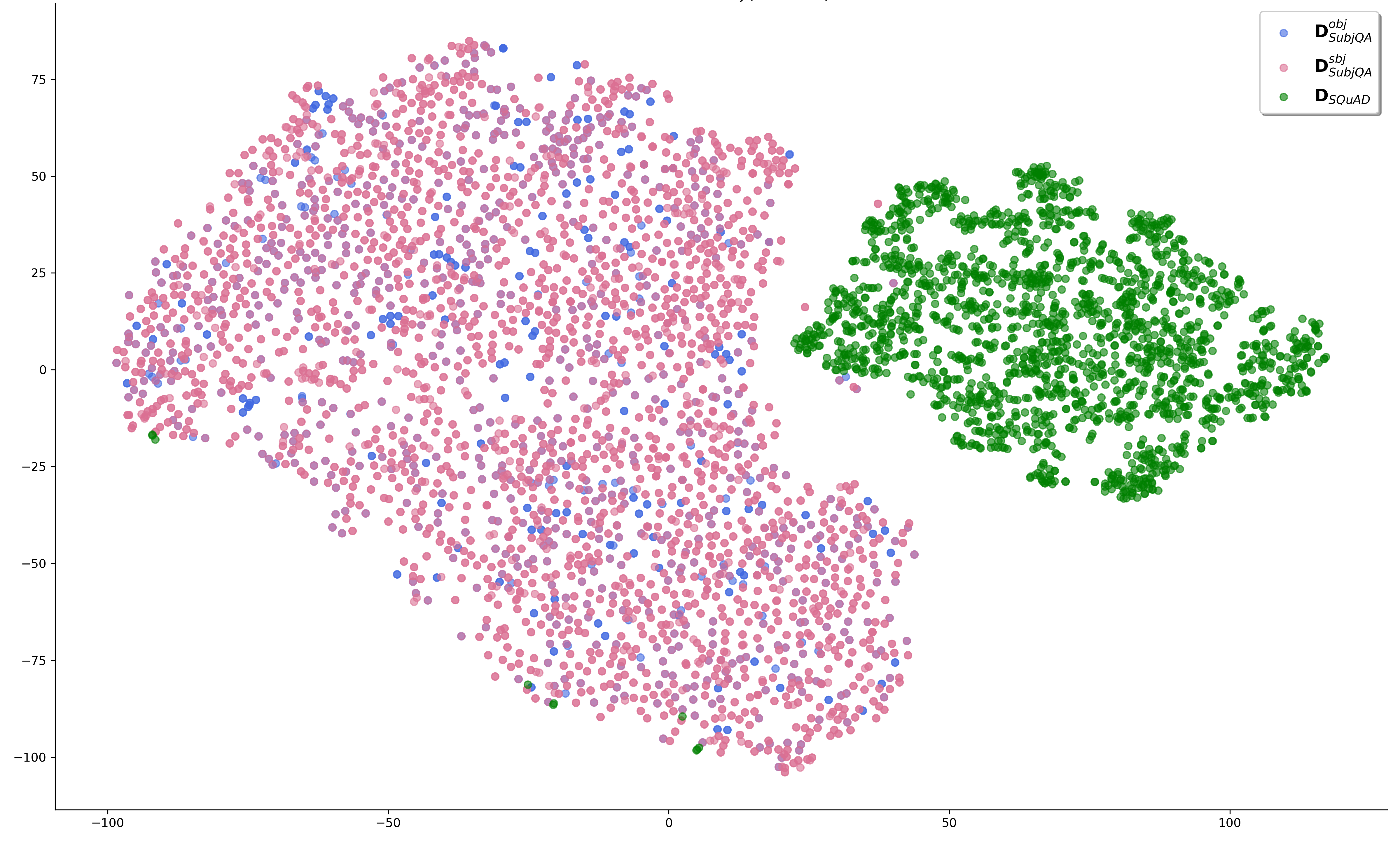}
    \caption[short]{Layer 2}
\end{subfigure}
\begin{subfigure}{.42\textwidth}
    \centering
    \captionsetup{justification=centering}
    \includegraphics[width=0.95\textwidth]{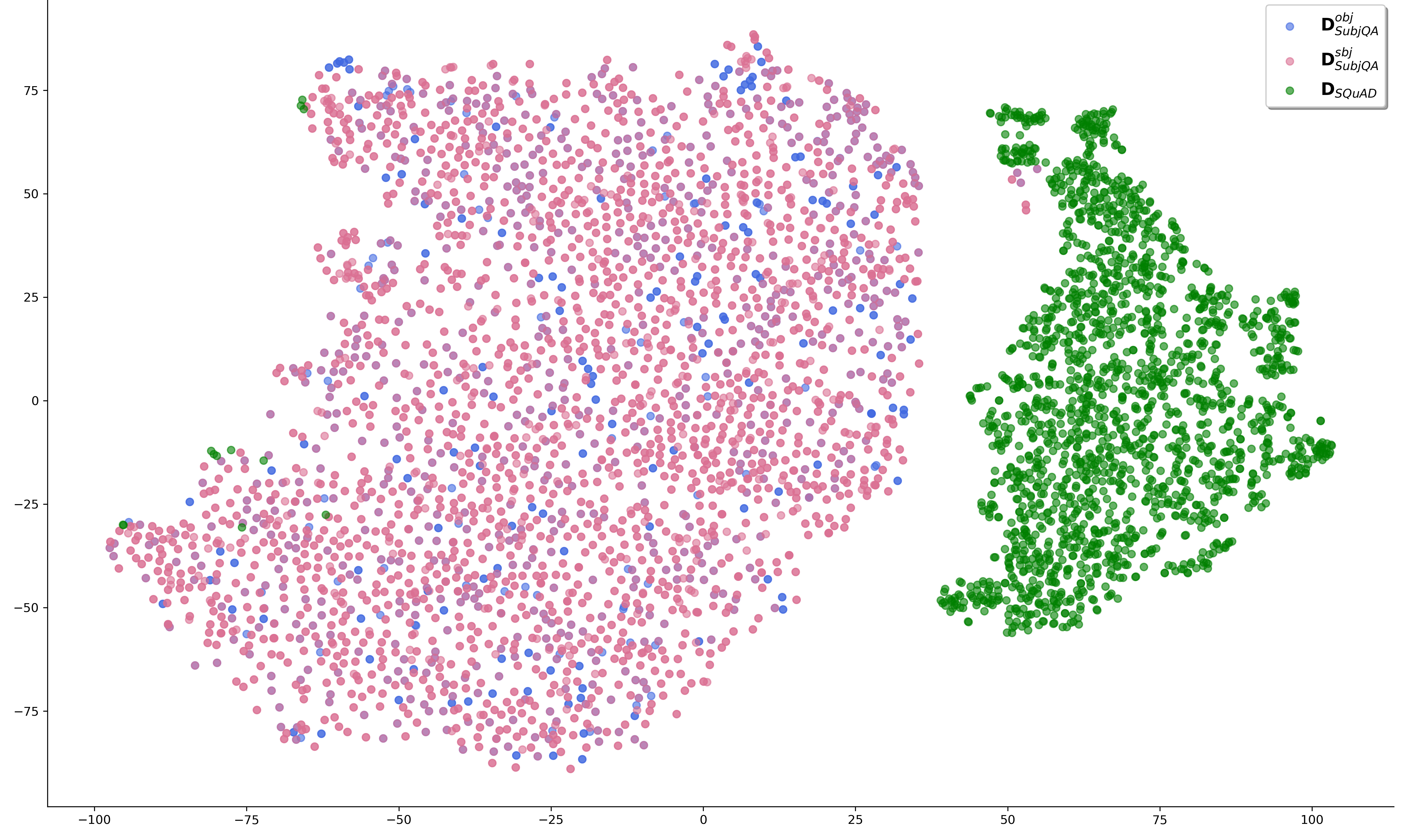}
    \caption[short]{Layer 3}
\end{subfigure}%
\begin{subfigure}{.42\textwidth}
    \centering
    \captionsetup{justification=centering}
    \includegraphics[width=0.95\textwidth]{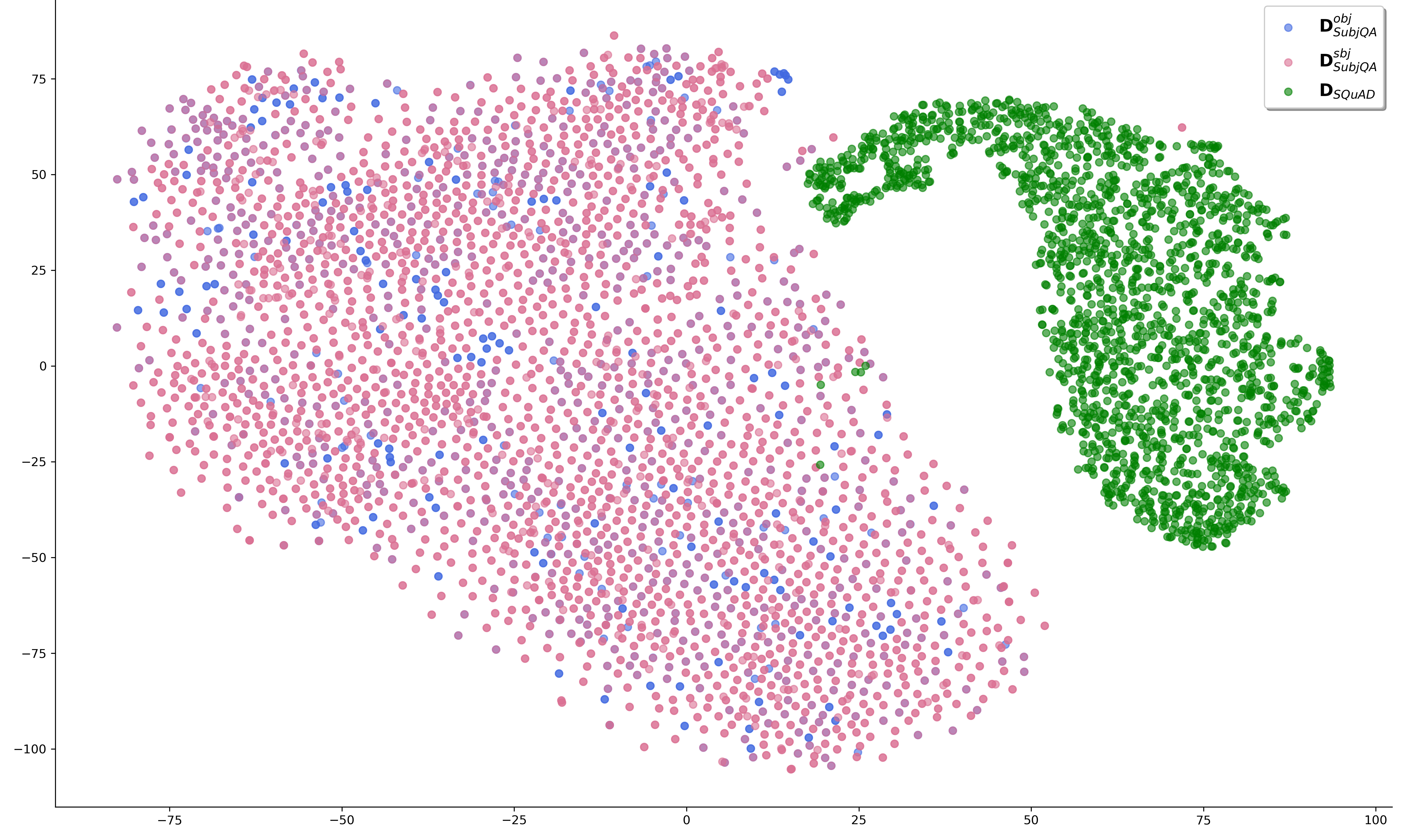}
    \caption[short]{Layer 4}
\end{subfigure}
\begin{subfigure}{.42\textwidth}
    \centering
    \captionsetup{justification=centering}
    \includegraphics[width=0.95\textwidth]{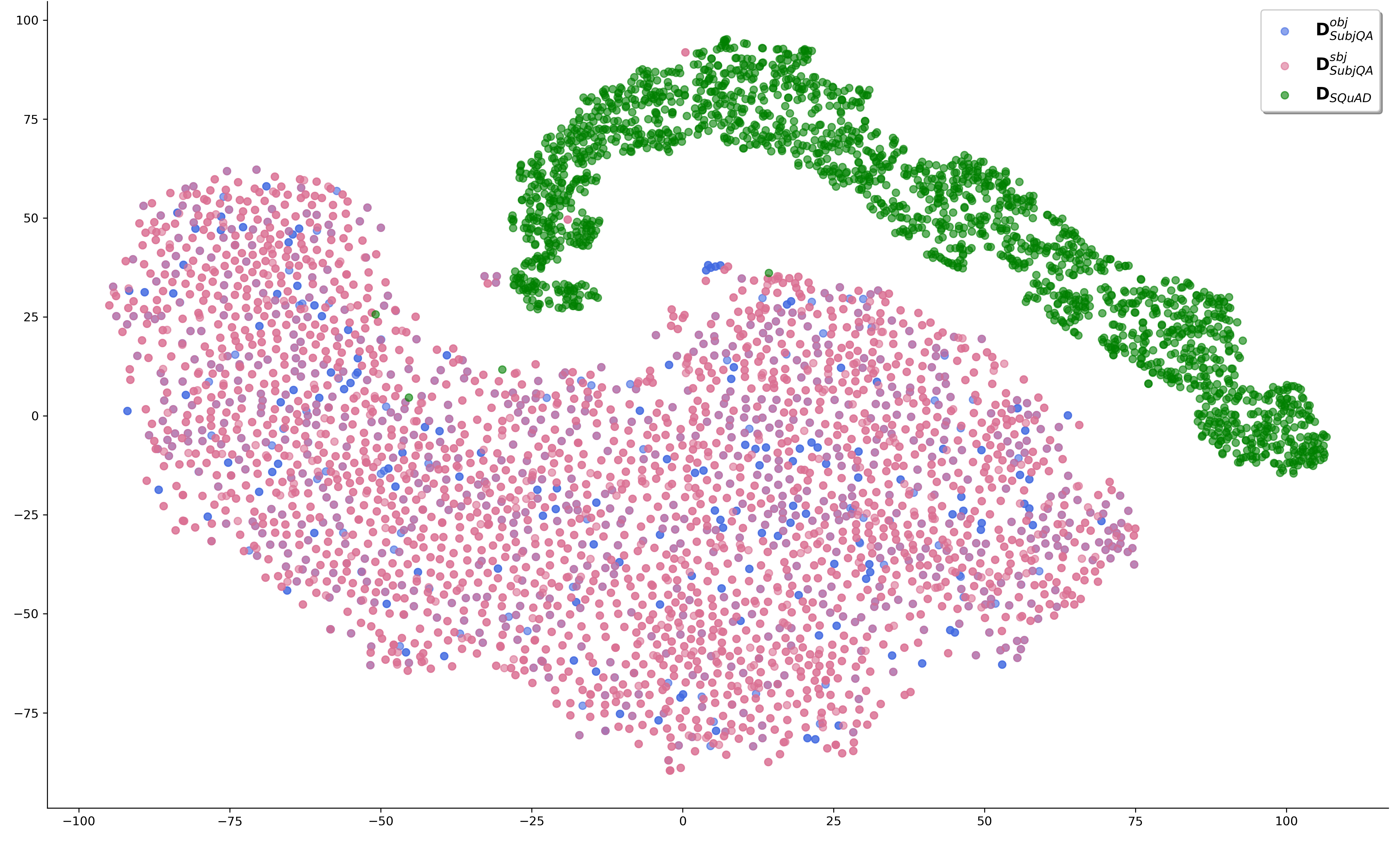}
    \caption[short]{Layer 5}
\end{subfigure}%
\begin{subfigure}{.42\textwidth}
    \centering
    \captionsetup{justification=centering}
    \includegraphics[width=0.95\textwidth]{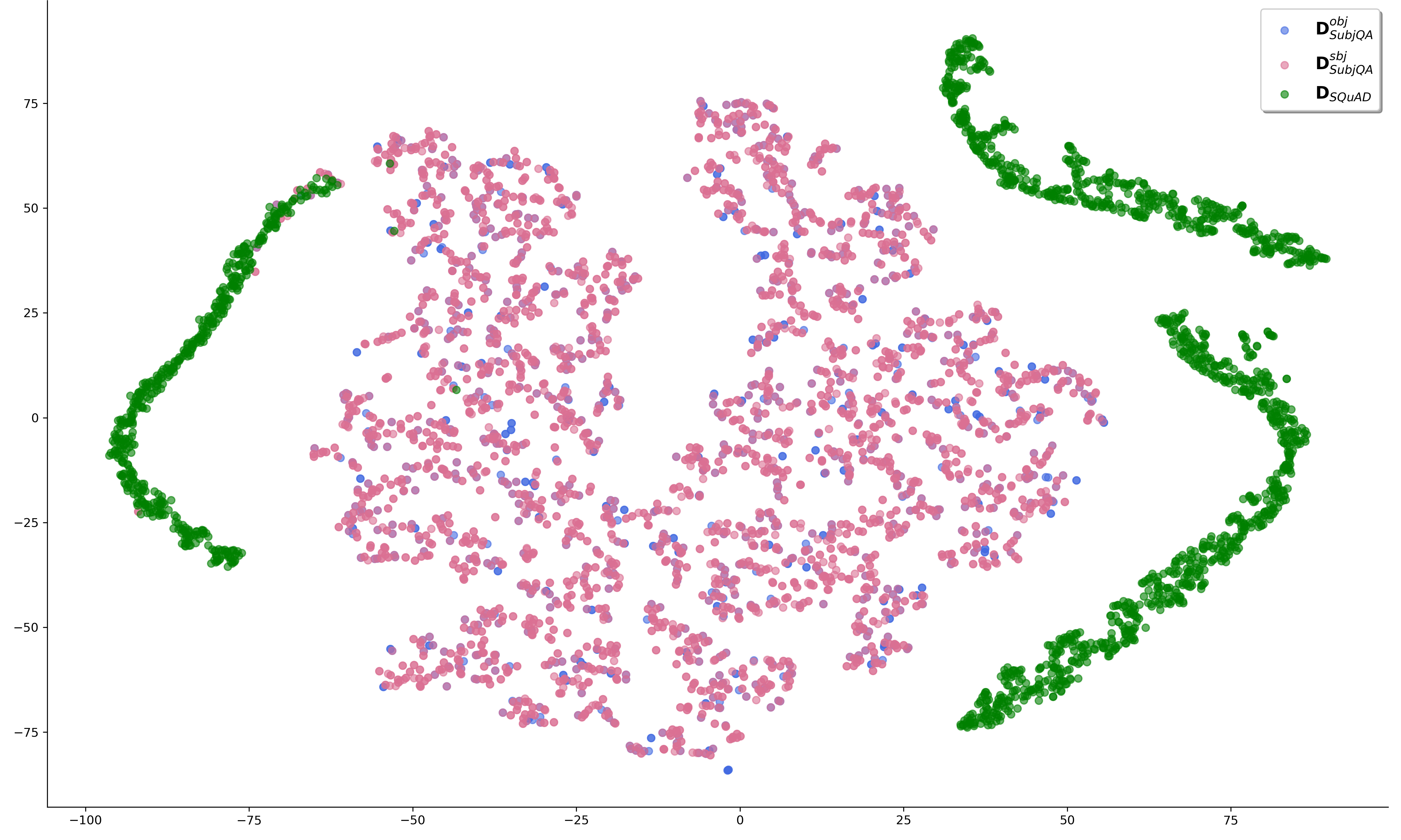}
    \caption[short]{Layer 6}
\end{subfigure}
\caption{Dataset agnostic MTL model fine-tuned on $\mathbf{D}_{SubjQA} \: \cup \: \mathbf{D}_{SQuAD}$. Depicted are the model's hidden representation w.r.t. \textsc{BERT}'s special \texttt{[CLS]} token at each layer for every sentence pair example in the combined test set $\mathbf{D}_{comb}$. Feature representations across the different layers are represented from left-to-right and top-to-bottom in the usual bottom-up representational hierarchy starting at the first (1\textsuperscript{st}) (top-left) and stopping at the last (6\textsuperscript{th}) (bottom-right) layer. Retained variance in PCA: $99\%$. Pink: \textcolor{pink}{subjective} questions $\in \mathbf{D}_{SubjQA}$. Blue: \textcolor{blue}{objective} questions $\in \mathbf{D}_{SubjQA}$. Green: \textcolor{green}{objective} questions $\in \mathbf{D}_{SQuAD}$.}
\label{fig:hidden_reps_cls_sbj_class_ds_agnostic}
\end{figure}
\newpage

The MTL training was performed in an \textsc{alternating} batch setting, where examples for subjectivity classification were drawn from batches that consisted of $(\mathbf{q, a})$ sequences, whereas the model received $(\mathbf{q, c})$ sequences as inputs for all other tasks. This was done, to ensure that the model does not learn the subjectivity classification task dependent on contextual features but question and answer tokens alone. Moreover, \hyperref[section:sbj_class_results]{\textbf{quantitative results}} have shown that solely classifying $(\mathbf{q, a})$ sequence pairs yields better results than classifying $(\mathbf{q, c})$ sequences (see Table~\ref{tab:sbjclass_main_results}). Note that the subjectivity classification task was optimized as a binary and not as a multi-way classification task. The objective class was synthetically split into \textsc{SQuAD} and \textsc{SubjQA} post hoc. This time, the model's hidden representations are depicted for each layer to investigate whether differences between objective and subjective questions occur exclusively at later or even at earlier stages of the neural network.

As clearly indicated by the model's hidden representations projected into $\mathbf{R}^{2}$ (see Figure~\ref{fig:hidden_reps_cls_sbj_class_ds_agnostic}), the differences in the linguistic signals between objective and subjective questions $\in \mathbf{D}_{SubjQA}$ appear to be too marginal to be distinguished from one another. Objective questions that belong to \textsc{SQuAD}, however, are embedded in a notably different part of the vector space. The latter becomes apparent even in the 1\textsuperscript{st} layer of the network (see Figure~\ref{fig:hidden_reps_cls_sbj_class_ds_agnostic} a). This indicates that the model is simply not able to differentiate between objective and subjective questions within \textsc{SubjQA}, even if the model is trained adversarially to be agnostic concerning the source of the data point, but can easily separate objective questions that belong to \textsc{SQuAD} from any question that belongs to \textsc{SubjQA} in latent space. 
%\newpage

\section{Sequential Transfer for Question Answering}

\paragraph{Inspecting sequentially transferred representations} To further investigate into the inability of the model to distinguish subjective from objective questions within \textsc{SubjQA}, I examined the hidden representations of a sequential transfer model that was sequentially trained on all tasks, namely QA, context-domain and subjectivity classification, until convergence (see Section~\ref{method:seq_transfer} for implementation details). To ensure that the model performed subjectivity classification prior to QA, the inspected model was firstly optimized on context-domain classification, followed by subjectivity classification and QA. Similarly to the MTL model, subjectivity classification was performed with batches that contained $(\mathbf{q, a})$ sequences to ensure that the classification is done solely with respect to question and answer tokens respectively and not interfered by linguistic signals of the context.  

\begin{figure}[h!]
\centering
\begin{subfigure}{.53\textwidth}
    \centering
    \captionsetup{justification=centering}
    \includegraphics[width=0.95\textwidth]{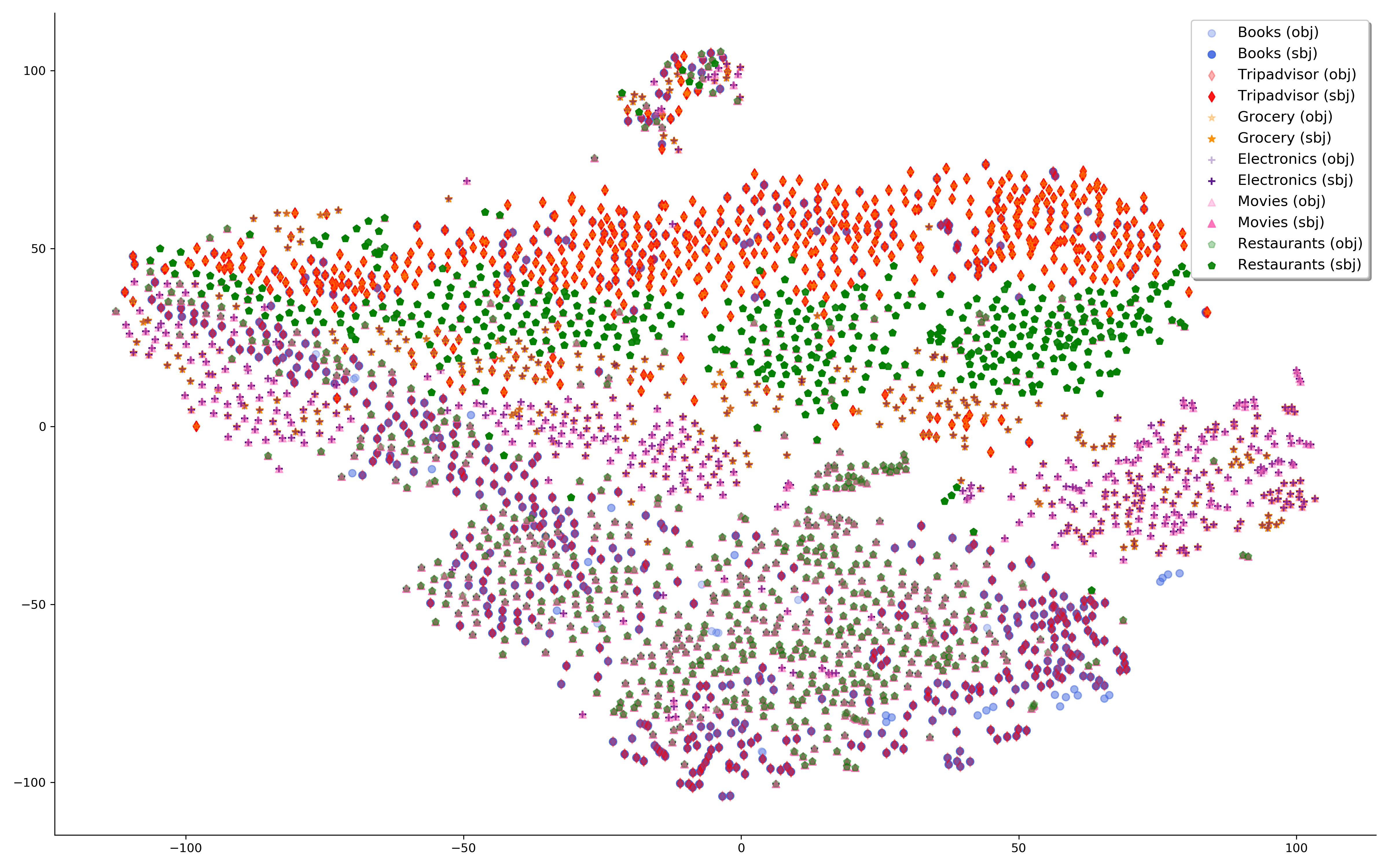}
    \caption[short]{Layer 1}
\end{subfigure}%
\begin{subfigure}{.53\textwidth}
    \centering
    \captionsetup{justification=centering}
    \includegraphics[width=0.95\textwidth]{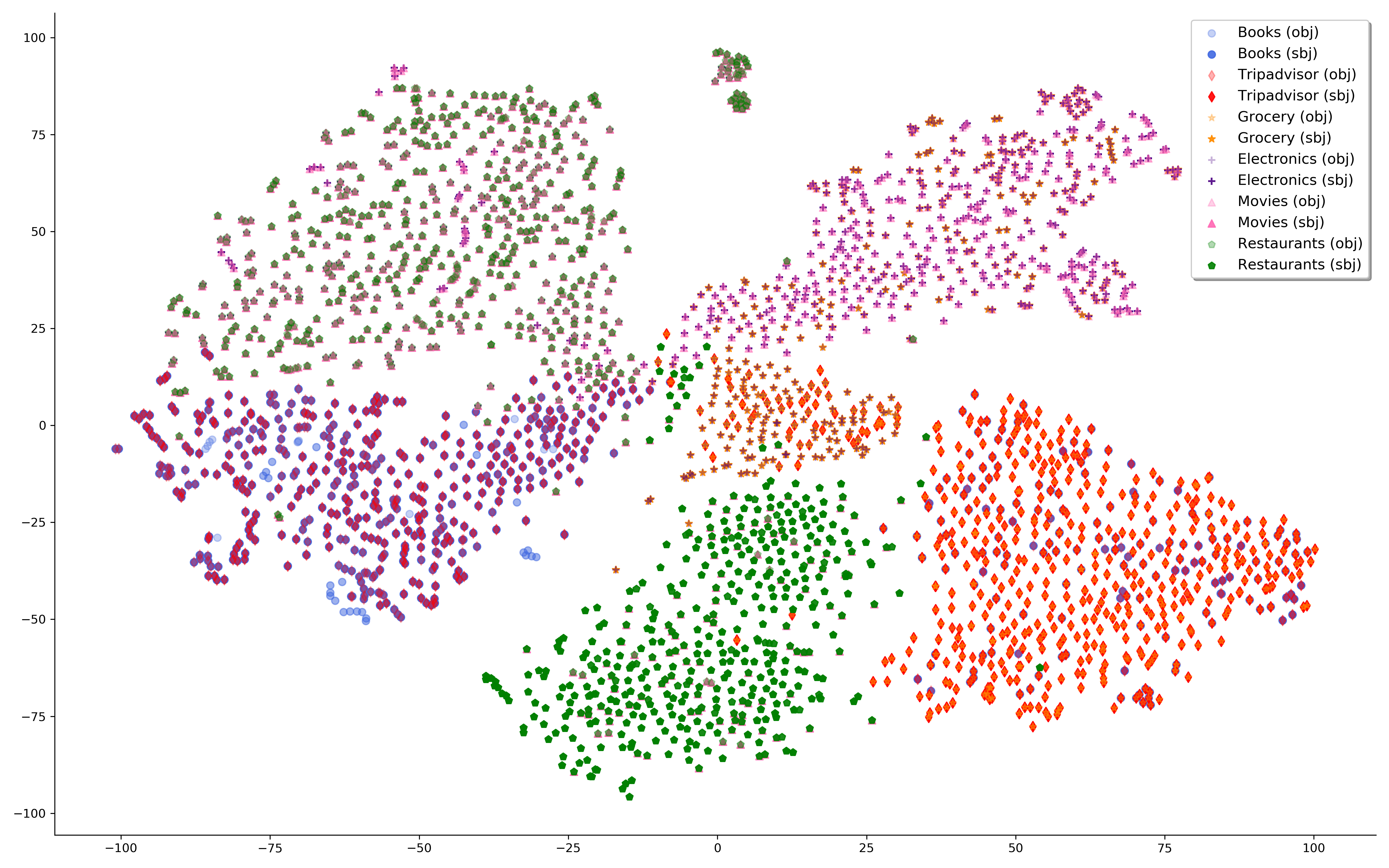}
    \caption[short]{Layer 2}
\end{subfigure}
\begin{subfigure}{.53\textwidth}
    \centering
    \captionsetup{justification=centering}
    \includegraphics[width=0.95\textwidth]{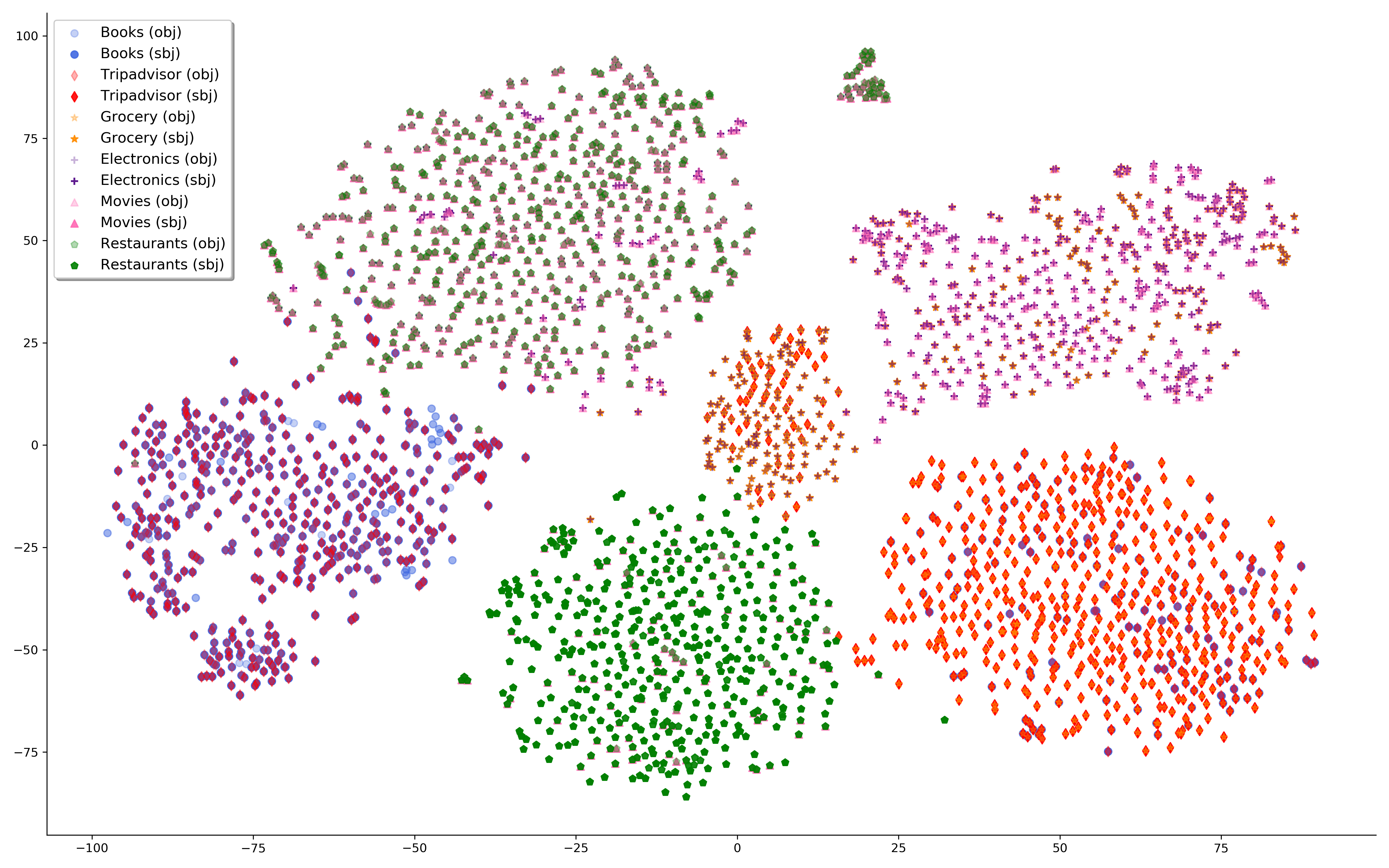}
    \caption[short]{Layer 3}
\end{subfigure}%
\begin{subfigure}{.53\textwidth}
    \centering
    \captionsetup{justification=centering}
    \includegraphics[width=0.95\textwidth]{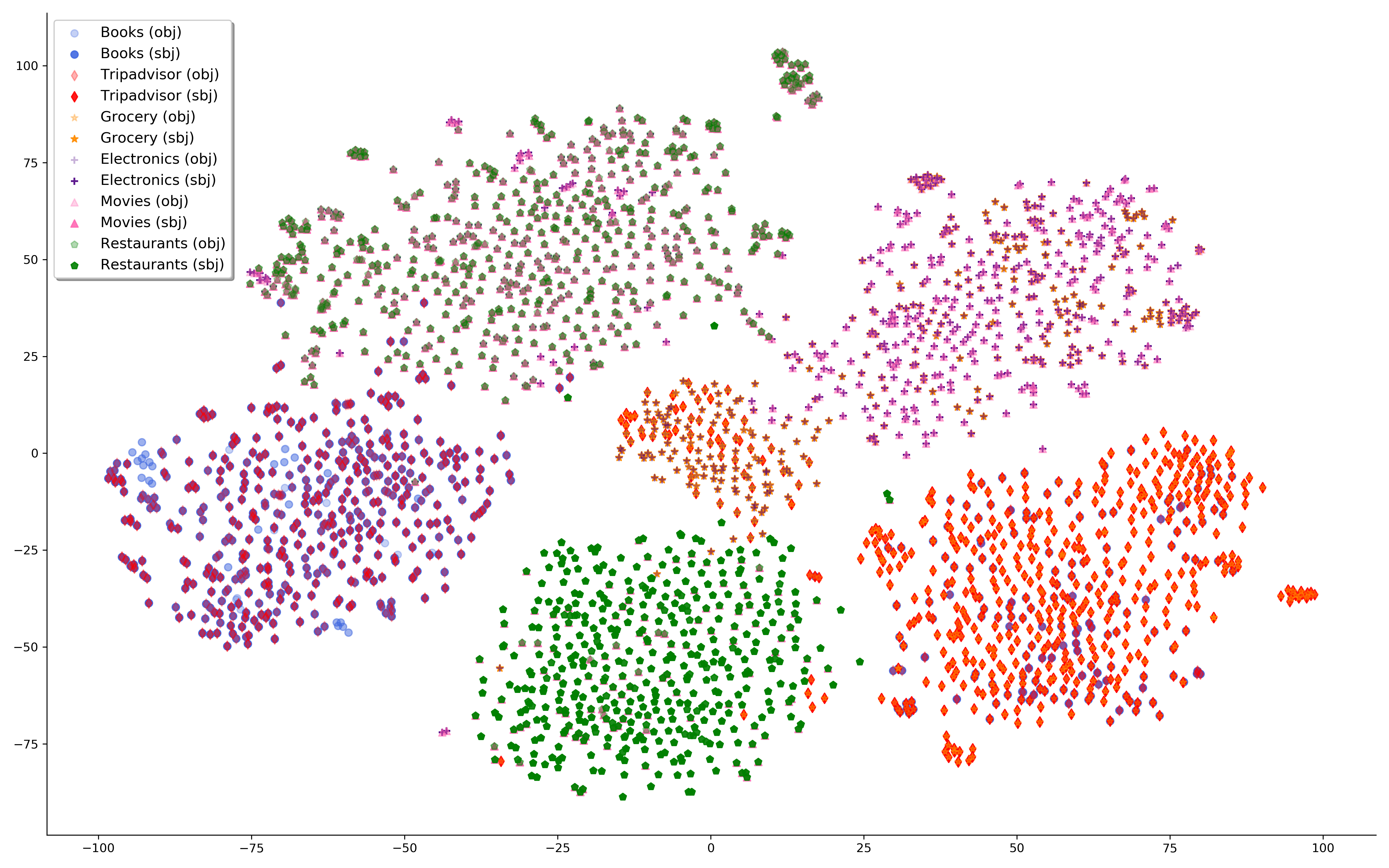}
    \caption[short]{Layer 4}
\end{subfigure}
\begin{subfigure}{.53\textwidth}
    \centering
    \captionsetup{justification=centering}
    \includegraphics[width=0.95\textwidth]{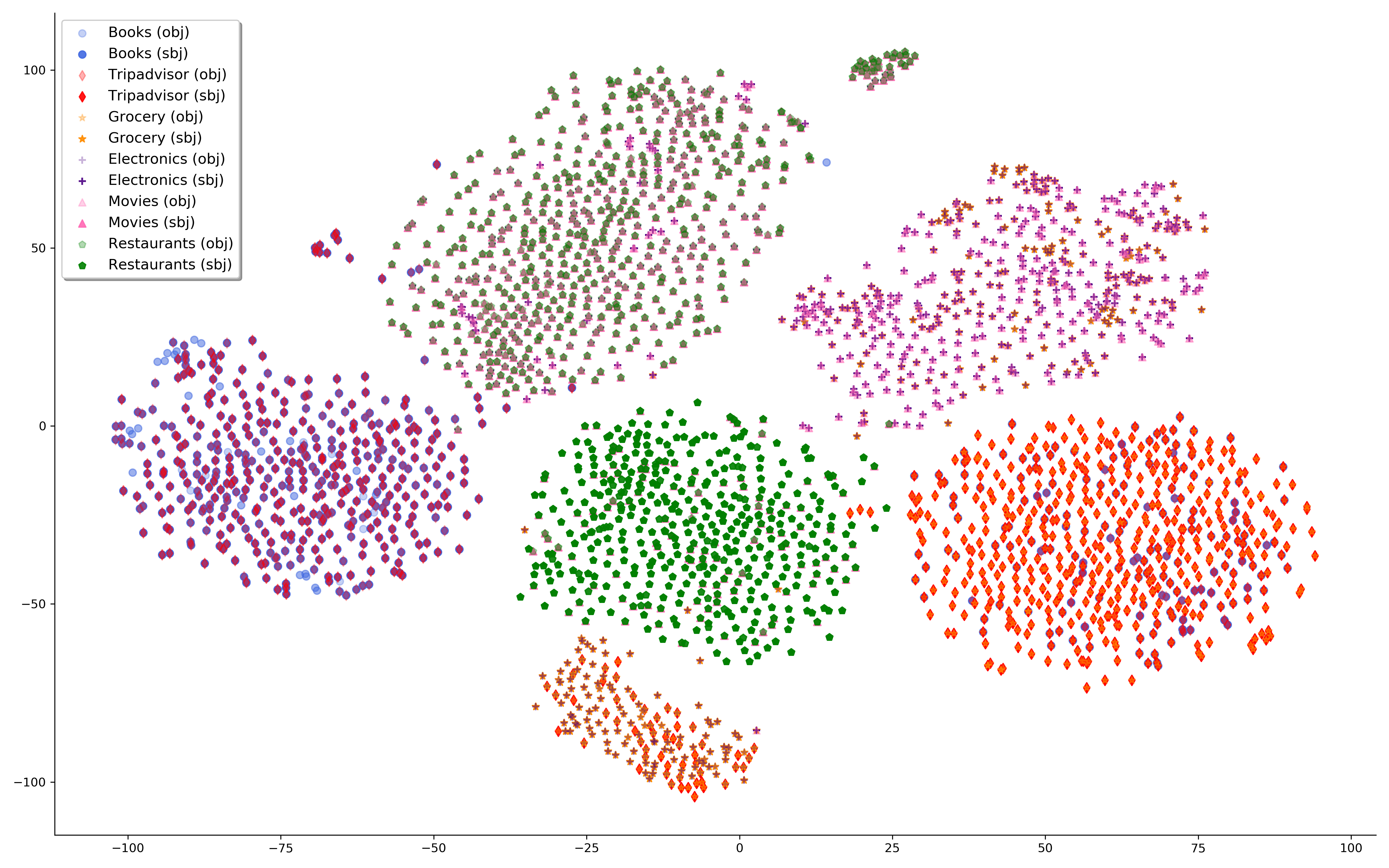}
    \caption[short]{Layer 5}
\end{subfigure}%
\begin{subfigure}{.53\textwidth}
    \centering
    \captionsetup{justification=centering}
    \includegraphics[width=0.95\textwidth]{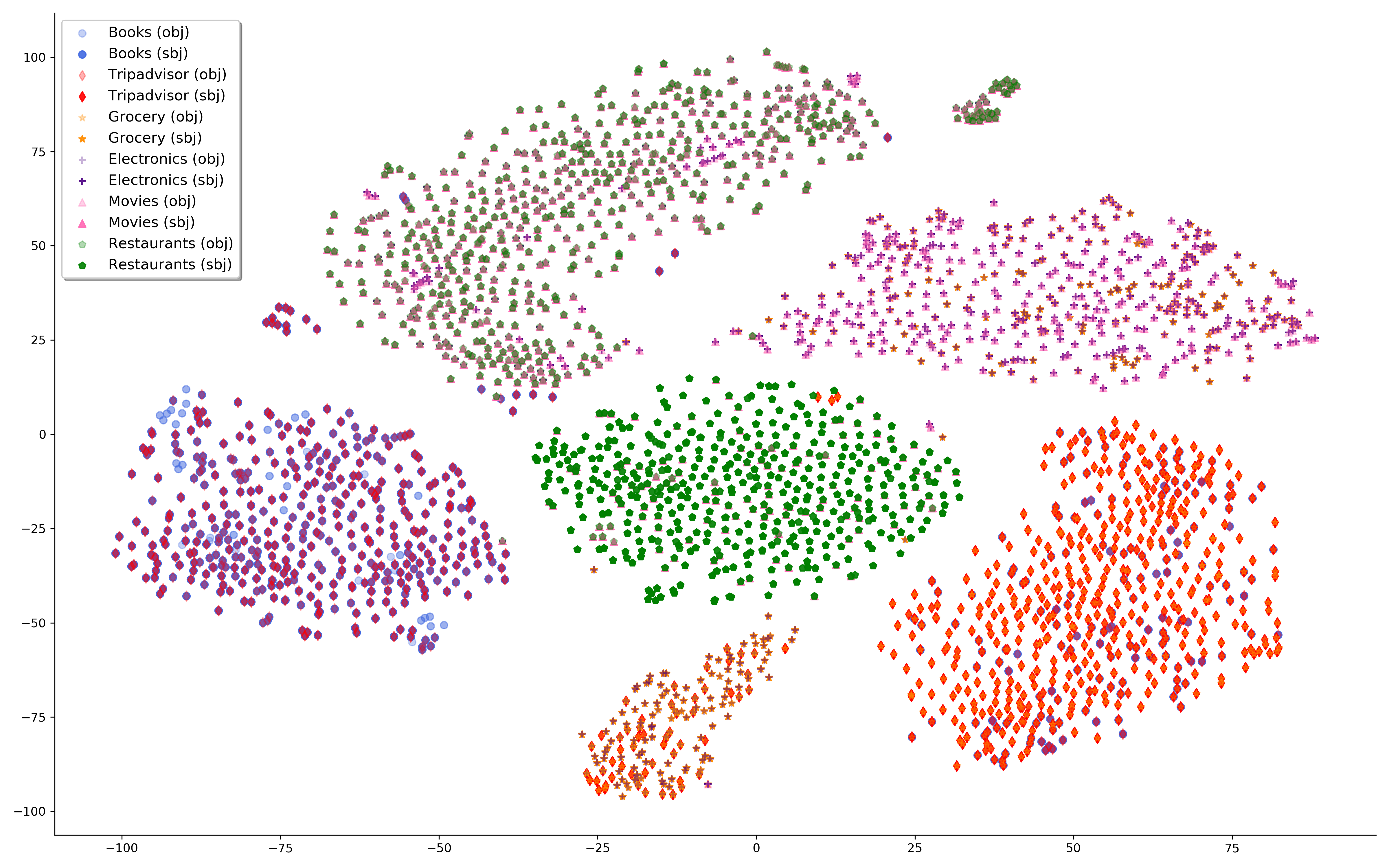}
    \caption[short]{Layer 6}
\end{subfigure}
\caption{Sequential transfer model fine-tuned exclusively on $\mathbf{D}_{SubjQA}$. Depicted are the model's hidden representation w.r.t. \textsc{BERT}'s special \texttt{[CLS]} token at each layer for every sentence pair example in SubjQA's test set $\mathbf{D}_{comb}$. Feature representations across the different layers are represented from left-to-right and top-to-bottom in the usual bottom-up representational hierarchy starting at the first (1\textsuperscript{st}) (top-left) and stopping at the last (6\textsuperscript{th}) (bottom-right) layer. Blue: \textcolor{blue}{\texttt{books}}. Red: \textcolor{red}{\texttt{tripadvisor}}. Dark orange: \textcolor{orange}{\texttt{grocery}}. Indigo: \texttt{electronics}. Pink: \textcolor{pink}{\texttt{movies}}. Green: \textcolor{green}{\texttt{restaurants}}. Within each domain, colors with higher intensity represent \textbf{subjective} and lower intensity colors \textcolor{gray}{objective} questions. Retained variance in PCA: $99\%$.}
\label{fig:hidden_reps_cls_seq_transfer}
\end{figure}

As can be inferred from the model's feature representations projected into $\mathbf{R}^{2}$ (see Figure~\ref{fig:hidden_reps_cls_seq_transfer}), the model could to a large extent separate the domains from each other - although overlaps between certain domains can be observed -, but could on the other hand not distinguish between subjective and objective questions within domains. Recall that context-domain classification was performed as the first task in the task sequence $\mathbf{T} = [T, T', T'']$, that is prior to subjectivity classification. Since context-domain clusters are visible in the 2D projections even after the model converged on all three tasks sequentially (see Figure~\ref{fig:hidden_reps_cls_seq_transfer}), one may draw two (alternative) conclusions that are not mutually exclusive.

Firstly, the linguistic signals extracted from review domains are both strong enough to be retained after fine-tuning on two other tasks and more notable than signals extracted from the subjectivity classification part. Secondly, $(\mathbf{q, c})$ sentence pairs are distinguishable due to their context-domains but not because of their difference in subjectivity levels labeled through human crowd workers (see Figure~\ref{fig:subj_levels_domains} for a depiction of the subjectivity level distributions across domains). The latter indicates what has been observed in the sections above, namely that the difference in linguistic signals between objective and subjective questions in SubjQA appears to be too weak to be modeled.   
\newpage

\newpage

\section{Error Analysis}

This section is entirely dedicated to the understanding of general error sources concerning QA. To enhance a network's performance and improve its learning, it is crucial to examine \texttt{which} errors a model made, and \texttt{why} those errors happened in the first place.   

\subsection{Question Answering in Vector Space}

Solely inspecting erroneous predictions regarding the question type (e.g., objective vs. subjective) or the corresponding domain (e.g., movies vs. grocery), does not yield insights into \texttt{why} and \texttt{where} along the way a learner made mistakes. Therefore, I've deviated from the usual error analysis, and instead of just showing examples of correct and erroneous predictions in the form natural language text taken a slightly different approach, inspired by one recently published paper \cite{bert-qa-layerwise}. This study has for the first time analyzed BERT's hidden representations after performing QA and thus contributed to a more thorough understanding of the inner-workings of Transformers \cite{DBLP:journals/corr/abs-2002-12327} (see Section~\ref{section:related_work} for more information). Following the approach of \cite{bert-qa-layerwise}, I've investigated the model's hidden representations at each layer for every token in a randomly chosen sentence pair. In so doing, I've projected them - similarly to the visualizations of hidden states concerning the different classes - with PCA and t-SNE from $\mathbf{R}^{768}$ into $\mathbf{R}^{2}$. This layer-wise analysis reveals information about the model's clustering of natural language utterances in latent space at each stage of the model at inference time.

To yield visualizations of hidden states for each token in a word sequence, I've chosen one random sentence pair among the following three sets: correct predictions w.r.t. answerable questions, correct predictions w.r.t. unanswerable questions, erroneous predictions w.r.t. answerable questions. The former and the latter set  reveal particular insights into \texttt{why} a model made a wrong prediction.

\begin{figure}
\centering
\begin{subfigure}{.7\textwidth}
    \centering
    \captionsetup{justification=centering}
    \includegraphics[width=0.99\textwidth]{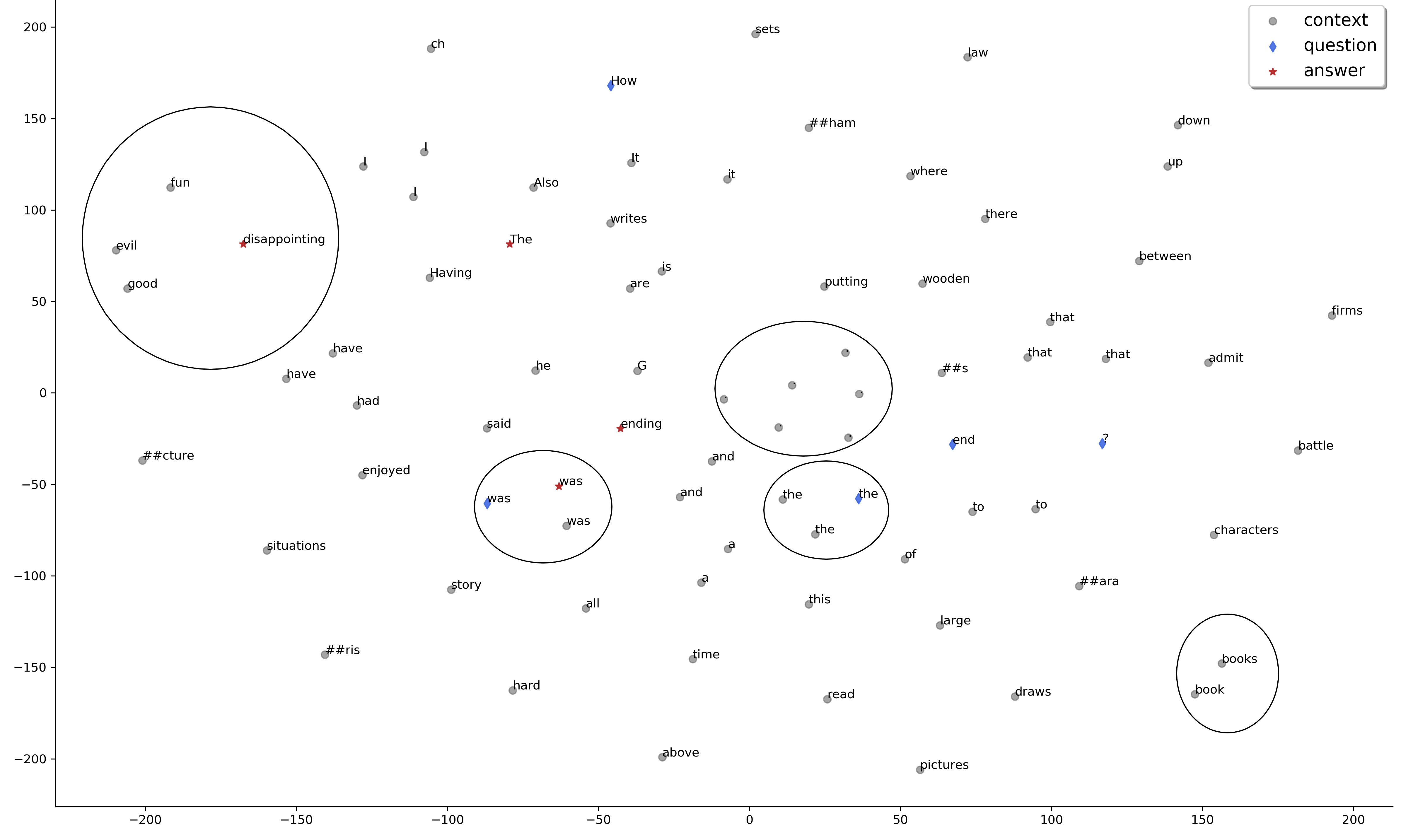}
    \caption[short]{Layer 1}
\end{subfigure}
%\begin{subfigure}{.55\textwidth}
    %\centering
    %\captionsetup{justification=centering}
    %\includegraphics[width=0.99\textwidth]{Figures/qualitative_analysis/hidden_reps_per_token/correct_prediction_answerable_question/tsne_question_context_qa_per_token_99_var_correct_answerable_layer_3_post_processed_no_title.png}
    %\caption[short]{Layer 3}
%end{subfigure}
\begin{subfigure}{.7\textwidth}
    \centering
    \captionsetup{justification=centering}
    \includegraphics[width=0.99\textwidth]{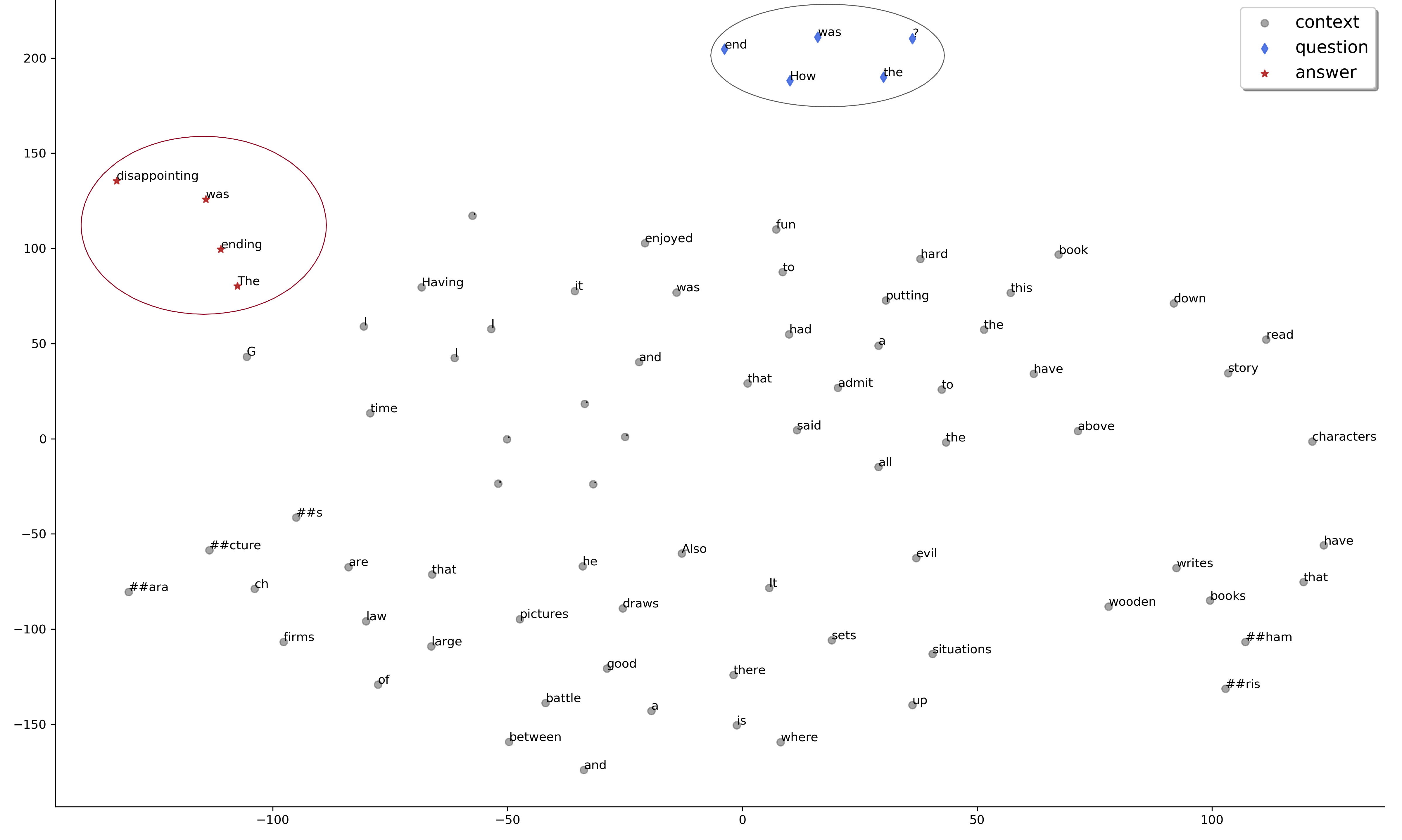}
    \caption[short]{Layer 4}
\end{subfigure}
\begin{subfigure}{.7\textwidth}
    \centering
    \captionsetup{justification=centering}
    \includegraphics[width=0.99\textwidth]{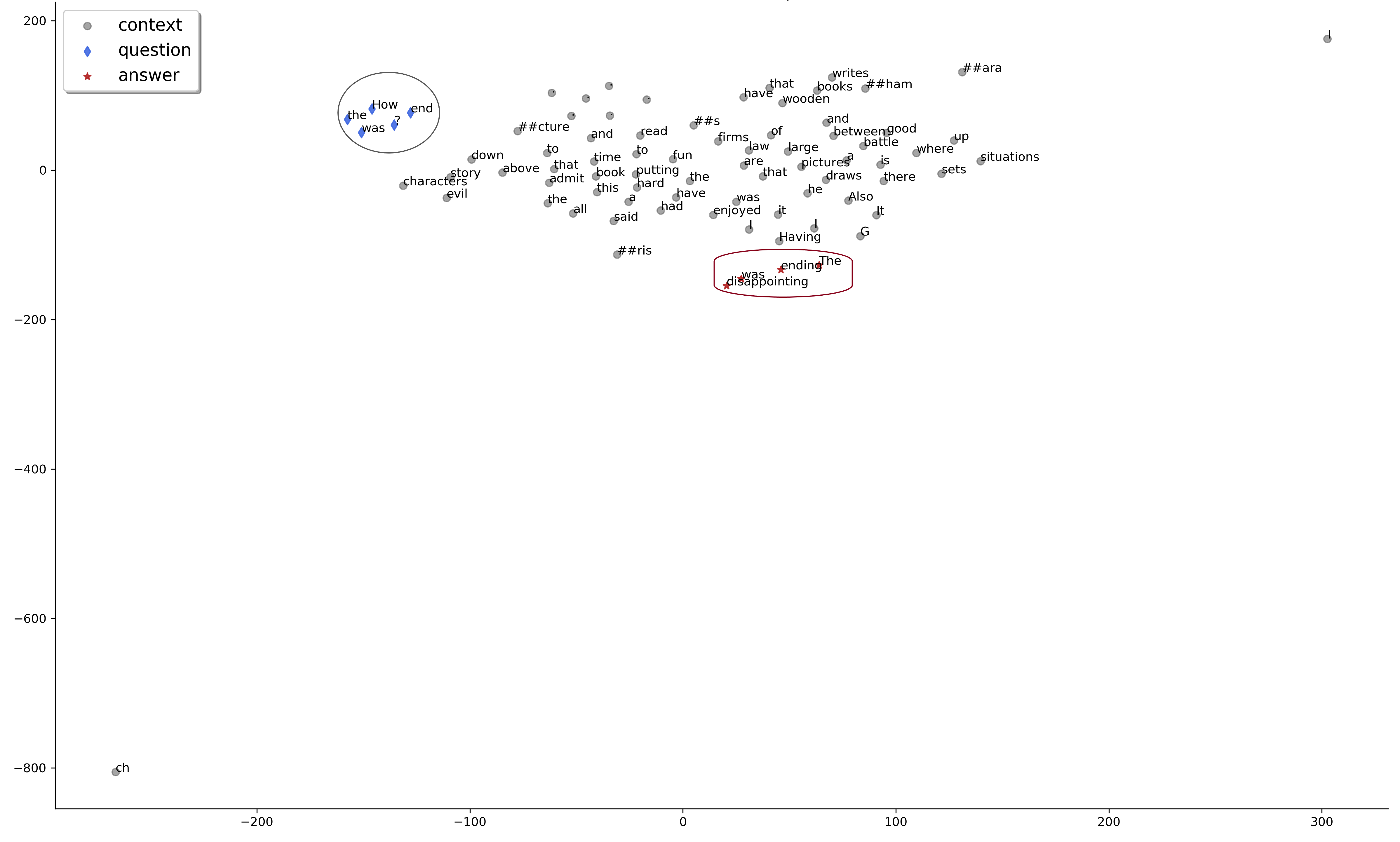}
    \caption[short]{Layer 5}
\end{subfigure}
%\begin{subfigure}{.55\textwidth}
    %\centering
    %\captionsetup{justification=centering}
    %\includegraphics[width=0.99\textwidth]{Figures/qualitative_analysis/hidden_reps_per_token/correct_prediction_answerable_question/tsne_question_context_qa_per_token_99_var_correct_answerable_layer_6_post_processed_no_title.png}
    %\caption[short]{Layer 6}
%\end{subfigure}
\caption{Correct answer-span prediction for an answerable question. Depicted are \textsc{bert}'s hidden representations at different stages of the model from bottom to top (i.e., Layer 1, 4 - 5) projected into $\mathbf{R}^{2}$ for every token in a randomly chosen input sequence (q, c)$_{i}$ among the set of sentence pairs for which the model correctly predicted the answer-span a$_{i}$ w.r.t. q$_{i}$ $\in \mathbf{q}_{answerable}$. Blue diamonds: \textcolor{blue}{question}. Red stars: \textcolor{red}{answer}. Grey circles: \textcolor{gray}{context}. (Q: "\texttt{how} was the end?", A: "The ending was disappointing")}
\label{fig:hidden_reps_correct_pred_answerable_question}
\end{figure}

Figure~\ref{fig:hidden_reps_correct_pred_answerable_question} illustrates hidden states for every token in a randomly chosen sentence pair for which the model correctly answered the questions. Depicted are representations for layers 1, 4, and 5. One can see that low-level features concerning language are depicted in the first layer. Here, tokens that are generally syntactically or semantically similar are clustered together. For instance, definite determiners such as \texttt{the}, indefinite determiners such as \texttt{a} or conjunctions such as \texttt{and} are each clustered in a similar space that is distinguishable from the other linguistic classes. Hence, the model has not yet grasped the high-level concept of question and answer but draws attention to the general features of natural language. In layer 4 the model has projected all tokens that belong to either the question, the answer, or the context into a similar latent space. This indicates an understanding of high-level features with respect to the notion of question and answer. The same holds for layer 5. What is compelling, is that the model clustered both the question and the answer in a separate space from the context even in layer 4. This could indicate that knowledge from later layers might not be necessary to answer the question.

\begin{figure}
\centering
\begin{subfigure}{.7\textwidth}
    \centering
    \captionsetup{justification=centering}
    \includegraphics[width=0.99\textwidth]{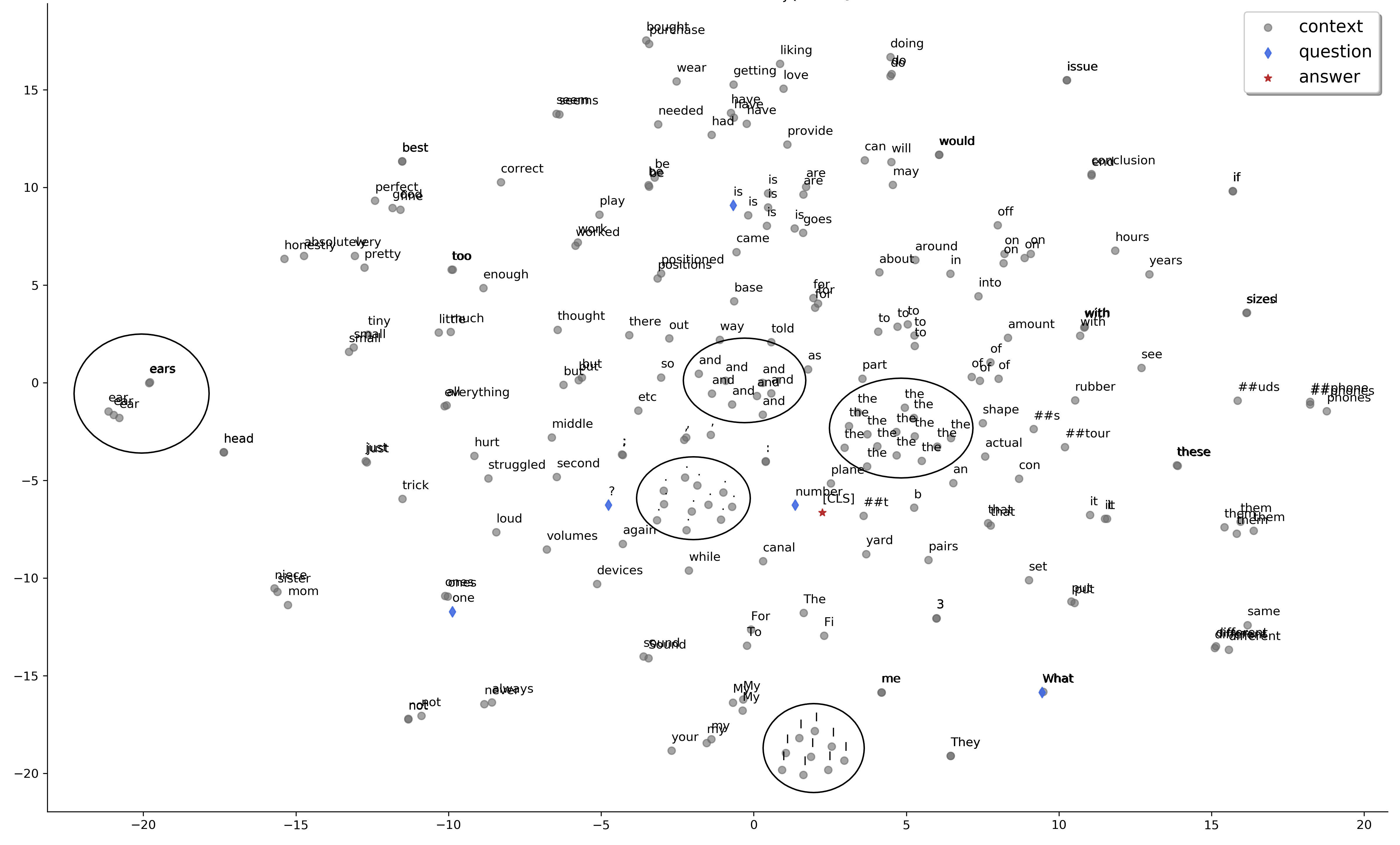}
    \caption[short]{Layer 1}
\end{subfigure}
\begin{subfigure}{.7\textwidth}
    \centering
    \captionsetup{justification=centering}
    \includegraphics[width=0.99\textwidth]{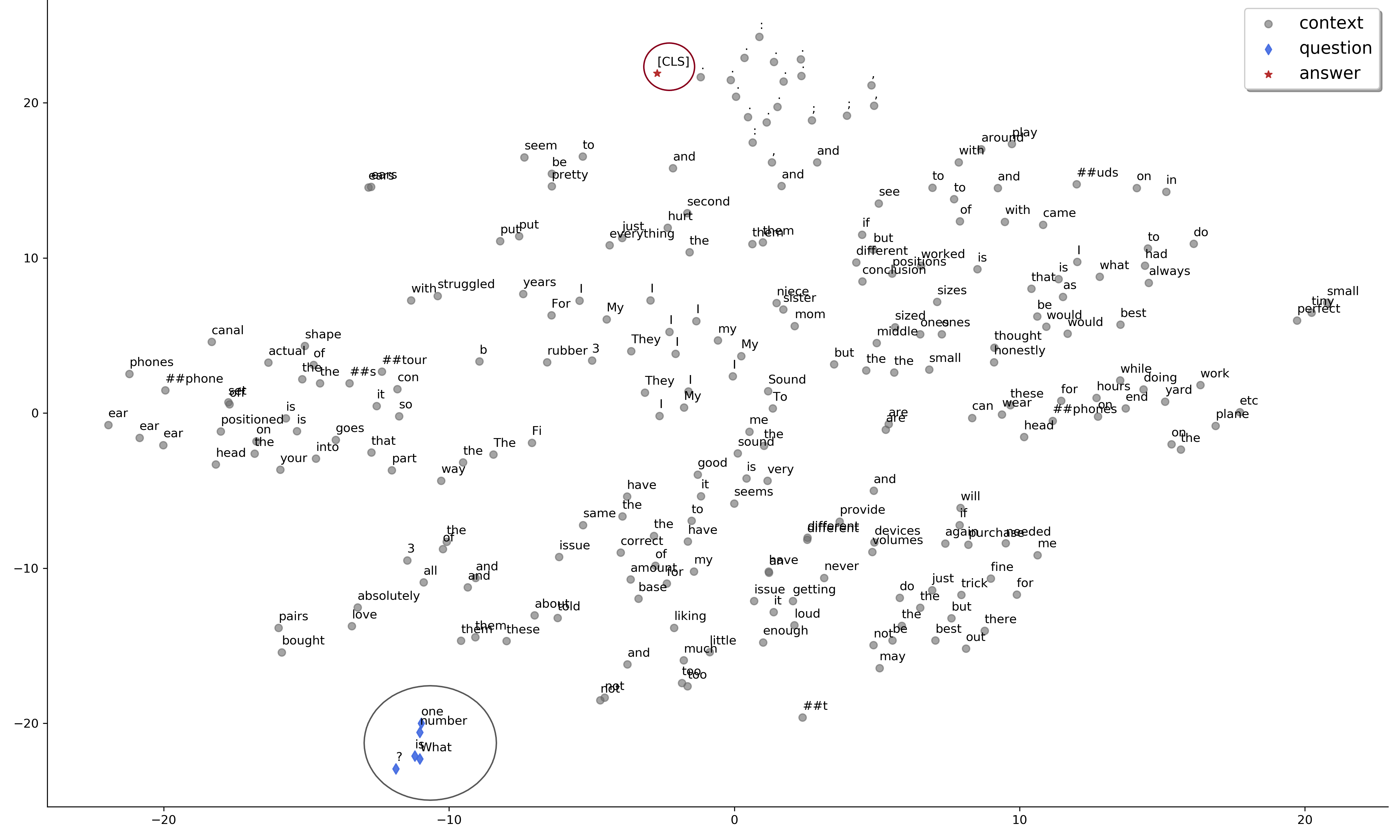}
    \caption[short]{Layer 3}
\end{subfigure}
%\begin{subfigure}{.55\textwidth}
    %\centering
    %\captionsetup{justification=centering}
    %\includegraphics[width=0.99\textwidth]{Figures/qualitative_analysis/hidden_reps_per_token/correct_prediction_unanswerable_question/tsne_question_context_qa_per_token_99_var_correct_unanswerable_layer_4_post_processed_no_title.png}
    %\caption[short]{Layer 4}
%\end{subfigure}%
\begin{subfigure}{.7\textwidth}
    \centering
    \captionsetup{justification=centering}
    \includegraphics[width=0.99\textwidth]{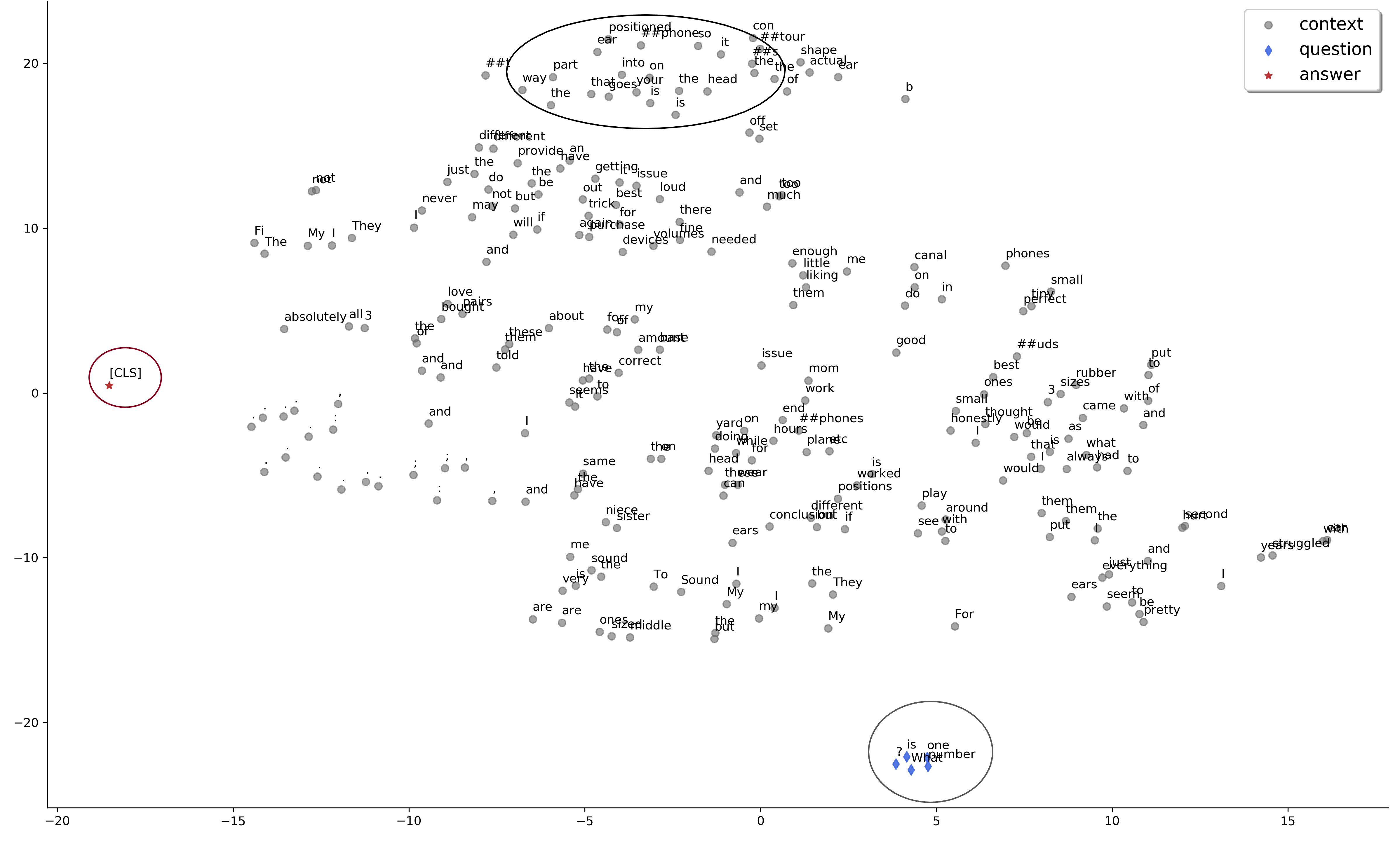}
    \caption[short]{Layer 5}
\end{subfigure}
%\begin{subfigure}{.55\textwidth}
    %\centering
    %\captionsetup{justification=centering}
    %\includegraphics[width=0.99\textwidth]{Figures/qualitative_analysis/hidden_reps_per_token/correct_prediction_unanswerable_question/tsne_question_context_qa_per_token_99_var_correct_unanswerable_layer_6_post_processed_no_title.png}
    %\caption[short]{Layer 6}
%\end{subfigure}
\caption{Correct answer-span prediction for an unanswerable question. Depicted are \textsc{bert}'s hidden representations at different stages of the model from bottom to top (i.e., Layer 1, 3, 5) projected into $\mathbf{R}^{2}$ for every token in a randomly chosen input sequence (q, c)$_{i}$ among the set of sentence pairs for which the model correctly predicted the special \texttt{[CLS]} token w.r.t. q$_{i}$ $\in \mathbf{q}_{unanswerable}$. Blue diamonds: \textcolor{blue}{question}. Red stars: \textcolor{red}{answer}. Grey circles: \textcolor{gray}{context}. (Q: "\texttt{what} is number one?", A: \texttt{[CLS]})}
\label{fig:hidden_reps_correct_pred_unanswerable_question}
\end{figure}

Looking at Figure~\ref{fig:hidden_reps_correct_pred_unanswerable_question} unveils that it was fairly easy to separate the vector representation of the special \texttt{[CLS]} token from the rest of the text. Recall that the \texttt{[CLS]} token must be predicted, if and only if a question is not answerable from the given context. The model is therefore required to only predict a single token which by the mere length of the answer span is easier than predicting multiple tokens. Again, tokens corresponding to the question were projected into a space that is distinguishable from the spatial representations of context and answer.

\begin{figure}
\centering
\begin{subfigure}{.7\textwidth}
    \centering
    \captionsetup{justification=centering}
    \includegraphics[width=0.99\textwidth]{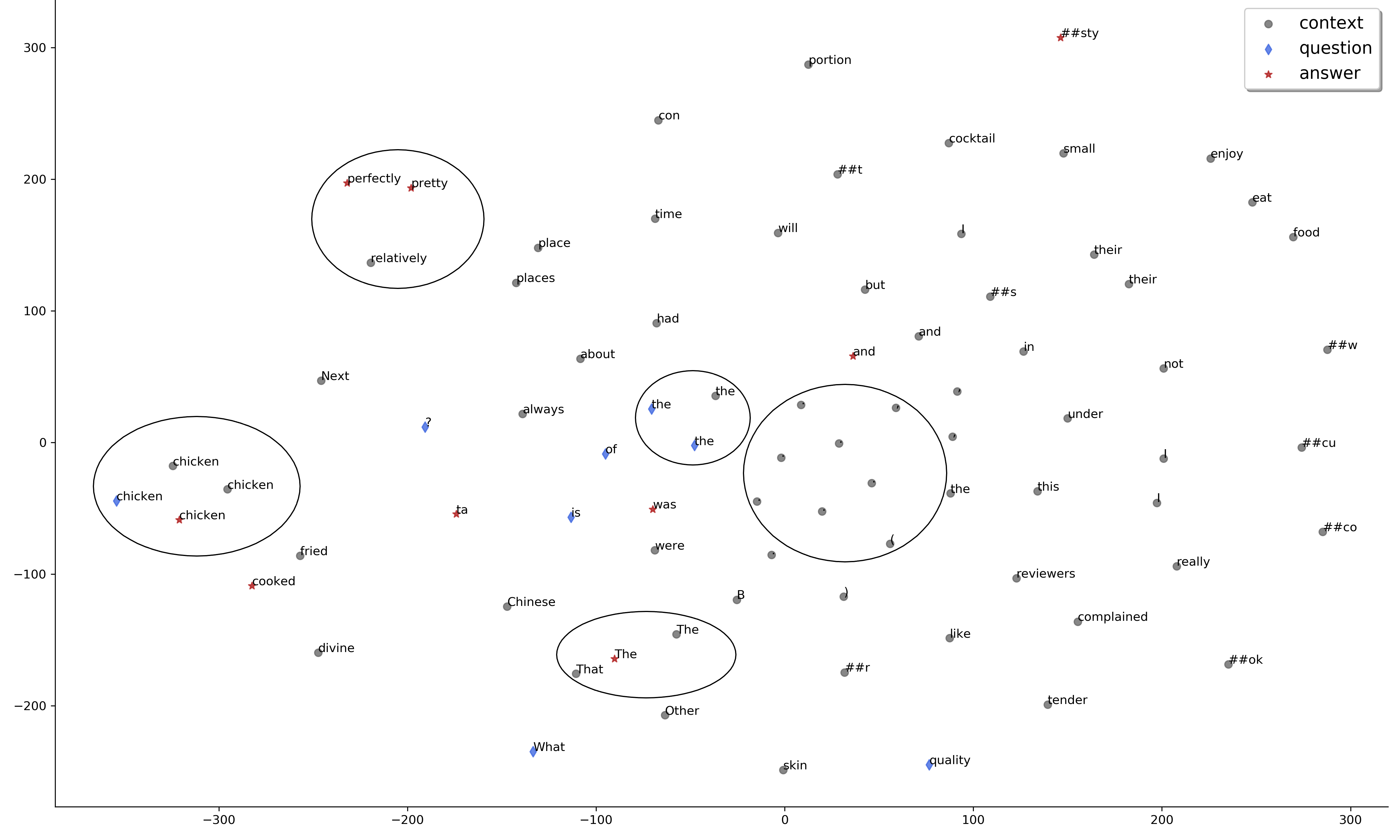}
    \caption[short]{Layer 1}
\end{subfigure}
\begin{subfigure}{.7\textwidth}
    \centering
    \captionsetup{justification=centering}
    \includegraphics[width=0.99\textwidth]{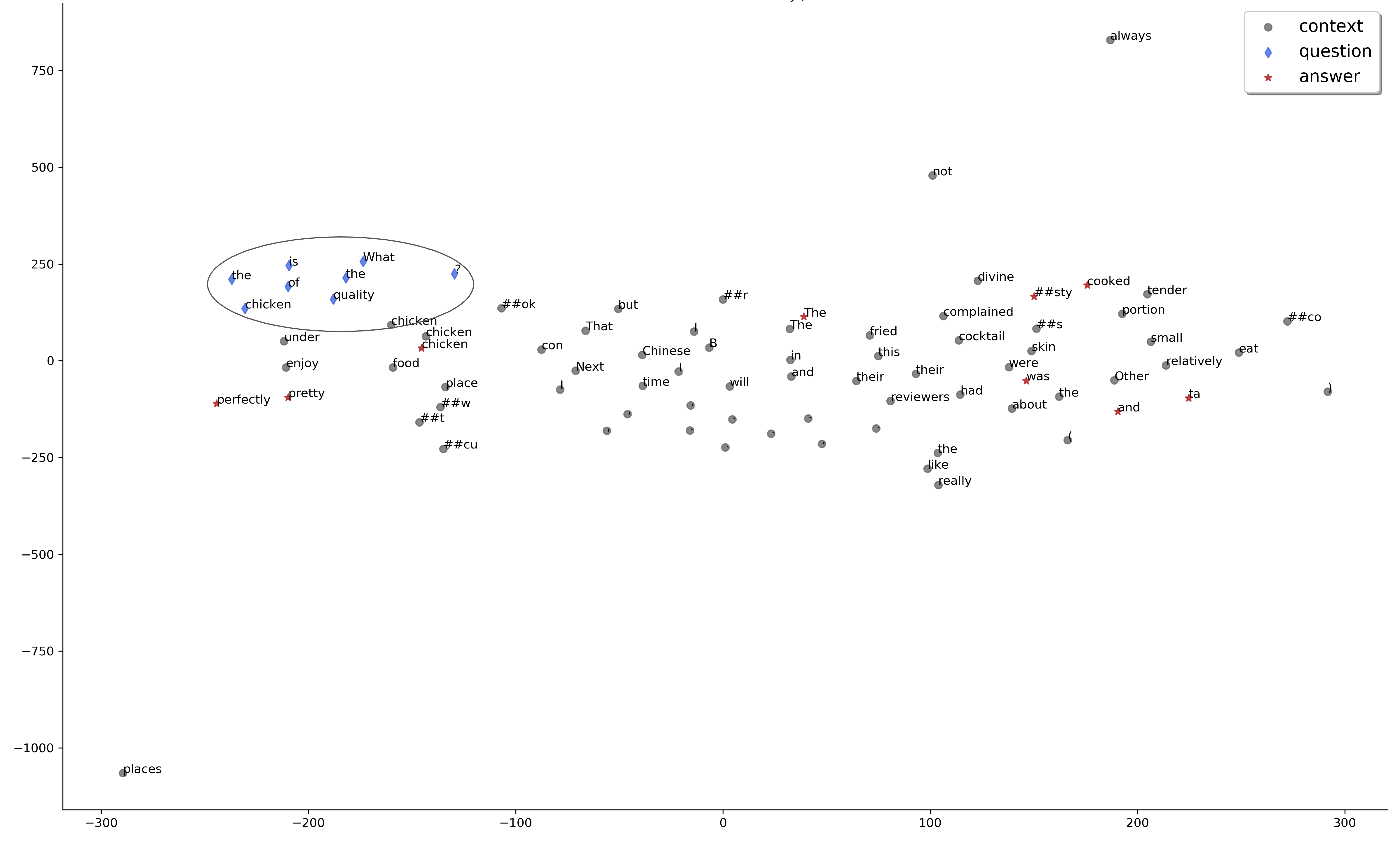}
    \caption[short]{Layer 3}
\end{subfigure}
%\begin{subfigure}{.55\textwidth}
    %\centering
    %\captionsetup{justification=centering}
    %\includegraphics[width=0.99\textwidth]{Figures/qualitative_analysis/hidden_reps_per_token/wrong_prediction/tsne_question_context_qa_per_token_99_var_wrong_layer_4_post_processed_no_title.png}
    %\caption[short]{Layer 4}
%\end{subfigure}%
\begin{subfigure}{.7\textwidth}
    \centering
    \captionsetup{justification=centering}
    \includegraphics[width=0.99\textwidth]{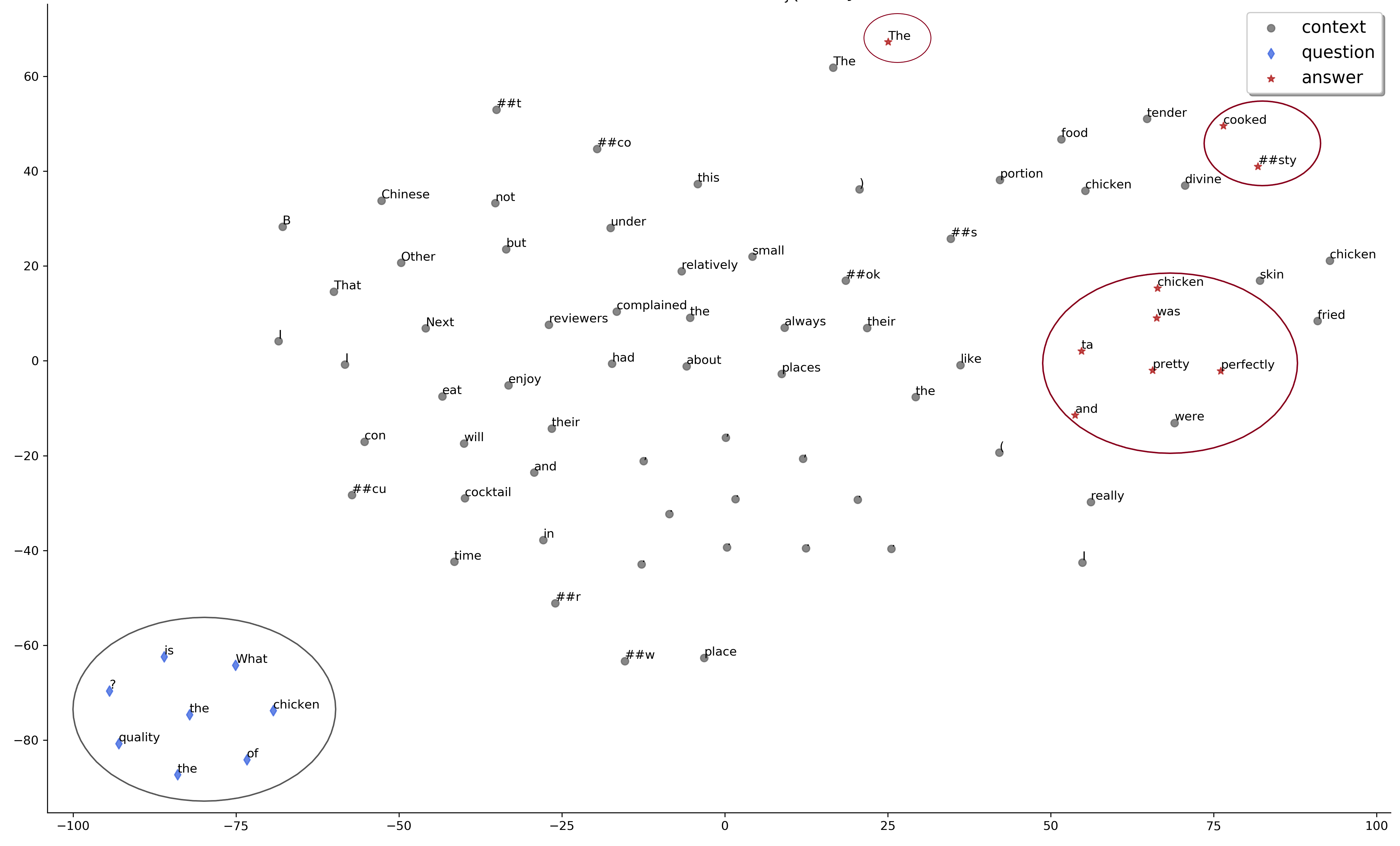}
    \caption[short]{Layer 5}
\end{subfigure}
%\begin{subfigure}{.55\textwidth}
    %\centering
    %\captionsetup{justification=centering}
    %\includegraphics[width=0.99\textwidth]{Figures/qualitative_analysis/hidden_reps_per_token/wrong_prediction/tsne_question_context_qa_per_token_99_var_wrong_layer_6_post_processed_no_title.png}
    %\caption[short]{Layer 6}
%\end{subfigure}
\caption{Erroneous answer-span prediction for an answerable question. Depicted are \textsc{bert}'s hidden representations at different stages of the model from bottom to top (i.e., Layer 1, 3, 5) projected into $\mathbf{R}^{2}$ for every token in a randomly chosen input sequence (q, c)$_{i}$ among the set of sentence pairs for which the model could not predict the answer-span a$_{i}$ w.r.t. q$_{i}$ $\in \mathbf{q}_{answerable}$. Blue diamonds: \textcolor{blue}{question}. Red stars: \textcolor{red}{answer}. Grey circles: \textcolor{gray}{context}. (Q: "\texttt{what} is the quality of the chicken?", A: "The chicken was perfectly cooked and pretty tasty")}
\label{fig:hidden_reps_wrong_pred_answerable_question}
\end{figure}
\newpage

\subsubsection{Answer vector agreements} 
\label{section:ans_vec_agreements}

As depicted and analyzed in the previous section, the model's hidden representations with respect to the answer span are clustered more closely in vector space for correct compared to wrong answer span predictions (see Figures~\ref{fig:hidden_reps_correct_pred_answerable_question},~\ref{fig:hidden_reps_wrong_pred_answerable_question}). This is particularly visible in the top three layers of the model, where generally high-level rather than low-level linguistic features are represented. To verify this observation more quantitatively, I've computed the average cosine similarities among all hidden representations for each token in the answer span whenever the correct answer contained more than one token. Hence, the following analysis was conducted exclusively for answerable questions since the correct answer span for unanswerable questions corresponds to the special \texttt{[CLS]} token.

Prior to the latter computation, I've removed all feature representations corresponding to the special \texttt{[PAD]} token and transformed the matrix of hidden representations $\mathbf{H}_{i} \in \mathbf{R}^{T \times D}$\footnote{$T$ is equal to the number of tokens in the word sequence $(\mathbf{q, c})_{i}$ without appended \texttt{[PAD]} tokens and $D$ = $768$ which is the model's hidden size in each layer.} for each sentence pair sequence $(\mathbf{q, c})_{i}$ into a lower-dimensional space to remove noise and exclusively keep those principal components that explain the most variance among the feature representations. In so doing, I've leveraged Principal Components Analysis (PCA) \cite{DBLP:journals/corr/Shlens14} and retained 95\% of the hidden representations' variance. In initial experiments, I've experimented with retaining only 90\% of the variance but results obtained from those transformations revealed less insightful analyses. Thus, I've proceeded with maintaining 95\% of the features' variance throughout all analyses.  This yielded a matrix of transformed hidden representations $\tilde \mathbf{H}_{i} \in \mathbf{R}^{T \times P}$, for each sentence pair $(\mathbf{q, c})_{i}$. From the transformed matrix of hidden representations, I've extracted the matrix of hidden representations corresponding to answer span tokens $\tilde \mathbf{H}_{a(i)} \in \mathbf{R}^{T_{a} \times P}$ to compute the average cosine similarity solely across all answer vectors. The average cosine similarity among the rows (i.e., vectors) of the matrix $\tilde \mathbf{H}_{a(i)} \in \mathbf{R}^{T_{a} \times P}$ was computed as follows:

\begin{equation}
    \cos_{a(i)} = \frac{1}{T_{a}T_{a}-T_{a}} \sum_{j}^{T_{a}} \sum_{k (k \neq j)}^{T_{a}} \cos(H_{a(i)}^{j}, H_{a(i)}^{k}) \in \mathbf{R}^{P},
\label{equation:average_cosine}
\end{equation}

where the cosine similarity between two non-zero vectors $\mathbf{u}$ and $\mathbf{v}$ is defined such as:

\begin{equation}
    \cos (\theta)=\frac{\mathbf{u} \cdot \mathbf{v}}{\|\mathbf{u}\|\|\mathbf{v}\|}=\frac{\sum_{i=1}^{n} u_{i} v_{i}}{\sqrt{\sum_{i=1}^{n} u_{i}^{2}} \sqrt{\sum_{i=1}^{n} v_{i}^{2}}}
\label{equation:cosine_sim}
\end{equation}

The cosine similarity evaluates to a normalized dot product (i.e., dot product normalized by the magnitudes of the respective vectors) between two non-zero vectors and ranges from $-1$ (exactly opposite) to $1$ (exactly the same), where $0$ refers to orthogonal or decorrelated vector pairs. Hence, the cosine similarity between two vectors is always in the interval [$-1$, $1$]. The closer the cosine similarity is to $1$, the more similar two vectors are, and vice versa, the closer it is to $-1$, the more dissimilar the vector pair is. The latter computation (see Equation~\ref{equation:average_cosine}) was performed for the two sets of correct and erroneous answer span predictions separately to inspect potential differences between the two with respect to their average cosine similarities. This was done at each transformer layer $l \in L$, where $L$ = $6$, to examine shifts in the cosine similarity distributions across space. 

\begin{figure}[h!]
\centering
\begin{subfigure}{.53\textwidth}
    \centering
    \captionsetup{justification=centering}
    \includegraphics[width=0.95\textwidth]{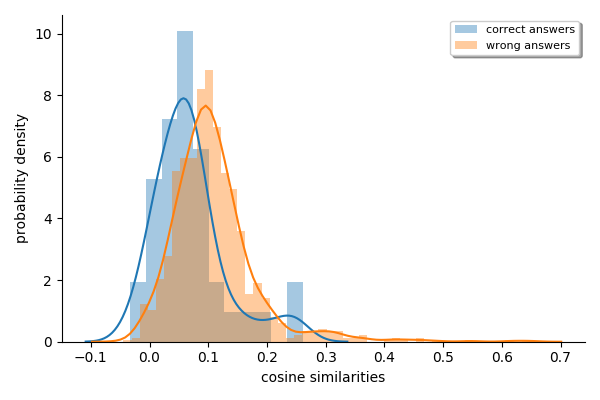}
    \caption[short]{Layer 1 \: ($p < 1e-4$)***}
\end{subfigure}%
\begin{subfigure}{.53\textwidth}
    \centering
    \captionsetup{justification=centering}
    \includegraphics[width=0.95\textwidth]{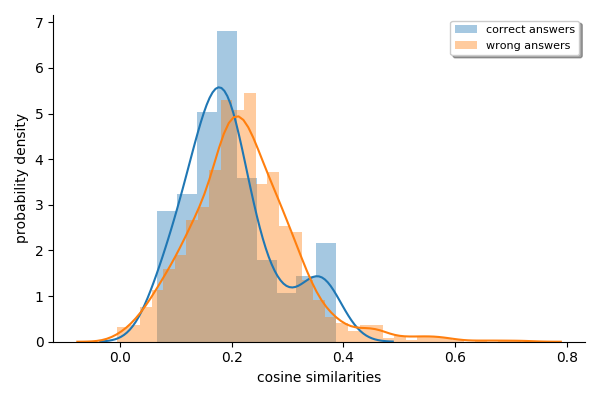}
    \caption[short]{Layer 2 \: ($p = .18$)*}
\end{subfigure}
\begin{subfigure}{.53\textwidth}
    \centering
    \captionsetup{justification=centering}
    \includegraphics[width=0.95\textwidth]{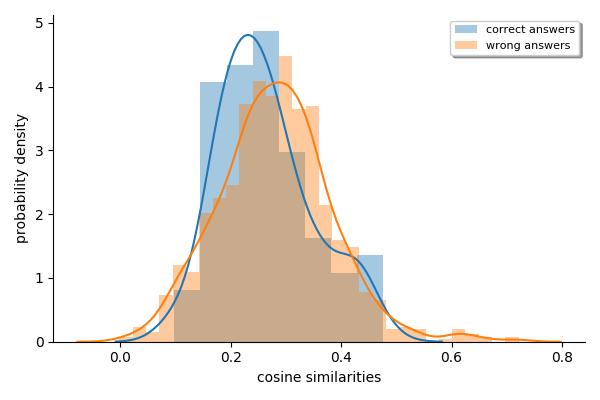}
    \caption[short]{Layer 3 \: ($p = .78$)}
\end{subfigure}%
\begin{subfigure}{.53\textwidth}
    \centering
    \captionsetup{justification=centering}
    \includegraphics[width=0.95\textwidth]{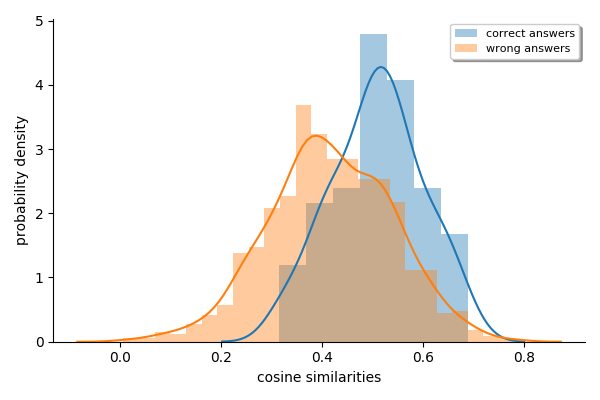}
    \caption[short]{Layer 4 \: ($p < 1e-8$)***}
\end{subfigure}
\begin{subfigure}{.53\textwidth}
    \centering
    \captionsetup{justification=centering}
    \includegraphics[width=0.95\textwidth]{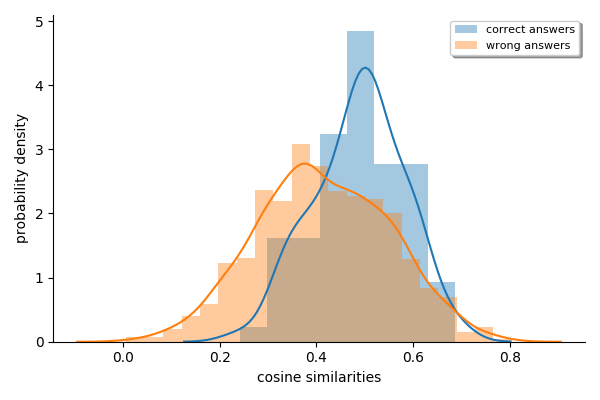}
    \caption[short]{Layer 5 \: ($p < 1e-5$)***}
\end{subfigure}%
\begin{subfigure}{.53\textwidth}
    \centering
    \captionsetup{justification=centering}
    \includegraphics[width=0.95\textwidth]{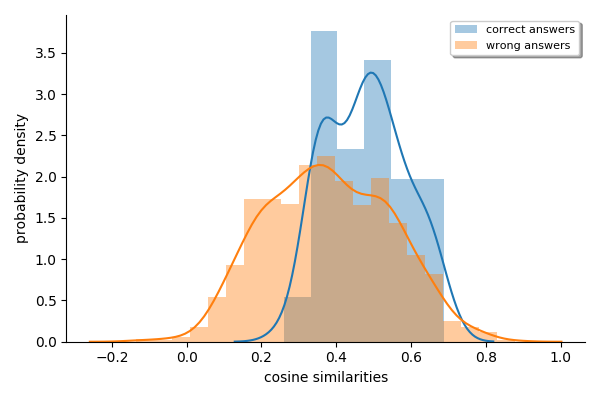}
    \caption[short]{Layer 6 \: ($p < 1e-6$)***}
\end{subfigure}
\caption{Probability density functions (PDFs) with respect to the average \texttt{cosine} similarities among hidden representations for each answer span token at each layer in the $D_{test} \in$ SubjQA, split into correct and erroneous model predictions. The respective model was fine-tuned on SubjQA. Cosine similarity distributions across the different layers are represented from left-to-right and top-to-bottom in the usual bottom-up representational hierarchy starting at the first (1\textsuperscript{st}) (top-left) and stopping at the last (6\textsuperscript{th}) (bottom-right) layer. Blue: \textcolor{blue}{correct} answers. Orange: \textcolor{orange}{wrong} answers. \textit{p}-values to the right of each caption refer to the difference significance w.r.t. the mean \texttt{cosine} similarities according to independent \textit{t}-tests (\textit{p} $<$ .05 = *, \textit{p} $<$ .01 = **, \textit{p} $<$ .001 = ***).}
\label{fig:subjqa_cosine_sims_answer_spans_pdf}
\end{figure}

\paragraph{Subjective questions} As can be inferred from the probability density functions (PDFs) with respect to $\cos_{a}$ for the hidden representations of a model that was fine-tuned and evaluated on SubjQA (see Figure~\ref{fig:subjqa_cosine_sims_answer_spans_pdf}), the difference between the two PDFs with respect to $\cos_{a}$ is statistically significant for the layers 4, 5, and 6 with $p < .001$ according to an independent \textit{t}-test. Bonferroni correction was applied to counteract the multiple comparisons problem, that is an increase in the likelihood of rare observations due to multiple statistical tests. Each observed $p$-value was multiplied by the number of tested hypotheses $m$, where $m = L = 6$. Hence, $p_{corrected} = p_{observed} \times m$. This $p$-value adjustment was applied to all subsequent statistical analyses.

Interestingly, the difference between PDFs with respect to $\cos_{a}$ for correct and wrong answer span predictions respectively at layer 1 also is statistically significant. This might, however, be an artifact  rather than insightful information as layer 1 exclusively reflects low-level linguistic features, and answer tokens are not expected to be clustered closely in latent space (see Figures~\ref{fig:hidden_reps_correct_pred_answerable_question}a,~\ref{fig:hidden_reps_correct_pred_unanswerable_question}a,~\ref{fig:hidden_reps_wrong_pred_answerable_question}a). At both hidden layers 2 and 3, $\cos_{a}$ appears to be equally distributed for correct and erroneous answer predictions respectively. In layers 4, 5, and 6, however, $P(\cos_{a} > 0.5)$ is notably higher for correct compared to incorrect predictions (see Figure~\ref{fig:subjqa_cosine_sims_answer_spans_pdf}d, e, f). Hence, the probability of a high cosine similarity among  elements of the answer vector matrix $\tilde \mathbf{H}_{a(i)} \in \mathbf{R}^{T_{a} \times P}$ is significantly larger for correct compared to erroneous answers. This is in line with the answer token clusters in 2D space (see Figures~\ref{fig:hidden_reps_correct_pred_answerable_question}b-c,~\ref{fig:hidden_reps_correct_pred_unanswerable_question}b-c,~\ref{fig:hidden_reps_wrong_pred_answerable_question}b-c). 

Another key difference between the matrices $\tilde \mathbf{H}_{a(i)} \in \mathbf{R}^{T_{a} \times P}$ corresponding to correct and erroneous answer predictions respectively is the fact that probability mass with respect to $\cos_{a}$ constantly travels through space towards the right of the $x$-axis for the former but appears almost stagnant for the latter. This indicates that the likelihood for $\cos_{a}$ being close to $1$ consistently increases through the layer hierarchy of the network whenever the model correctly predicted an answer span. In layer 1 most probability mass is distributed to $P(\cos_{a} < 0.1$) (see Figure~\ref{fig:subjqa_cosine_sims_answer_spans_pdf}a), whereas in both the penultimate and last layer the PDF is centered around $P(0.4 < \cos_{a} < 0.7$) (see Figure~\ref{fig:subjqa_cosine_sims_answer_spans_pdf}e, f). This pattern does not seem to be evident when the model did not get an answer span correct. The area of the PDFs corresponding to the latter scenario is spread in space more horizontally and hence does not clearly indicate a center. 
 
If an answer span was predicted correctly, their hidden representations were most probably clustered closely in vector space in the hidden layers 4, 5, and 6, and vice versa, if the hidden representations with respect to the answer span were clustered closely in vector space, the model most probably predicted the correct start and end positions. This, however, is not always the case, as can be inferred from the PDFs. Sometimes $\cos_{a}$ is high although the model did not predict the correct answer span. Why this happens goes beyond the scope of this project and is encouraged to be inspected in future studies. The probability for the latter scenario is, however, fairly low. To provide an example: At hidden layer 5, $P(\cos_{a} > 0.5) \approx 0.64$ and $P(\cos_{a} > 0.5)\approx 0.29$ for answer vectors corresponding to correct and incorrect model predictions respectively (see Figure~\ref{fig:subjqa_cosine_sims_answer_spans_pdf}e).

\begin{figure}[h!]
\centering
\begin{subfigure}{.53\textwidth}
    \centering
    \captionsetup{justification=centering}
    \includegraphics[width=0.95\textwidth]{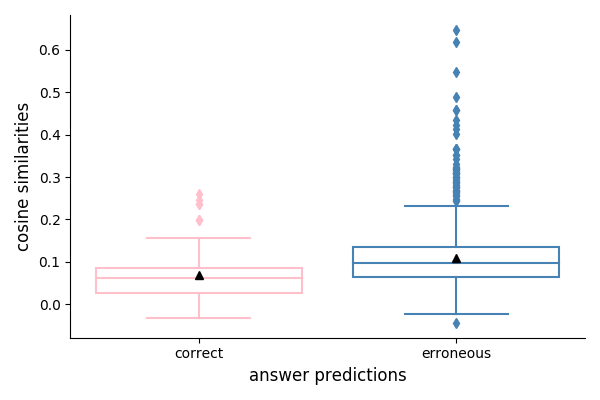}
    \caption[short]{Layer 1 \: ($p < 1e-4$)***}
\end{subfigure}%
\begin{subfigure}{.53\textwidth}
    \centering
    \captionsetup{justification=centering}
    \includegraphics[width=0.95\textwidth]{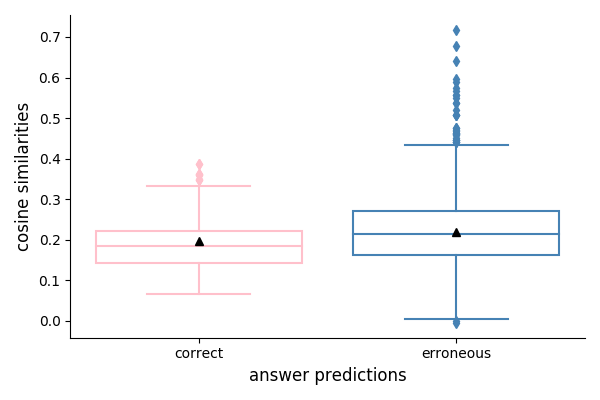}
    \caption[short]{Layer 2 \: ($p = .18$)}
\end{subfigure}
\begin{subfigure}{.53\textwidth}
    \centering
    \captionsetup{justification=centering}
    \includegraphics[width=0.95\textwidth]{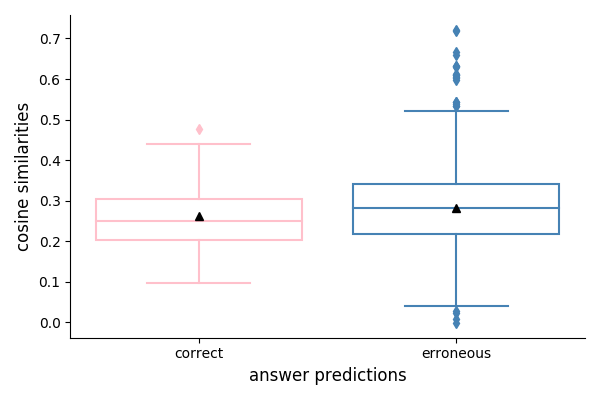}
    \caption[short]{Layer 3 \: ($p = .78$)}
\end{subfigure}%
\begin{subfigure}{.53\textwidth}
    \centering
    \captionsetup{justification=centering}
    \includegraphics[width=0.95\textwidth]{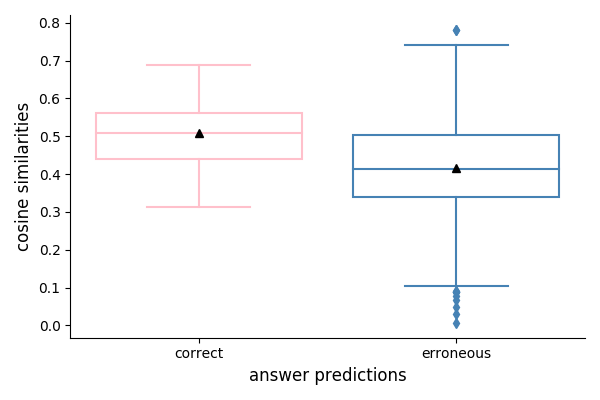}
    \caption[short]{Layer 4 \: ($p < 1e-8$)***}
\end{subfigure}
\begin{subfigure}{.53\textwidth}
    \centering
    \captionsetup{justification=centering}
    \includegraphics[width=0.95\textwidth]{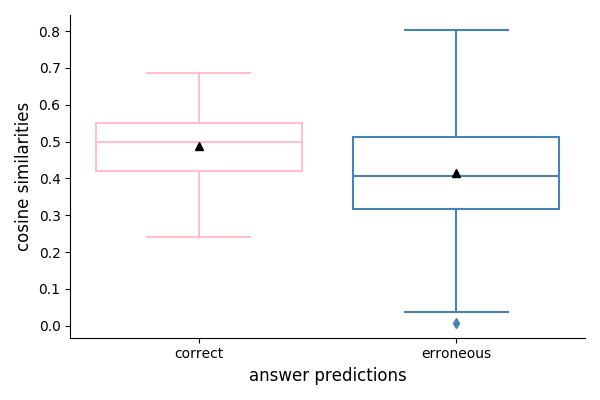}
    \caption[short]{Layer 5 \: ($p < 1e-5$)***}
\end{subfigure}%
\begin{subfigure}{.53\textwidth}
    \centering
    \captionsetup{justification=centering}
    \includegraphics[width=0.95\textwidth]{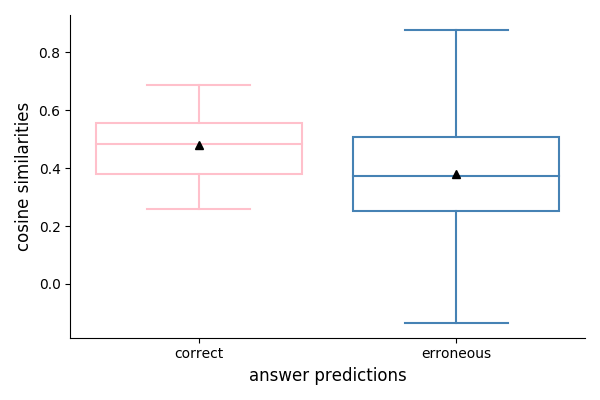}
    \caption[short]{Layer 6 \: ($p < 1e-6$)***}
\end{subfigure}
\caption{Box plots with respect to the average \texttt{cosine} similarities among hidden representations for each answer span token at each layer in $D_{test} \in$ SubjQA, split into correct and erroneous model predictions. The respective model was fine-tuned on SubjQA. Cosine similarity distributions across the different layers are represented from left-to-right and top-to-bottom in the usual bottom-up representational hierarchy starting at the first (1\textsuperscript{st}) (top-left) and stopping at the last (6\textsuperscript{th}) (bottom-right) layer. Pink: \textcolor{pink}{correct} answers. Blue: \textcolor{blue}{wrong} answers. $\bigtriangleup$: mean. \textit{p}-values to the right of each caption refer to the difference significance w.r.t. the mean \texttt{cosine} similarities according to independent \textit{t}-tests (\textit{p} $<$ .05 = *, \textit{p} $<$ .01 = **, \textit{p} $<$ .001 = ***).}
\label{fig:subjqa_cosine_sims_answer_spans_boxplot}
\end{figure}

The same pattern as depicted in the probability distributions is shown in the box plots (see Figure~\ref{fig:subjqa_cosine_sims_answer_spans_boxplot}). The difference between $\bar{\cos_{a}}$ corresponding to correct and incorrect answer span predictions respectively is statistically significant at hidden layers 4, 5, and 6 with $p < .001$ according to both an independent \textit{t}-test and a one-way ANOVA, and is higher for answer vectors corresponding to correct compared to incorrect predictions. Furthermore, the spread of $\cos_{a}$ values is notably higher for answer vectors corresponding to erroneous predictions as can be inferred from the whiskers of the box plots.

\begin{figure}[h!]
\centering
\begin{subfigure}{.53\textwidth}
    \centering
    \captionsetup{justification=centering}
    \includegraphics[width=0.95\textwidth]{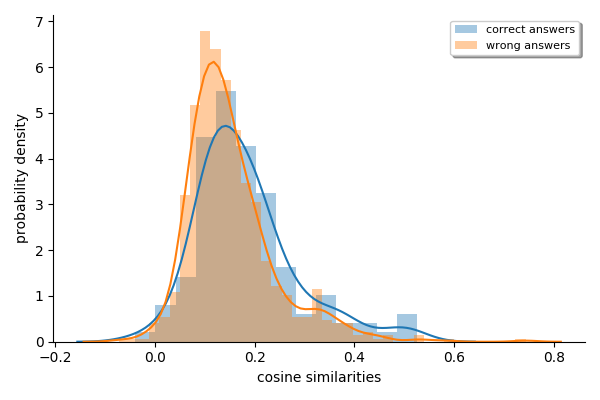}
    \caption[short]{Layer 1 \: ($p < 1e-2$)**}
\end{subfigure}%
\begin{subfigure}{.53\textwidth}
    \centering
    \captionsetup{justification=centering}
    \includegraphics[width=0.95\textwidth]{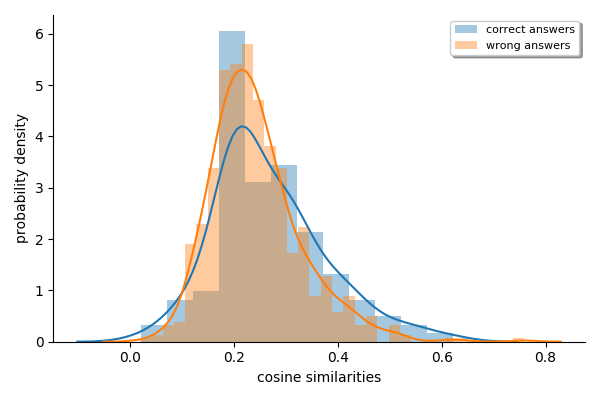}
    \caption[short]{Layer 2 \: ($p < .05$)*}
\end{subfigure}
\begin{subfigure}{.53\textwidth}
    \centering
    \captionsetup{justification=centering}
    \includegraphics[width=0.95\textwidth]{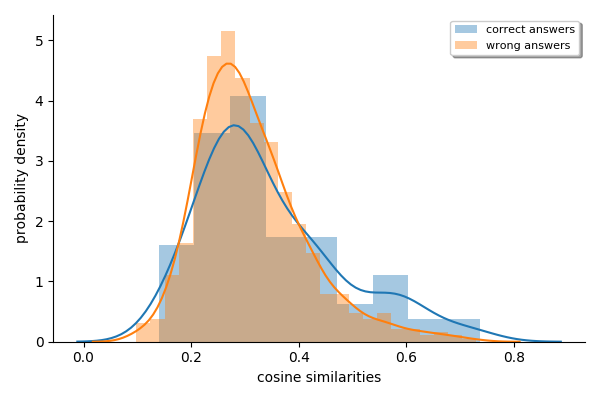}
    \caption[short]{Layer 3 \: ($p < .05$)*}
\end{subfigure}%
\begin{subfigure}{.53\textwidth}
    \centering
    \captionsetup{justification=centering}
    \includegraphics[width=0.95\textwidth]{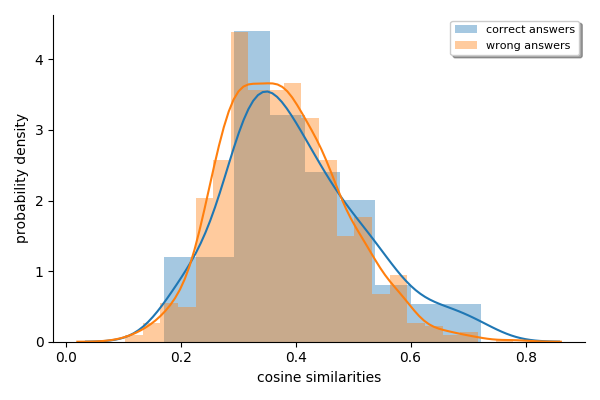}
    \caption[short]{Layer 4 \: ($p = .24$)}
\end{subfigure}
\begin{subfigure}{.53\textwidth}
    \centering
    \captionsetup{justification=centering}
    \includegraphics[width=0.95\textwidth]{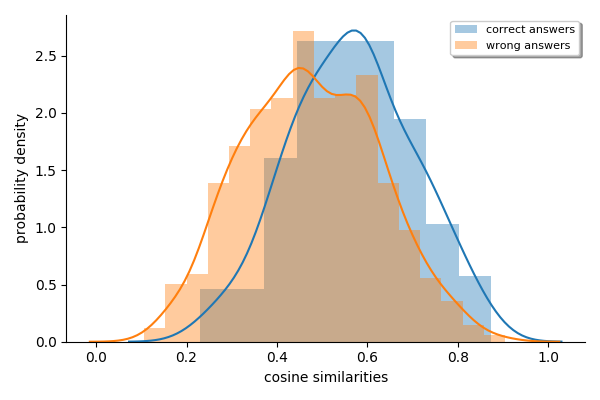}
    \caption[short]{Layer 5 \: ($p < 1e-8$)***}
\end{subfigure}%
\begin{subfigure}{.53\textwidth}
    \centering
    \captionsetup{justification=centering}
    \includegraphics[width=0.95\textwidth]{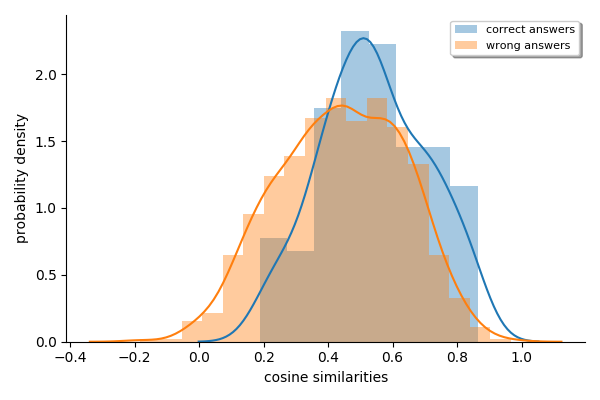}
    \caption[short]{Layer 6 \: ($p < 1e-7$)***}
\end{subfigure}
\caption{Probability density functions (PDFs) with respect to the average \texttt{cosine} similarities among hidden representations for each answer span token at each layer in the $D_{test} \in$ SQuAD, split into correct and erroneous model predictions. The respective model was fine-tuned on SQuAD. Cosine similarity distributions across the different layers are represented from left-to-right and top-to-bottom in the usual bottom-up representational hierarchy starting at the first (1\textsuperscript{st}) (top-left) and stopping at the last (6\textsuperscript{th}) (bottom-right) layer. Blue: \textcolor{blue}{correct} answers. Orange: \textcolor{orange}{wrong} answers. \textit{p}-values to the right of each caption refer to the difference significance w.r.t. the mean \texttt{cosine} similarities according to independent \textit{t}-tests (\textit{p} $<$ .05 = *, \textit{p} $<$ .01 = **, \textit{p} $<$ .001 = ***).}
\label{fig:squad_cosine_sims_answer_spans_pdf}
\end{figure}

\begin{figure}[h!]
\centering
\begin{subfigure}{.53\textwidth}
    \centering
    \captionsetup{justification=centering}
    \includegraphics[width=0.95\textwidth]{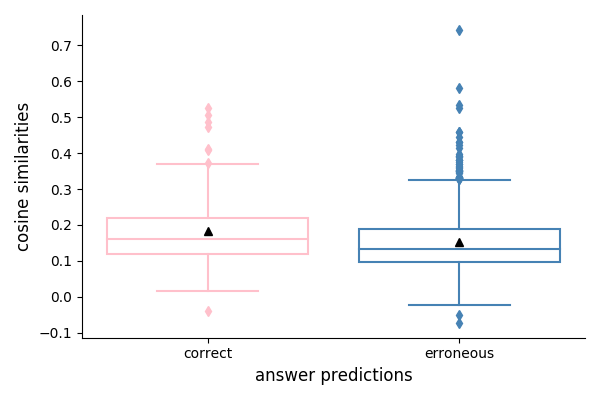}
    \caption[short]{Layer 1 \: ($p < 1e-2$)**}
\end{subfigure}%
\begin{subfigure}{.53\textwidth}
    \centering
    \captionsetup{justification=centering}
    \includegraphics[width=0.95\textwidth]{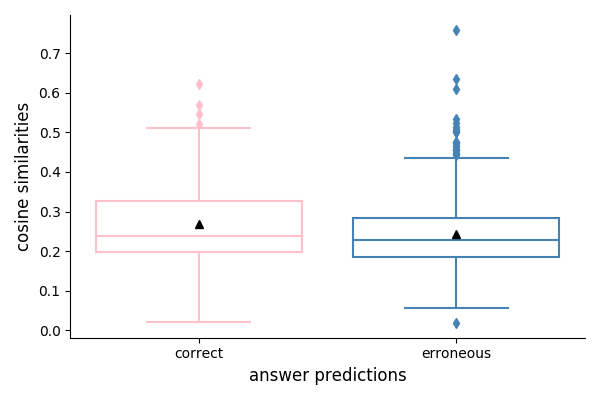}
    \caption[short]{Layer 2 \: ($p < .05$)*}
\end{subfigure}
\begin{subfigure}{.53\textwidth}
    \centering
    \captionsetup{justification=centering}
    \includegraphics[width=0.95\textwidth]{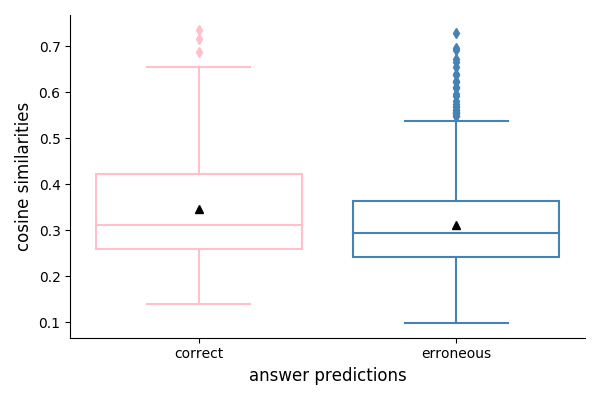}
    \caption[short]{Layer 3 \: ($p < .05$)*}
\end{subfigure}%
\begin{subfigure}{.53\textwidth}
    \centering
    \captionsetup{justification=centering}
    \includegraphics[width=0.95\textwidth]{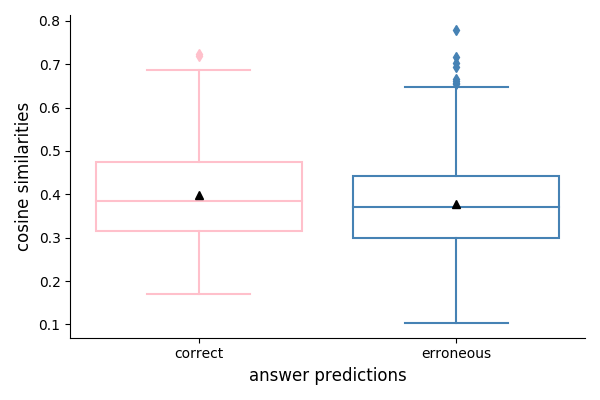}
    \caption[short]{Layer 4 \: ($p = .24$)}
\end{subfigure}
\begin{subfigure}{.53\textwidth}
    \centering
    \captionsetup{justification=centering}
    \includegraphics[width=0.95\textwidth]{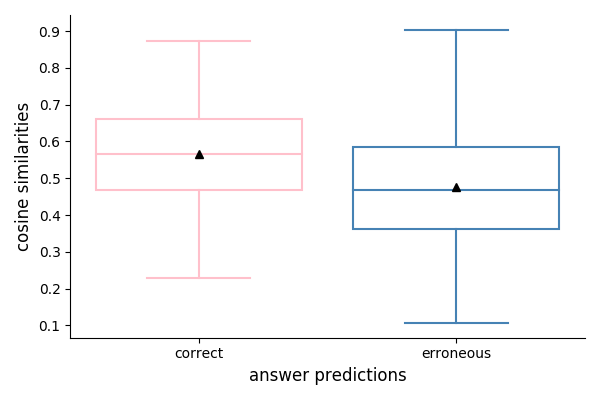}
    \caption[short]{Layer 5 \: ($p < 1e-8$)***}
\end{subfigure}%
\begin{subfigure}{.53\textwidth}
    \centering
    \captionsetup{justification=centering}
    \includegraphics[width=0.95\textwidth]{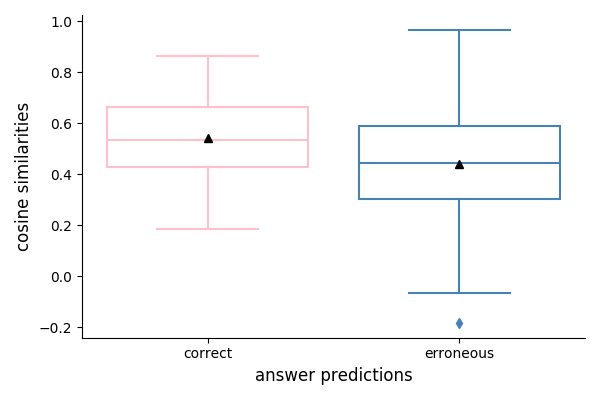}
    \caption[short]{Layer 6 \: ($p < 1e-7$)***}
\end{subfigure}
\caption{Box plots with respect to the average \texttt{cosine} similarities among hidden representations for each answer span token at each layer in $D_{test} \in$ SQuAD, split into correct and erroneous model predictions. The respective model was fine-tuned on SQuAD. Cosine similarity distributions across the different layers are represented from left-to-right and top-to-bottom in the usual bottom-up representational hierarchy starting at the first (1\textsuperscript{st}) (top-left) and stopping at the last (6\textsuperscript{th}) (bottom-right) layer. Pink: \textcolor{pink}{correct} answers. Blue: \textcolor{blue}{wrong} answers. $\bigtriangleup$: mean. \textit{p}-values to the right of each caption refer to the difference significance w.r.t. the mean \texttt{cosine} similarities according to independent \textit{t}-tests (\textit{p} $<$ .05 = *, \textit{p} $<$ .01 = **, \textit{p} $<$ .001 = ***).}
\label{fig:squad_cosine_sims_answer_spans_boxplot}
\end{figure}

\paragraph{Objective questions} I've performed the same computations as outlined above for the hidden representations of a model that was fine-tuned and evaluated on SQuAD (see Figures~\ref{fig:squad_cosine_sims_answer_spans_pdf},~\ref{fig:squad_cosine_sims_answer_spans_boxplot}) to investigate whether a different pattern holds for objective questions. The distributions with respect to $\cos_{a}$ are highly similar to the subjective case as can be inferred from both the PDFs (see Figure~\ref{fig:squad_cosine_sims_answer_spans_pdf}) and box plots (see Figure~\ref{fig:squad_cosine_sims_answer_spans_boxplot}). The only difference between subjective and objective question - context pairs is that $\cos_{a}$ with respect to objective questions is significantly higher for answer vectors corresponding to correct compared to incorrect predictions in layers 5, 6 - compared to layers 4, 5 and 6 with respect to subjective questions. Again, differences in the mean cosine similarities between correct and erroneous model predictions are significant according to an independent \textit{t}-test with $\alpha = .5$. $p$-values were adjusted following Bonferroni correction (see above). It seems as if the separation of objective answers from the context started later in latent space than the separation of subjective answers from its context, whenever the model correctly predicted start and end positions of an answer span (otherwise scarcely any separation is visible which holds for both types of questions).

%% file: discussion.tex
In this section, I elaborate on both quantitative and qualitative results, discuss caveats and shortcomings with respect to experiments and analyses, and envision potential directions for future research.

\section{General} 
In general, results have shown that it is crucial to fine-tune BERT on SubjQA to achieve state-of-the-art (SOTA) performance with respect to this dataset (see Table~\ref{tab:stl_main_results}). A BERT model that was previously fine-tuned on SQuAD appears to not generalize well to SubjQA. This is not surprising, given that SQuAD is a span-selection QA dataset that exclusively contains objective questions with respect to a single domain, namely Wikipedia \cite{DBLP:journals/corr/RajpurkarZLL16,DBLP:journals/corr/abs-1806-03822}. In contrast, SubjQA consists of questions that ask for subjective opinions with respect to six domains (see Table~\ref{tab:overview_squad_subjqa}) which requires different, or rather, additional abilities from a learner. This finding is in line with other work that emphasized the necessity of fine-tuning BERT on a specific dataset to achieve SOTA \cite{devlin2018bert,sanh2019distilbert,dodge2020finetuning,DBLP:journals/corr/abs-2002-12327}. 

\section{Single-task Learning} 
Quantitative analyses revealed insights into the performance of different model architectures. In single-task learning (STL) settings, additional \textsc{post-bert} encoding layers yielded an increase in $F$1 over the baseline model, which represents the most common QA set-up, where a fully-connected linear layer follows a pre-trained BERT feature extractor \cite{devlin2018bert,sanh2019distilbert}. This is in line with one recent study that has shown that an additional encoding of input sequences through Highway or recurrent layers improves QA performance for SQuAD \cite{hu2019question}. The best STL model in this work leveraged a BiLSTM \cite{hochreiter1997long} in-between the pre-trained BERT model and the task-specific linear output layer. Thus, one can draw the conclusion that the computation of temporal dependencies forward and backward in time yields an additional rise in performance when applied to BERT. This result is reasonable given the fact that temporal dependencies between tokens are not computed in Transformer based architectures \cite{DBLP:conf/nips/VaswaniSPUJGKP17,devlin2018bert} (see Section~\ref{section:transformers}) and have proven to unveil crucial information about the relationships between timesteps in temporal data \cite{birnns,DBLP:journals/corr/SeoKFH16,hu2019question} of which word sequences are a subset.

\section{Multi-task Learning} 
Rather surprisingly, models that were trained in multi-task learning (MTL) settings could not significantly improve over models trained in STL set-ups. This finding is against results obtained from previous studies concerning MTL \cite{DBLP:conf/eacl/PlankA17,DBLP:conf/nodalida/Bjerva17,DBLP:conf/eacl/SogaardB17,DBLP:journals/corr/Ruder17a}. Note, however, that none of these studies investigated neural architectures based on the Transformer \cite{DBLP:conf/nips/VaswaniSPUJGKP17} or examined QA behaviour. Recall that QA is a span-selection task whose nature is different from simple classification (i.e., sequence classification) and structured prediction (e.g., Part-of-Speech tagging, Named Entity Recognition) problems evaluated in the aforementioned  studies. Although my best performing model was a learner that was adversarially trained on the binary task of subjectivity classification in an MTL set-up, the differences to STL are not significant and may thus be considered some sort of regularization yielded through the injection of noise. This most likely had multiple reasons.

Firstly, the binary classification task of labeling sentence pairs as reflecting a subjective opinion, or containing an objective, measurable fact has proven to be not solvable by any of the implemented architectures. None of the models achieved an $F$1-score $> 54\%$ which is scarcely better than tossing a coin (see Table~\ref{tab:sbjclass_main_results}). More detailed analyses concerning this auxiliary task revealed that the poor performances are primarily due to the inability of a learner to distinguish between objective and subjective questions in SubjQA (see Table~\ref{tab:sbj_class_detailed} and Figure~\ref{fig:sbj_multi_conf_mats}). Hence, evidence suggests that the human-provided labels either lack quality and require revision, or are too ambiguous - owing to the nature of subjectivity itself - in order to be utilized for a binary sequence classification task. SubjQA shall rather be considered an entirely subjective dataset. Both questions and answers might show varying degrees of subjectivity, but they cannot be classified under the umbrella of objectivity. The latter is not only encouraged by the quantitative results but even more so by qualitative analyses of the model's hidden representation in vector space. 

The hidden representations of a model that was fine-tuned solely on subjectivity classification projected into $\mathbf{R}^2$ unveil that the model did not distinguish between objective and subjective questions with respect to SubjQA in latent space (see Figure~\ref{fig:sbj_class_multi_tsne}). The model could, however, perfectly separate objective questions belonging to SQuAD from any question belonging to SubjQA. One potential caveat with the former analysis is that the model might have simply learned differences between the two datasets and not between subjectivity and objectivity, and therefore clustered both objective and subjective questions belonging to SubjQA in the same latent space but separated questions with respect to SQuAD. To alleviate the latter objection, I've projected the hidden states of an MTL QA model that was adversarially trained on the binary task of classifying input sequences into their respective datasets with PCA and t-SNE into $\mathbf{R}^2$. The synthetic labels were assigned to a sentence pair after inference solely for the purpose of visualization in 2D space. Even in this dataset agnostic set-up, the transformed hidden representations depict a clear separation of objective questions belonging to SQuAD from questions with respect to SubjQA but show no differentiation within SubjQA at any layer stage (see Figure~\ref{fig:hidden_reps_cls_sbj_class_ds_agnostic}). Hence, it appears reasonable to modify the binary task of subjectivity classification into a regression problem, where subjectivity is measured on a continuous scale. Whether this helps in an MTL setting is to be evaluated in future studies.

Secondly, although each of the implemented model architectures could easily solve the multi-way classification problem of labeling question - context sentence pairs with one of the six review domains (see Table~\ref{tab:domainclass_main_results}) it seems as if knowledge about the latter task either maintained or deteriorated rather than enhanced performance (see Table~\ref{tab:mtl_domain_main_results}). Adversarial training with respect to context-domain classification in an MTL set-up made the model perform slightly better than without forcing the model to learn domain-invariant features, but did not help either. This might indicate that a Transformer based architecture such as BERT \cite{devlin2018bert} encodes domain-invariant features by itself without the necessity of an additional adversarially trained auxiliary task. The latter conclusion is in line with recent research about BERT \cite{DBLP:journals/corr/abs-2002-12327}, but requires more investigation into the feature representations of BERT and its ability to generalize across domains. This is an avenue of research I highly encourage to pursue as it will yield apprehension of both MTL and adversarial training for BERT in particular and Transformers in general.

\section{Interrogative Words \& Review Domains} 
Questions that start with \texttt{how} were the most difficult to answer (see Table~\ref{tab:results_per_q_word}) despite appearing more frequently in the train set of SubjQA than questions starting with any other interrogative word (see Table~\ref{tab:question_prefix_distrib}). This indicates that a considerable number of subjective questions begins with \texttt{how}, and subjective questions are difficult to answer in general. Why QA performance with respect to the domain \texttt{tripadvisor} was significantly worse compared to model performances across other domains is yet to be deciphered. This is another avenue I leave for future research that is concerned with SubjQA. An intuitive explanation, however, might be the fact that \texttt{tripadvisor} appeared considerably more often in the test set of SubjQA than any other domain, while being the least frequently encountered review domain during training (see Table~\ref{tab:domain_distribution}).

\section{Hidden Representations} 
Fine-grained error analyses unveiled that hidden representations in top layers with respect to the correct answer tokens are clustered closely in vector space and separated from the context for correct answer span prediction, and vice versa neither clustered together nor separated from the context for erroneous predictions (see Figures~\ref{fig:hidden_reps_correct_pred_answerable_question},~\ref{fig:hidden_reps_correct_pred_unanswerable_question},~\ref{fig:hidden_reps_wrong_pred_answerable_question}). The latter insights hold for both subjective and objective questions. These qualitative results are in line with one recently published study that has conducted a similar analysis of BERT's hidden representations in vector space with respect to SQuAD \cite{bert-qa-layerwise}. 

\section{Cosine Similarity Distributions} 
To the best of my knowledge, I am, however, the first to date who has investigated further into BERT's hidden representations with respect to both correct and erroneous answer predictions. I have analyzed the cosine similarities among answer token representations at every layer stage to both decipher why the aforementioned differences in latent space between the two prediction sets occur and provide a mathematical formalization to quantitatively verify the results obtained from the qualitative analyses (see Section~\ref{section:ans_vec_agreements}). Without the latter, one could not draw general conclusions. Probability density functions (PDFs) and box-plots displayed that there are no statistically significant differences between the two prediction sets with respect to their average cosine similarity among answer token hidden representations, $\cos_{a}$, (see Equation~\ref{equation:cosine_sim} for the computation of $\cos_{a}$), at early Transformer layers, namely layers 1, 2, and 3. There are, however, statistically significant differences between $\bar{\cos_{a}}$ corresponding to correct and erroneous answer predictions respectively at later Transformer layers, namely layers 4, 5, and 6 (see Figures~\ref{fig:subjqa_cosine_sims_answer_spans_pdf},~\ref{fig:subjqa_cosine_sims_answer_spans_boxplot},~\ref{fig:squad_cosine_sims_answer_spans_pdf},~\ref{fig:squad_cosine_sims_answer_spans_boxplot}). Note that these results reflect correlation rather than causation. As can be inferred from the PDFs (e.g., Figure~\ref{fig:subjqa_cosine_sims_answer_spans_pdf}), $P(\cos_{a} > .4)$ in layers 4, 5, and 6 is higher for correct answer predictions compared to erroneous predictions, and vice versa, the probability for a correct model prediction is greater when $\cos_{a}$ is high in top layers of the model. Hence, one can draw the conclusion that $\cos_{a}$ and $P(\hat{y} = y)$ are positively correlated.

If one leverages the knowledge about similarities of answer token representations in top layers of a Transformer model to anticipate whether an answer span prediction will be correct or erroneous, an erroneous answer could simply be skipped without the necessity to inspect its validity a posteriori. In a follow-up study that was performed alongside this master's thesis, we show that this insight has decisive implications for down-stream applications. In Muttenthaler et al. \cite{mutten-qa2020}, we propose an unsupervised evaluation method that can almost faultlessly predict the correctness of an answer span prediction, without the need for any labeled data at inference time. This method might be applied to semi-automatic generation of QA datasets, where a predicted answer span could be considered as gold label, if it was identified as correct by our method. Hence, the need for tedious annotation work could notably be reduced.  

Moreover, instead of exclusively utilizing cross-entropy loss with respect to the model's output logits to train a neural network for QA, one could additionally exploit a different loss function (e.g., cosine-embedding loss) that aims at increasing the similarity between token representations with respect to the correct answer span in top layers of the model (e.g., in the penultimate and last Transformer layer). I plan to investigate further into this line of research in follow-up work.
\newpage

\section{Conclusions}
\label{section:conclusions}

To summarize, the main conclusions that can be drawn from this study are as follows.

\begin{enumerate}
    \item Fine-tuning on a dataset that contains subjective questions is indispensable to answer subjective questions since objective questions do not appear to generalize to the realm of subjectivity. This answers \hyperref[section:rq]{\textbf{RQ}} 1.
    \item Encoding temporal dependencies among tokens forward and backward in time as a \textsc{post-bert} encoding step before performing QA enhances performance. This reveals insights about \hyperref[section:rq]{\textbf{RQ}} 2.
    \item Multi-task learning (MTL) does not appear to particularly enhance BERT's answering behavior. This is in part due to the impossibility of solving one of the two auxiliary tasks that were leveraged in this study. Hence, SubjQA shall be considered an entirely subjective dataset or a dataset with varying degrees of subjectivity that does not contain objective questions. This is most likely due to the ambiguity of crowd-sourced labels. I, therefore, encourage the measuring of subjectivity on a continuous scale and modifying of the binary classification task into a multi-class or regression problem accordingly. This unveils additional information that was not considered an area of examination prior to conducting experiments. Thus, \hyperref[section:rq]{\textbf{RQ}} 3 remains open to investigation with the exploitation of different auxiliary tasks, improved labels, and modified subjectivity measurements. 
    \item Questions that start with \texttt{how} or belong to the domain \texttt{tripadvisor} are the most difficult to solve in SubjQA. This corresponds to both \hyperref[section:rq]{\textbf{RQ}} 4 \& 5.
    \item A model's hidden representations with respect to the correct answer tokens are clustered more closely in low-dimensional vector space in top Transformer layers than answer token representations corresponding to erroneous predictions. With this insight, \hyperref[section:rq]{\textbf{RQ}} 6 can be regarded as investigated.
    \item The probability to achieve a high cosine similarity (e.g., $ > .5$) among the answer token vectors in latent space is significantly greater for correct compared to erroneous predictions. This may be considered an additional analysis that appeared insightful while scrutinizing \hyperref[section:rq]{\textbf{RQ}} 6. To the best of my knowledge, I am the first to date who has ever investigated this. In Muttenthaler et al. \cite{mutten-qa2020}, we have shown that this information can easily be leveraged for down-stream applications to predict correctness of an answer span prediction in Transformer-based models across two datasets and seven domains. Further research is encouraged to examine whether this holds across other QA datasets and domains.
\end{enumerate}
\newpage

%% file: summary.tex
First, I provided an \hyperref[section:overview]{\textbf{Overview}} of the sections in this thesis. Second, I introduced the overarching \hyperref[section:background_and_motivation]{\textbf{Topic}} of the thesis, namely QA with respect to subjective natural language questions, motivated the research goals and explained the general task of \hyperref[section:question_answering]{\textbf{QA}}. In so doing, I explored various neural network architectures such as the \hyperref[section:transformers]{\textbf{Transformer}}, which served as the foundation for each implemented model, \hyperref[method:birnns]{\textbf{RNNs}}, and \hyperref[method:highway]{\textbf{Highway Networks}}, all of which were exploited in the various experiments to encode token sequences differently and determine potential benefits. Third, I mentioned \hyperref[section:related_work]{\textbf{Research}} that is related to my analyses. Fourth, I first explained the mechanisms and conceptual details behind the leveraged models, namely BERT, RNNs, Highway networks, and later in Section~\ref{section:method} discussed the fine-tuning procedure, and most crucially clarified how the networks are optimized with respect to the particular tasks.
Fifth, I provided a thorough analysis of the utilized \hyperref[section:datasets]{\textbf{Datasets}}, discussed both characteristics of and differences between the two. Sixth, in Section~\ref{section:results}, I presented and explained results concerning all conducted experiments.The seventh section regarding  \hyperref[section:qualitative_analysis]{\textbf{Qualitative Analyses}} aimed at deciphering the network's feature representations in latent space with respect to both classes and individual sentence pairs. In so doing, I analyzed \texttt{why} and \texttt{where} along the way (i.e., in vector space) the selected model made mistakes in its predictions to investigate deeper into the model's errors. 
Finally, I \hyperref[section:discussion]{\textbf{discussed}} the results obtained from both quantitative and qualitative analyses as thoroughly as space and time allowed, drew conclusions and envisioned potential directions for future research.

%% file: acknowledgements.tex
This thesis was almost entirely written during the Covid-19 crisis. A few months back in time, I would have never expected to write my master's thesis mainly at my mother's apartment back in my hometown Vienna. As ever, life holds surprises, not even the human brain - which is considered to be the holy grail of intelligence - is able to foresee. I am incredibly grateful that my mother hosted me during this precarious time, and provided a desk on which I could conduct experiments, analyze data, inspect algorithms and let creativity flow as much as space and constrained conditions allowed. Special thanks to my wonderful supervisor, Johannes Bjerva, who has always supplied constructive feedback and listened to me when I needed it the most. Alexander Haseler and Nora Hollenstein have provided particularly valuable feedback on earlier drafts of this master's thesis, which definitely contributed to improved reading clarity. I am grateful that they took the time to read over my work. I would also like to thank my three closest friends, Julian Blumenschein, Tobias Kovats and Alicja Grządziel, who have provided lots of mental support during the lockdown. I don't know what my mind would have done without them. 
However, I have been lucky enough that the Covid-19 pandemic has neither affected me physically nor with respect to my future job prospects. Unfortunately, many people have lost their jobs, suffered from the virus, or even witnessed their beloved ones pass away. This thesis is dedicated to each of these people. We, as scientists, must draw attention to the ones who suffer, and strive for knowledge to alleviate their burdens.  This might sound idealistic, but this is most definitely one of the reasons why I have decided to pursue the path of science. I hope this piece of research contributed to this pursuit of knowledge, even if just a little. Nevertheless, I am grateful that I could work on a fulfilling, truly satisfying, and sometimes mind-bending project throughout the pandemic.